\documentclass[twoside,11pt]{article}
\usepackage{jmlr2e}

\usepackage{amsmath,epsfig}
\usepackage{nccmath}
\usepackage{multirow}
\usepackage{amssymb}
\usepackage{epstopdf}
\usepackage{algorithm}
\usepackage{algorithmic}
\usepackage{graphicx}
\usepackage{epstopdf}
\usepackage{subfig}
\usepackage{enumitem}
\usepackage{color,soul}
\usepackage{mathrsfs}
\usepackage{cases}
\usepackage{cleveref}
\usepackage{array}
\usepackage{amsthm}
\usepackage{cases}
\newtheoremstyle{definitionAnh}
  {14pt} 
  {14pt} 
  {} 
  {} 
  {\bfseries} 
  {} 
  {.5em} 
  {} 
\makeatletter
\g@addto@macro\th@definition{\thm@headpunct{}}
\makeatother
\theoremstyle{definitionAnh}
\newtheorem{theorem}{Theorem}
\newtheorem{proposition}[theorem]{Proposition}

\newcommand{\argmax}{\arg\!\max}
\newcommand{\ie}{i.e.}
\newcommand{\eg}{e.g.}
\makeatletter
\newcommand{\eqnum}{\leavevmode\hfill\refstepcounter{equation}\textup{\tagform@{\theequation}}}
\makeatother
\makeatletter
\newcommand{\specialcell}[1]{\ifmeasuring@#1\else\omit$\displaystyle#1$\ignorespaces\fi}
\makeatother
\crefname{equation}{equation}{equations}%

\ShortHeadings{Dynamic programming for Instance Annotation in MIML Learning}{Pham, Raich, and Fern}
\firstpageno{1}
\begin{document}

\title{Dynamic Programming for Instance Annotation in Multi-instance Multi-label Learning}

\author{\name Anh T. Pham \email phaman@eecs.oregonstate.edu \\
       \addr School of EECS\\
       Oregon State University\\
       Corvallis, OR 97331-5501, USA
       \AND
       \name Raviv Raich \email raich@eecs.oregonstate.edu \\
       \addr School of EECS\\
       Oregon State University\\
       Corvallis, OR 97331-5501, USA
       \AND
       \name Xiaoli Z. Fern \email xfern@eecs.oregonstate.edu \\
       \addr School of EECS\\
       Oregon State University\\
       Corvallis, OR 97331-5501, USA}

\editor{Editors}
\maketitle

\begin{abstract}
Labeling data for classification requires significant human effort. To reduce labeling cost, instead of labeling every instance, a group of instances (bag) is labeled by a single bag label. Computer algorithms are then used to infer the label for each instance in a bag, a process referred to as instance annotation. This task is challenging due to the ambiguity regarding the instance labels. We propose a discriminative probabilistic model for the instance annotation problem and introduce an expectation maximization framework for inference, based on the maximum likelihood approach. For many probabilistic approaches, brute-force computation of the instance label posterior probability given its bag label is exponential in the number of instances in the bag. Our key contribution is a dynamic programming method for computing the posterior that is linear in the number of instances. We evaluate our methods using both benchmark and real world data sets, in the domain of bird song, image annotation, and activity recognition. In many cases, the proposed framework outperforms, sometimes significantly, the current state-of-the-art MIML learning methods, both in instance label prediction and bag label prediction.
\end{abstract}


\section{Introduction}
Multiple instance multiple label (MIML) learning is a framework where learning is carried out under label uncertainty. Conventional single instance single label (SISL) learning assumes that each instance in the training data is labeled. In the MIML setting, instances are grouped into bags and labels are provided at the bag level. For example, an image can be viewed as a bag of segments tagged with names of objects present in the image (\eg,~`house', `grass', and `sky') without associating individual segments with a label. In bird species recognition from audio recording, bags are long intervals containing multiple syllables (instances). Intervals are labeled with list of species without providing an explicit label for each syllable. Various problems are considered in the MIML setting including: (i) learning a bag level label classifier and (ii) learning an instance level classifier. We refer readers to \citep{zhou2012multi} for a detailed review of MIML methods.

A bag level label classifier can be constructed without explicitly reasoning about the instance labels. This is the approach taken by MIMLBoost, MIMLSVM \citep{zhou2007multi}, Citation-kNN, and Bayesian-kNN \citep{wang2000solving}. In MIMLSVM, the training bags are first clustered using the Hausdorff distances among them. Then, each bag is encoded with a vector of similarities to each of the cluster centers. Finally, an SVM classifier is trained and used to predict the bag label of a new bag. A similar method using Hausdorff distance is applied in Citation-kNN and Bayesian-kNN. The training phase of the aforementioned methods does not provide an instance level classifier.

The focus of our paper is instance level label prediction, \ie,~instance annotation \citep{briggs2012, anh2014a}. Even though most MIML methods focus on bag level prediction, a few of the existing methods resort to instance level classifiers as a means to obtain bag level predictions. For example, M3MIML \citep{zhang2008m3miml} aims at maximizing the margin among classes where the score for each class is computed from score of bags and the score of each bag is computed from a single score-maximizing instance in the bag. As a result, M3MIML may not use information from many instances in each bag. A smaller number of methods directly aim at solving the instance annotation problem. For example, rank-loss support instance machines (SIM) \citep{briggs2012} considers both max score and softmax score taking into account all of the instance scores in each bag. Probabilistic graphical models have been proposed for the instance annotation problem in different applications. Due to the high computational complexity in the inference step, they employ approximation techniques, such as sampling \citep{nguyen2013multi, zha2008joint} and variational inference \citep{yang2009dirichlet}. \cite{foulds2011multi} propose a generative model with an exact inference based on the expectation maximization framework. However, compared to discriminative methods, generative methods often achieve lower accuracy \citep{vapnik1998statistical,jaakkola1999exploiting}.

We develop a discriminative probabilistic model with an efficient inference method that takes into account all instances in each bag. The contributions of this paper are as follows. First, we propose the discriminative ORed-logistic regression model for the instance annotation problem. Second, we propose an expectation maximization framework to facilitate maximum likelihood inference. Third, we introduce a computationally efficient and exact algorithm for posterior probability calculation in the E-step. Finally, we demonstrate the superiority of this approach over various domains such as bird song, image annotation, and activity recognition, for both bag level prediction and instance level prediction.

\section{Related work}
\begin{figure}
\center
\includegraphics[width=4.5in]{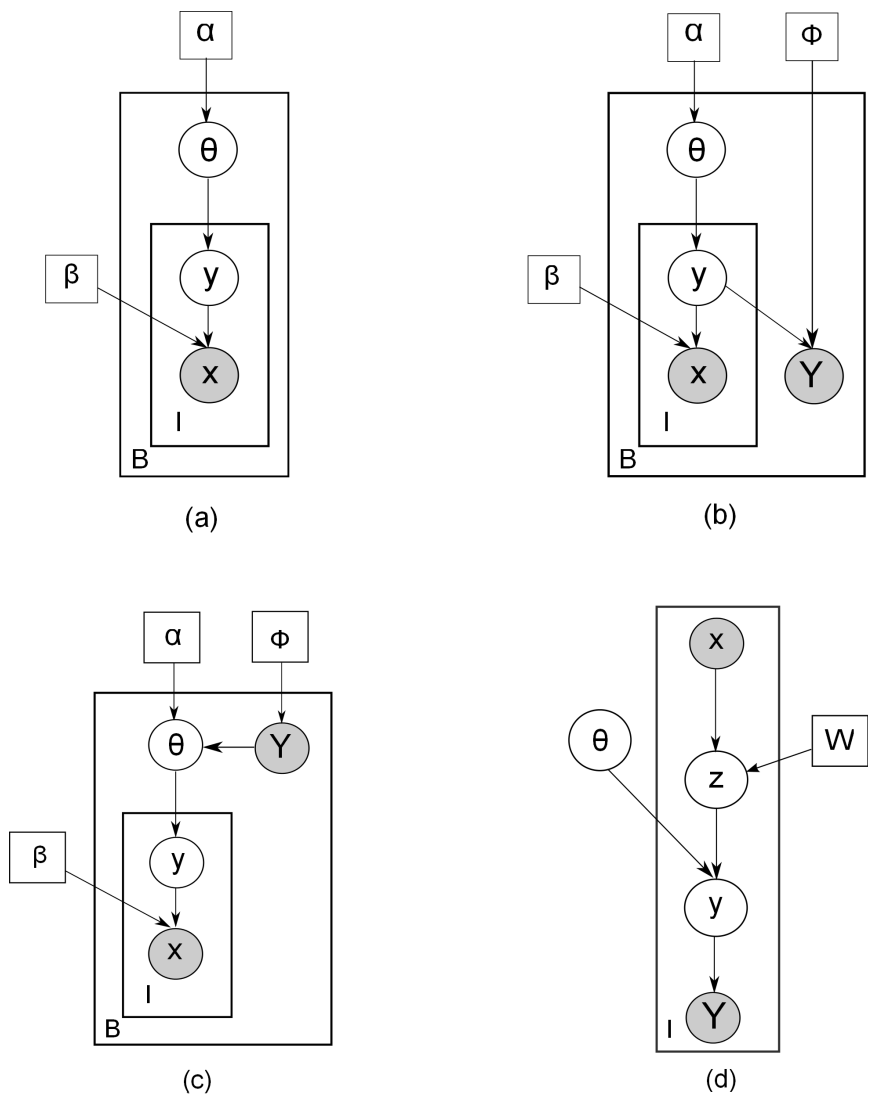}
\caption{Graphical models for (a) LDA, (b) supervised LDA, (c) labeled LDA, (d) and LSB-CMM. Square nodes denote parameters, circle nodes denote random variables. Shaded nodes denote observed variables. }
\label{fig:allmodels}
\end{figure}

Multi-instance multi-label learning problems have been implicitly addressed by {\em probabilistic graphical models}. In text data processing, Latent Dirichlet Allocation (LDA) \citep{blei2003latent} is a well-known generative topic model for processing a corpus of text documents. The graphical model for LDA is illustrated in Figure \ref{fig:allmodels}(a). For each document (bag), a topic proportion $\theta$ is generated. Then, from the topic proportion, a topic $y$ is randomly selected and a word (instance) $\textbf{x}$ is selected at random based on the topic. However, different from the MIML setting, LDA is an unsupervised model in which words are observed but their topics are hidden. Supervised/labeled LDA models incorporate a bag label \citep{mcauliffe2008supervised, ramage2009labeled}. In supervised LDA \citep{mcauliffe2008supervised}, as shown in Figure \ref{fig:allmodels}(b), the observed document label $\textbf{Y}$ is generated based on the hidden topics $y$ in that document. From the observation (observed labels and words) parameters are estimated using approximate maximum likelihood through variational expectation maximization. In labeled LDA \citep{ramage2009labeled}, as illustrated in Figure \ref{fig:allmodels}(c), the topic proportion $\theta$ of each document is generated based on the observed document label $\textbf{Y}$. For inference, labeled LDA uses collapsed Gibbs sampling to estimate parameters.

Several {\em maximum margin based methods} have been considered for MIML learning. ASVM-MIL \citep{yang2006region} poses the support vector machine (SVM) for region-based image annotation problem. The challenge lies in how to construct a bag level label error from instance level label errors. With an assumption that there are a large number of true positive instances in positive bags and hence a false negative at the instance level does not necessarily lead to a bag error. However, a false positive at the instance level absolutely results in a positive prediction for a negative bag. Consequently, ASVM-MIL approximates the bag level label error by the false positive. In practice, the assumption may be violated since there may be only one instance from the positive class in a positive bag. mi-SVM \citep{andrews2002support} uses a heuristic approach that alternates between updating instance level labels constrained on bag labels and maximizing the margin among hyperplanes obtained from the instance labels. MIMLfast \citep{huang2013fast} maximizes the margin among classes by maximizing difference of class scores which are evaluated from bag scores. To construct a score for each bag, MIMLfast uses only the instance which maximizes the scores. This principle may ignore information from most instances in each bag. Another maximum margin based method that uses the max principle is M3MIML \citep{zhang2008m3miml}. SIM \citep{briggs2012} uses both max principle and softmax principle where the score of each bag is computed from all instance scores by the softmax function.  \citet{briggs2012} show that using softmax score for SIM yields some performance improvement compared to using max score.

{\em Super set label learning} (SSLL) or partial label learning has been considered for instance annotation. Different from the instance annotation setting, SSLL does not consider the bag concept. Instead, from a MIML dataset, it transforms the data so that each instance label is replaced with the bag label. A limitation of this setting is that it cannot capture the assumption that bag labels are union of instance labels. LSB-CMM \citep{liu2012conditional} uses logistic stick breaking encoding scheme to link each instance vector to its labels. Due to the complexity of the model, variational expectation maximization and MCMC sampling are employed. The discriminative LSB-CMM model is presented in Figure \ref{fig:allmodels}(d). For each instance $\textbf{x}$, a mixture component $z$ is identified. From the mixture component, the instance hidden label $y$ is generated. Finally, the noisy superset label $\textbf{Y}$ is generated from that instance label. In the LSB-CMM model, each instance is considered independently therefore the model does not take advantage of the union relationship among instance labels in each bag. In addition, several SVM-based solutions have been proposed for the partial label learning setting such as SVM based partial label learning \citep{cour2011learning} and PL-SVM \citep{nguyen2008classification}.

In this paper, we would like to develop an instance annotation approach that addresses some of the aforementioned challenges associated with MIML methods. We would like to construct a probabilistic framework that uses instance class membership probability instead of using a single score-maximizing instance. Furthermore, even though generative models are well suited to deal with missing or small amounts of data, when the data size is sufficiently large, generative models are outperformed by discriminative models \citep{vapnik1998statistical,jaakkola1999exploiting,ng2001discriminative,taskar2002discriminative}. We are motivated by this argument to develop a discriminative model for instance annotation. Moreover, we would like to design a model sufficiently simple that allows for exact inference. Finally, we would like to maintain the relation among instance labels in each bag.

\begin{table}[H]
\centering
    \begin{tabular}{  l | l }
    \hline
    $B$                                     & the number of bags in the dataset\\
    $C$                                     & the number of classes in the dataset\\
    $N$                                     & the total number of instances in the dataset\\
    $\textbf{x}_{bi}$                       & the $i$th instance of the $b$th bag\\
    $y_{bi}$                                & the label for the $i$th instance of the $b$th bag\\
    $n_{b}$                                 & the number of instances for the $b$th bag\\
    $\textbf{X}_{b}$                        & the set of instances in the $b$th bag\\
    $\textbf{Y}_{b}$                        & the label for the $b$th bag\\
    $\textbf{y}_{b}$                        & $[y_{b1},y_{b2},\dots,y_{bn_b}]$\\
    $d$                                     & the dimension of every instance $\textbf{x}_{bi}$\\
    $\textbf{X}_{D}$                        & $\{\textbf{X}_{1},\textbf{X}_{2},\dots,\textbf{X}_{B}\}$\\
    $\textbf{Y}_{D}$                        & $\{\textbf{Y}_{1},\textbf{Y}_{2},\dots,\textbf{Y}_{B}\}$\\
    $\underline{\textbf{y}}$                & $\{\textbf{y}_{1},\textbf{y}_{2},\dots,\textbf{y}_{B}\}$\\
    $\textbf{w}_c$                          & instance level weight for class $c$\\
    $\textbf{w}$                            & $[\textbf{w}_1,\textbf{w}_2,\dots,\textbf{w}_C]$\\
    $\textbf{Y}_{b}^{k}$                    & $\bigcup_{i=1}^{k}y_{bi}$\\
    $\textbf{Y}_{b}^{\backslash k}$         & $\bigcup_{\substack{i=1;i\neq k}}^{n_b}y_{bi}$\\
    $\textbf{\L}_{\backslash c}$            & a set includes all labels in $\textbf{\L}$ excluding $c$\\
    \hline
    \end{tabular}
\caption{Notations used in this paper}
\label{table:notation}
\end{table}

\section{Problem formulation and the proposed model}
\label{s:pf}
This paper considers the instance annotation problem in the MIML framework. We consider a collection of bags and their labels $\{(\textbf{X}_b,\textbf{Y}_b)\}_{b=1}^B$. Specifically, each $\textbf{X}_b$ denotes the set of instance feature vectors of the $b$th bag, $\textbf{x}_{b1},\textbf{x}_{b2},\dots,$ and $\textbf{x}_{bn_b}$, where $\textbf{x}_{bi} \in \mathscr{X} \subseteq \mathbb{R}^d$ is the feature vector for the $i$th instance in the $b$th bag and $n_b$ denotes the number of instances in the $b$th bag. Moreover, the bag level label $\textbf{Y}_b$ is a subset of the set $\mathscr{Y}=\{1, 2, \dots, C\}$, where $C$ is the number of classes. To simplify the notation, we use $(\textbf{X}_D,\textbf{Y}_D)$ where $\textbf{X}_D=\{\textbf{X}_1,\dots, \textbf{X}_B\}$ and $\textbf{Y}_D=\{\textbf{Y}_1,\dots, \textbf{Y}_B\}$ as an abbreviated notation for $\{(\textbf{X}_b,\textbf{Y}_b)\}_{b=1}^B$. The goal of instance annotation is to train a classifier that maps an instance in $\mathscr{X}$ to a single label in $\mathscr{Y}$ under the MIML framework \ie,~given $(\textbf{X}_D,\textbf{Y}_D)$. Main notations used in this paper are in Table \ref{table:notation}.
\subsection{The proposed model: ORed-logistic regression}
\label{ssec:format}
\begin{figure}[H]
\center
\includegraphics[width=6in]{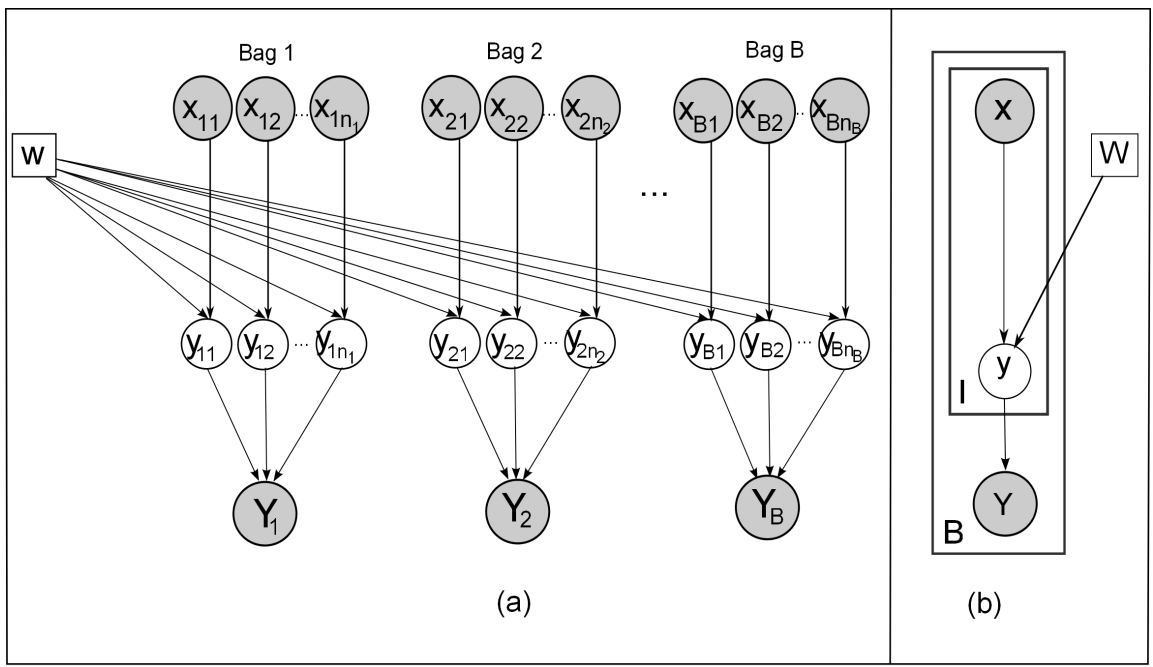}
\caption{(a) Graphical model for the proposed ORed-logistic regression model for instance annotation. (b) A simplified version of the proposed model. Observed variables are shaded. Square nodes denote parameters. }
\label{fig:MIML}
\end{figure}

The graphical representation of the proposed model is illustrated in Figure \ref{fig:MIML}. Following the notations of Section \ref{s:pf}, we assume that bags $\textbf{X}_1, \textbf{X}_2,\dots, \textbf{X}_B$ are independent and that instances in each bag $\textbf{x}_{b1},\textbf{x}_{b2},\dots,\textbf{x}_{bn_b}$ are independent: $p(\textbf{X}_1,\textbf{X}_2,\dots,\textbf{X}_B)=\prod_{b=1}^{B}p(\textbf{X}_b)=\prod_{b=1}^{B}\prod_{i=1}^{n_b}p(\textbf{x}_{bi})$. Next, we model the probability relationship between $y_{bi}$ the label of the $i$th instance in the $b$th bag and its feature vector $\textbf{x}_{bi}$ by a multinomial logistic regression function as follows
\begin{equation}
p(y_{bi}|\textbf{x}_{bi},\textbf{w})=\frac{\prod_{c=1}^{C}e^{I(y_{bi}=c)\textbf{w}_c^T\textbf{x}_{bi}}}{\sum_{c=1}^C{e^{\textbf{w}_c^T\textbf{x}_{bi}}}},
\label{e:prior}
\end{equation}
\noindent where $\textbf{w}_c \in \mathbb{R}^{d\times1}$ is the weight for the $c$th class score function\footnotemark \footnotetext{For simplicity, our derivations follow the linear relation $\textbf{w}^T\textbf{x}$. In practice, we implement a generalization to the affine model $\textbf{w}^T\textbf{x}+\beta$ as in \citet[p.~120]{hastie2009elements} by setting $\textbf{w}=[\textbf{w}^{T}, \beta]^{T}$ and $\textbf{x}=[\textbf{x}^{T}, 1]^{T}$.} and $\textbf{w}=[\textbf{w}_1,\textbf{w}_2,\allowbreak\dots,\textbf{w}_C]$. We use $I(\cdot)$ to denote the indicator function taking the value of $1$ when its argument is true and $0$ otherwise. Note that the probability of $\textbf{X}_1,\textbf{X}_2,\dots,\textbf{X}_B$ is not a function of $\textbf{w}$. This property is a key to the discriminative nature of our model. Next, we assume that the label of each bag is the union of its instance labels. Consequently, the probability of the label of each bag $\textbf{Y}_b$ given its instance labels $\textbf{y}_{b}=[y_{b1},y_{b2},\dots,y_{bn_b}]$ is expressed as follows
\begin{equation}
p(\textbf{Y}_b|\textbf{y}_{b})=I(\textbf{Y}_b=\bigcup_{j=1}^{n_b}y_{bj}).
\label{e:priornew}
\end{equation}
Based on the aforementioned description, the proposed model has the following properties. First, our model is a discriminative probabilistic model \citep{ng2001discriminative} since it learns a mapping from instance feature vectors to class labels as in \eqref{e:prior}. Second, the instance labels in each bag are constrained by their bag label using \eqref{e:priornew}. As a result, the model preserves the MIML structure from the dataset.

\subsection{Maximum Likelihood}
\label{sec:ml}
We consider the maximum likelihood principle for inference in our model. From the formula for conditional probability, $p(\textbf{Y}_D,\textbf{X}_D|\textbf{w})=p(\textbf{X}_D|\textbf{w})p(\textbf{Y}_D|\textbf{X}_D,\textbf{w})$. Moreover, based on our assumption that $\textbf{X}_D$ is independent of $\textbf{w}$, $p(\textbf{X}_D|\textbf{w})=p(\textbf{X}_D)$. Consequently, the probability of the observation given the unknown parameters is given by
\begin{equation}
\label{e:observationgivenvari}
p(\textbf{Y}_D,\textbf{X}_D|\textbf{w})=p(\textbf{X}_D)\prod_{b=1}^B p(\textbf{Y}_b|\textbf{X}_{b},\textbf{w}),
\end{equation}
where $p(\textbf{Y}_b|\textbf{X}_{b},\textbf{w})$ is obtained by marginalizing $p(\textbf{Y}_b, \textbf{y}_{b}|\textbf{X}_{b},\textbf{w})$ over $\textbf{y}_{b}$ as follows
\begin{equation}
p(\textbf{Y}_b|\textbf{X}_{b},\textbf{w})=\sum_{y_{b1}=1}^C\dots\sum_{y_{bn_{b}}=1}^C p(\textbf{Y}_b,\textbf{y}_{b}|\textbf{X}_{b},\textbf{w}).\label{e:observed1}
\end{equation}
From the conditional probability and from the graphical model that given $\textbf{y}_{b}$ the bag label $\textbf{Y}_b$ is independent of $\textbf{X}_{b}$ and $\textbf{w}$, we can rewrite \eqref{e:observed1} as
\begin{align}
p(\textbf{Y}_b|\textbf{X}_{b},\textbf{w})=&\sum_{y_{b1}=1}^C\dots\sum_{y_{bn_{b}}=1}^C p(\textbf{Y}_b|\textbf{y}_{b})p(\textbf{y}_{b}|\textbf{X}_{b},\textbf{w})\nonumber\\
=&\sum_{y_{b1}=1}^C\dots\sum_{y_{bn_{b}}=1}^C [I(\textbf{Y}_b=\bigcup_{j=1}^{n_b}y_{bj})\prod_{i=1}^{n_b}p(y_{bi}|\textbf{x}_{bi},\textbf{w})],\label{e:observed}
\end{align}
where the last step is obtained by substituting $p(\textbf{Y}_b|\textbf{y}_{b})$ from \eqref{e:priornew} and the assumption that instances in each bag are independent. Substituting \eqref{e:observed} back into \eqref{e:observationgivenvari} and taking the logarithm, the log-likelihood function can be written as
\begin{equation}
l_{MIML}(\textbf{w})=\sum_{b=1}^B\log(\sum_{y_{b1}=1}^C\dots\sum_{y_{bn_{b}}=1}^C [I(\textbf{Y}_b=\bigcup_{j=1}^{n_b}y_{bj})\prod_{i=1}^{n_b}p(y_{bi}|\textbf{x}_{bi},\textbf{w})])+\log p(\textbf{X}_D).\label{e:fullllh}
\end{equation}
Note that $\log p(\textbf{X}_D)$ is a constant w.r.t.~$\textbf{w}$ and hence does not affect the inference of $\textbf{w}$. In comparison, inference for the single instance single label setting where instance level labels are known can be done by maximizing the following log-likelihood function \citep{krishnapuram2005sparse,hastie2009elements}
\begin{equation}
l_{SISL}(\textbf{w})=\sum_{b=1}^B\sum_{i=1}^{n_b}[\sum_{c=1}^{C}I(y_{bi}=c)\textbf{w}_c^T\textbf{x}_{bi}-\log(\sum_{c=1}^C{e^{\textbf{w}_c^T\textbf{x}_{bi}}})]+\log p(\textbf{X}_D).\label{e:sisllearning}
\end{equation}
More detailed description of training a logistic regression in the SISL setting can be found in \cite{minka2003comparison}. To obtain the MIML maximum likelihood estimation of $\textbf{w}$, the log-likelihood in \eqref{e:fullllh} is maximized w.r.t.~$\textbf{w}$, and the maximizing $\hat{\textbf{w}}$ is selected, \ie,~$\hat{\textbf{w}}=\argmax_{\textbf{w}}{l_{MIML}(\textbf{w})}.$ Maximizing the log-likelihood in \eqref{e:fullllh} is difficult since to the best of our knowledge, no closed-form efficient solution exists. Therefore, we propose an expectation maximization (EM) approach \citep{moon1996expectation, dempster1977maximum} to maximize \eqref{e:fullllh}.

\section{EM for maximum likelihood inference}
Expectation maximization is a framework for indirectly maximizing the log-likelihood when there are hidden variables \citep{dempster1977maximum, moon1996expectation}. In the following, we introduce the EM approach (Section \ref{s:emr}) and present its application to inference for the proposed ORed-logistic regression (Ored-LR) model (Section \ref{ss:em}).
\subsection{Expectation maximization review}
\label{s:emr}
Denote the log-likelihood of the observation $\mathcal{X}$ given the parameter $\theta$ by $l(\theta)=\log p(\mathcal{X}|\theta)$. In EM, the hidden variable $\mathcal{Y}$ is introduced to develop a surrogate function to the log-likelihood given by $g(\theta,\theta')=E_{\mathcal{Y}}[\log p(\mathcal{X},\mathcal{Y}|\theta)|\mathcal{X},\theta']$. The EM framework alternates between the computation and the maximization of the surrogate function w.r.t.~$\theta$. Specifically, the two steps are:
\begin{itemize}[leftmargin=10pt]
\item E-step: Compute $g(\theta,\theta')=E_{\mathcal{Y}}[\log p(\mathcal{X},\mathcal{Y}|\theta)|\mathcal{X},\theta']$
\item M-step: $\theta^{(k+1)}$=$\argmax_{\theta}{g(\theta,\theta^{(k)})}$.
\end{itemize}
This paper uses the Generalized EM approach \citep{mclachlan2007algorithm} where in the M-step, instead of finding $\theta^{(k+1)}$ such that $\theta^{(k+1)}$=$\argmax_{\theta}{g(\theta,\theta^{(k)})}$, we obtain $\theta^{(k+1)}$ satisfying $g(\theta^{(k+1)},\theta^{(k)}) \ge g(\theta^{(k)},\theta^{(k)})$. As with EM, GEM guarantees non-decreasing log-likelihood $l(\theta^{(k)})$ as a function of $k$ \citep{mclachlan2007algorithm}.

\subsection{Expectation maximization for the proposed ORed-LR model}
\label{ss:em}
In our setting, the observed data $\mathcal{X}=\{\textbf{Y}_D, \textbf{X}_D\}$, the parameter $\theta=\textbf{w}$, and the hidden data $\mathcal{Y}=\underline{\textbf{y}}=\{\textbf{y}_{1},\textbf{y}_{2},\dots,\textbf{y}_{B}\}$. To compute the surrogate $g(\textbf{w},\textbf{w}')$, we begin with the derivation of the complete log-likelihood. We apply the conditional rule as follows
\begin{equation}
\begin{split}
p(\textbf{Y}_D,\textbf{X}_D,\underline{\textbf{y}}|\textbf{w})&=p(\textbf{Y}_D|\underline{\textbf{y}},\textbf{X}_D,\textbf{w})p(\underline{\textbf{y}}|\textbf{X}_D,\textbf{w})p(\textbf{X}_D|\textbf{w})\\
&=p(\textbf{Y}_D|\underline{\textbf{y}})[\prod_{b=1}^B\prod_{i=1}^{n_b}p(y_{bi}|\textbf{x}_{bi},\textbf{w})]p(\textbf{X}_D).
\label{e:joint}
\end{split}
\end{equation}
Then, the complete log-likelihood can be computed by taking the logarithm of \eqref{e:joint}, replacing $p(y_{bi}|\textbf{x}_{bi},\textbf{w})$ from \eqref{e:prior} into \eqref{e:joint}, and reorganizing as follows
\begin{align}
\log p(\textbf{Y}_D,\textbf{X}_D,\underline{\textbf{y}}|\textbf{w})=&\sum_{b=1}^B\sum_{i=1}^{n_b}\sum_{c=1}^{C}I(y_{bi}=c)\textbf{w}_c^T\textbf{x}_{bi}\label{e:llh}\\
&-\sum_{b=1}^B\sum_{i=1}^{n_b}\log(\sum_{c=1}^C{e^{\textbf{w}_c^T\textbf{x}_{bi}}})+\log p(\textbf{Y}_D|\underline{\textbf{y}})+\log p(\textbf{X}_D)\nonumber.
\end{align}
Finally, taking the expectation of \eqref{e:llh} w.r.t.~$\underline{\textbf{y}}$ given $\textbf{Y}_D$, $\textbf{X}_D$, and $\textbf{w}'$, we obtain the surrogate function $g(\cdot,\cdot)$ as follows
\begin{align}
&g(\textbf{w},\textbf{w}')=E_{\underline{\textbf{y}}}[\log p(\textbf{Y}_D,\textbf{X}_D,\underline{\textbf{y}}|\textbf{w})|\textbf{Y}_D,\textbf{X}_D,\textbf{w}']\label{e:sllh}\\
&=\sum_{b=1}^B\sum_{i=1}^{n_b}[\sum_{c=1}^{C}p(y_{bi}=c|\textbf{Y}_b,\textbf{X}_b,\textbf{w}')\textbf{w}_c^T\textbf{x}_{bi}-\log(\sum_{c=1}^C{e^{\textbf{w}_c^T\textbf{x}_{bi}}})]+ \zeta, \nonumber
\end{align}
where $\zeta=E_{\underline{\textbf{y}}}[\log p(\textbf{Y}_D|\underline{\textbf{y}})+\log p(\textbf{X}_D)|\textbf{Y}_D,\textbf{X}_D,\textbf{w}']$ is a constant w.r.t.~$\textbf{w}$. From \eqref{e:sllh}, we have the following expectation maximization iterations:\\

\vspace*{6px}
\noindent\fbox{%
\begin{minipage}{\textwidth}
\begin{itemize}[leftmargin=10pt]
  \item E-step: Compute instance posterior probabilities $p(y_{bi}=c|\textbf{Y}_b,\textbf{X}_b,\textbf{w}^{(k)})$ for $b=1,\dots,B$ and $i=1,\dots,n_b$.
  \item M-step: Find $\textbf{w}^{(k+1)}$ such that $g(\textbf{w}^{(k+1)},\textbf{w}^{(k)}) \ge g(\textbf{w}^{(k)},\textbf{w}^{(k)})$, for $g(\cdot,\cdot)$ in \eqref{e:sllh}.
\end{itemize}
\end{minipage}}
\vspace*{12px}\\
Note that the form of the EM surrogate function in \eqref{e:sllh} is similar to the SISL log-likelihood in \eqref{e:sisllearning}. While the EM approach for MIML requires updating $p(y_{bi}=c|\textbf{Y}_b,\textbf{X}_b,\textbf{w}')$, in the SISL case the equivalent term $I(y_{bi}=c)$ requires no update.

\subsubsection{The expectation step and our challenge}
\label{s:esaoc}
Computing $p(y_{bi}=c|\textbf{Y}_b,\textbf{X}_b,\textbf{w})$ can be done using $p(y_{bi}=c,\textbf{Y}_b|\textbf{X}_b,\textbf{w})$ and the following conditional rule
\begin{equation}
p(y_{bi}=c|\textbf{Y}_b,\textbf{X}_b,\textbf{w})=\frac{p(y_{bi}=c,\textbf{Y}_b|\textbf{X}_b,\textbf{w})}{\sum_{c=1}^C{p(y_{bi}=c,\textbf{Y}_b|\textbf{X}_b,\textbf{w})}}.
\end{equation}
The probability $p(y_{bi}=c,\textbf{Y}_b|\textbf{X}_b,\textbf{w})$ can be computed in a brute-force manner by keeping $y_{bi}=c$ and marginalizing over all other instances $y_{bj}$, where $j\in\{1,\dots,n_b\}$ and $j\neq i$, as follows
\begin{align}
p(y_{bi}=c,\textbf{Y}_b|\textbf{X}_b,\textbf{w})=&\sum_{y_{b1}=1}^C\dots\sum_{y_{b(i-1)}=1}^C\sum_{y_{b(i+1)}=1}^C\dots\sum_{y_{bn_b}=1}^Cp(y_{b1},\dots,y_{bi}=c,\dots,y_{bn_{b}},\textbf{Y}_b|\textbf{X}_b,\textbf{w})\nonumber\\
=&\sum_{y_{b1}=1}^C\dots\sum_{y_{b(i-1)}=1}^C\sum_{y_{b(i+1)}=1}^C\dots\sum_{y_{bn_b}=1}^C\Big[p(y_{b1},\dots,y_{bi}=c,\dots,y_{bn_{b}}|\textbf{X}_b,\textbf{w})\nonumber\\
&\hspace{56mm}\times p(\textbf{Y}_b|y_{b1},\dots,y_{bi}=c,\dots,y_{bn_{b}})\Big]\nonumber\\
=&\sum_{y_{b1}=1}^C\dots\sum_{y_{b(i-1)}=1}^C\sum_{y_{b(i+1)}=1}^C\dots\sum_{y_{bn_b}=1}^C\Big[p(y_{b1},\dots,y_{bi}=c,\dots,y_{bn_{b}}|\textbf{X}_b,\textbf{w})\nonumber\\
&\hspace{56mm}\times I(\textbf{Y}_b=\{c\}\cup\bigcup_{\substack{j=1;j\neq i}}^{n_b}y_{bj})\Big],\hspace{-5mm}\label{e:2}
\end{align}
where the last step is obtained by substituting $p(\textbf{Y}_b|y_{b1},\dots,y_{bi}=c,\dots,y_{bn_{b}})$ from \eqref{e:priornew}. From the last step in \eqref{e:2} and using the definition of the indicator function $I(\cdot)$, if $c\notin \textbf{Y}_b$ then $p(y_{bi}=c,\textbf{Y}_b|\textbf{X}_b,\textbf{w})=0$. Additionally, if there exists $j\in\{1,\dots,n_b\}$ and $j\neq i$ such that $y_{bj}\notin\textbf{Y}_b$ then $I(\textbf{Y}_b=\{c\}\cup\bigcup_{\substack{j=1;j\neq i}}^{n_b}y_{bj})=0$. As a result, we restrict the summation in the last step in \eqref{e:2} from $y_{bj}\in\{1,\dots,C\}$ to $y_{bj}\in\textbf{Y}_b$, for $j\in\{1,\dots,n_b\}$ and $j\neq i$. Moreover, from \eqref{e:2} and using the instance labels independence assumption, we obtain
\begin{eqnarray}
p(y_{bi}=c,\textbf{Y}_b|\textbf{X}_b,\textbf{w})=&\sum_{y_{b1}\in\textbf{Y}_b}\dots\sum_{y_{b(i-1)}\in\textbf{Y}_b}\sum_{y_{b(i+1)}\in\textbf{Y}_b}\dots\sum_{y_{bn_b}\in\textbf{Y}_b}\Big[p(y_{bi}=c|\textbf{x}_{bi},\textbf{w})\times\nonumber\\
&\prod_{\substack{j=1;j\neq i}}^{n_b}p(y_{bj}|\textbf{x}_{bj},\textbf{w})I(\textbf{Y}_b=\{c\}\cup\bigcup_{\substack{j=1;j\neq i}}^{n_b}y_{bj})\Big].\label{e:1}
\end{eqnarray}
The cost of computing prior probabilities $p(y_{bj}=l|\textbf{x}_{bj},\textbf{w})$ for $j=1,\dots,n_b$ and $l\in\textbf{Y}_b$ using \eqref{e:prior} is $O(n_bCd)$.  The cost of marginalizing using \eqref{e:1} is $O(|\textbf{Y}_b|^{n_b-1})$. Consequently, the computational complexity for the E-step is exponential w.r.t.~the number of instances in each bag. To overcome this challenge, we propose a computational method based on dynamic programming to compute $p(y_{bi}=c,\textbf{Y}_b|\textbf{X}_b,\textbf{w})$ from the prior probabilities $p(y_{bj}=l|\textbf{x}_{bj},\textbf{w})$. For many probability models, the computational complexity cost of exact calculation of the posterior probability necessitates approximate inference. In the following section, we introduce a dynamic programming approach for efficient and exact calculation of \eqref{e:1}.

\subsubsection{Forward algorithm for the E-step}
\begin{figure}[H]
\center
\includegraphics[width=5.5in]{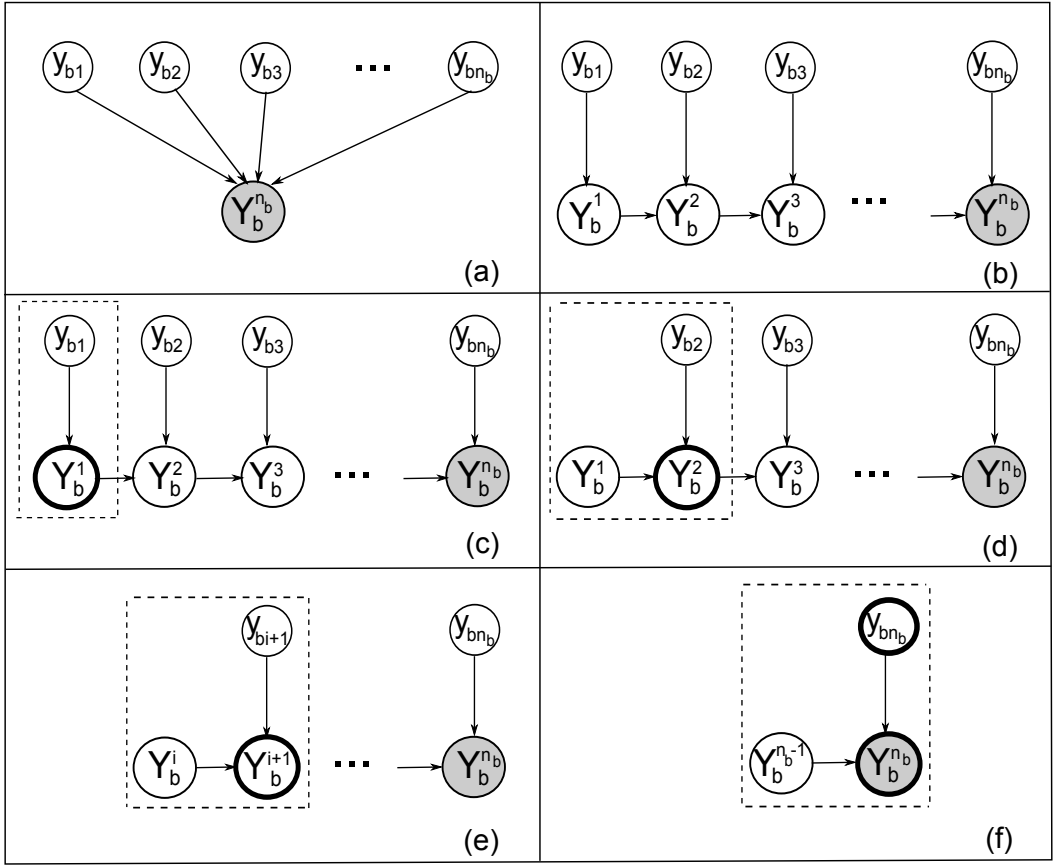}
\caption{A high level idea of our solution. Note that nodes that are currently computed on are bolded. (a) Our original model. (b) The model with new variables. (c) Compute $p(\textbf{Y}_b^{1}|\textbf{X}_b,\textbf{w})$. (d) Compute $p(\textbf{Y}_b^{2}|\textbf{X}_b,\textbf{w})$. (e) Compute $p(\textbf{Y}_b^{i+1}|\textbf{X}_b,\textbf{w})$ with $2 \le i \le n_b-2$ using equation \eqref{e:6}. (f) Compute $p(y_{bn_b}=c,\textbf{Y}_b|\textbf{X}_b,\textbf{w})$ using equation \eqref{e:last}.}
\label{fig:Dynamic4}
\end{figure}

In this section, we solve the aforementioned E-step challenge using a dynamic programming approach. To compute $p(y_{bi}=c,\textbf{Y}_b|\textbf{X}_b,\textbf{w})$ for all $c\in\textbf{Y}_b$ in \eqref{e:1}, we introduce $n_b$ sets $\textbf{Y}_b^1$, $\textbf{Y}_b^2$, $\dots$, $\textbf{Y}_b^{n_b}$ such that $\textbf{Y}_b^i=\bigcup_{j=1}^{i}y_{bj}$. Note that $\textbf{Y}_b^i$ represents the sub-bag label associated with the first $i$ instances in bag $b$. Consequently, $\textbf{Y}_b^{n_b}=\textbf{Y}_b$ can be computed recursively using $\textbf{Y}_b^i=\textbf{Y}_b^{i-1}\bigcup y_{bi}$. We illustrate the relation between these sets using Figure \ref{fig:Dynamic4}(b). By introducing sets $\textbf{Y}_b^1, \textbf{Y}_b^2, \dots, \textbf{Y}_b^{n_b}$ we can convert the graphical model in Figure \ref{fig:Dynamic4}(a) to a chain model in Figure \ref{fig:Dynamic4}(b). The chain structure allows us to perform a step by step grouping of independent factors in \eqref{e:1} resulting in a polynomial time complexity. Specifically, from the chain structure, we can dynamically compute $p(\textbf{Y}_b^{1}|\textbf{X}_b,\textbf{w})$, $p(\textbf{Y}_b^{2}|\textbf{X}_b,\textbf{w})$, $\dots$, $p(\textbf{Y}_b^{n_b-1}|\textbf{X}_b,\textbf{w})$ as illustrated in Figures \ref{fig:Dynamic4}(c), (d), and (e), respectively. Finally, the desired probability $p(y_{bn_b}=c,\textbf{Y}_b|\textbf{X}_b,\textbf{w})$ can be computed from $p(\textbf{Y}_b^{n_b-1}|\textbf{X}_b,\textbf{w})$ and $p(y_{n_b}=c|\textbf{x}_{bn},\textbf{w})$ as in Figure \ref{fig:Dynamic4}(f). Our forward algorithm is summarized by the following steps.

\noindent \underline{\textbf{Step 1.}} To compute $p(\textbf{Y}_b^{n_b-1}|\textbf{X}_b,\textbf{w})$, we use an incremental calculation of $p(\textbf{Y}_b^{i+1}|\textbf{X}_b,\textbf{w})$ from $p(\textbf{Y}_b^{i}|\textbf{X}_b,\textbf{w})$ using the following proposition. This process is also illustrated in Figures \ref{fig:Dynamic4}(c) to (e).

\begin{proposition} {\em The probability} $p(\textbf{Y}_b^{i+1}=\textbf{\L}|\textbf{X}_b,\textbf{w})$, {\em for any} $\textbf{\L}$ {\em in the power set of} $\textbf{Y}_b$ {\em excluding the empty set, can be computed using the following recursion}
\label{p:1}
\begin{itemize}[leftmargin=10pt]
\item {\em For $i=0$:}
\begin{equation}
\label{e:initdp}
\text{{\em If} }\textbf{\L}=\{l\}\text{, }p(\textbf{Y}_b^{1}=\textbf{\L}|\textbf{X}_b,\textbf{w})=p(y_{b1}=l|\textbf{x}_{b1},\textbf{w})\text{{\em, otherwise} } p(\textbf{Y}_b^{1}=\textbf{\L}|\textbf{X}_b,\textbf{w})=0.
\end{equation}
\item {\em For $i\ge 1$:}
\begin{align}
p(\textbf{Y}_b^{i+1}=\textbf{\L}|\textbf{X}_b,\textbf{w})=\sum_{c\in\textbf{\L}}&p(y_{b(i+1)}=c|\textbf{x}_{b(i+1)},\textbf{w})\times\label{e:6}\\
&[p(\textbf{Y}_b^{i}=\textbf{\L}|\textbf{X}_b,\textbf{w})+p(\textbf{Y}_b^{i}=\textbf{\L}_{\backslash c}|\textbf{X}_b,\textbf{w})],\nonumber
\end{align}
\end{itemize}
{\em where we use} $\textbf{\L}_{\backslash c}$ {\em to denote the set} $\{l|l\in\textbf{\L}, l\neq c\},$ {\em for any} $\textbf{\L}$ {\em and} $c\in\textbf{\L}$.
\end{proposition}

\noindent For a detailed proof, we refer the reader to Appendix A.

\begin{figure}[H]
\center
\includegraphics[width=4.8in]{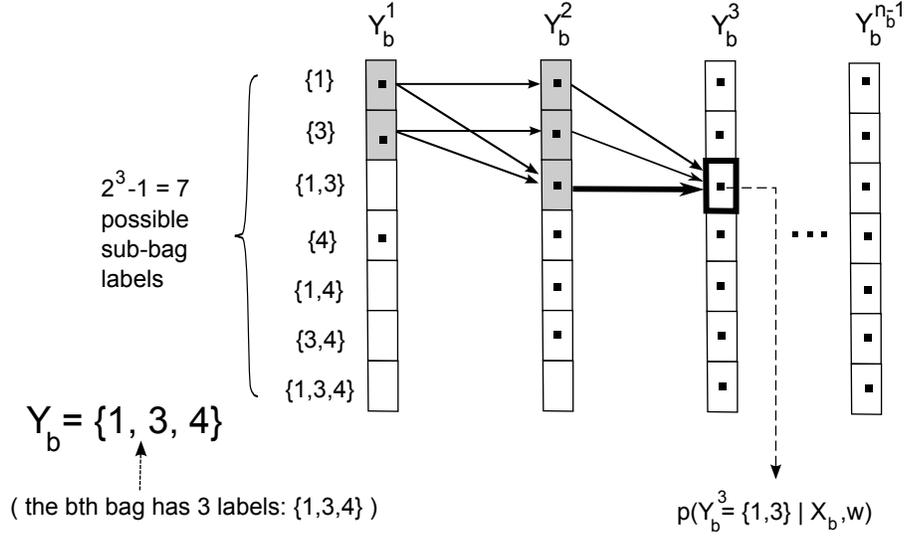}
\caption{Example for dynamically computing $p(\textbf{Y}_b^{n_b-1}|\textbf{X}_b,\textbf{w})$ in a bag having three labels. The bolded box representing $p(\textbf{Y}_b^{3}=\{1, 3\}|\textbf{X}_b,\textbf{w})$ contains the probability which is currently computed. The continuous arrows mean $\textit{`is computed from'}$. For example, $p(\textbf{Y}_b^{3}=\{1, 3\}|\textbf{X}_b,\textbf{w})$ $\textit{is computed from}$ $p(\textbf{Y}_b^{2}=\{1\}|\textbf{X}_b,\textbf{w})$, $p(\textbf{Y}_b^{2}=\{3\}|\textbf{X}_b,\textbf{w})$, and $p(\textbf{Y}_b^{2}=\{1,3\}|\textbf{X}_b,\textbf{w})$ using \eqref{e:6} with $i=2$ and $\textbf{\L}=\{1,3\}$ as in \eqref{e:example1}. Thick arrows denote multiple coefficients. For example, $p(\textbf{Y}_b^{2}=\{1,3\}|\textbf{X}_b,\textbf{w})$ connects to $p(\textbf{Y}_b^{3}=\{1, 3\}|\textbf{X}_b,\textbf{w})$ by a thick arrow since there are two coefficients for $p(\textbf{Y}_b^{2}=\{1,3\}|\textbf{X}_b,\textbf{w})$ in \eqref{e:example1} which are $p(y_{b3}=1|\textbf{x}_{b3},\textbf{w})$ and $p(y_{b3}=3|\textbf{x}_{b3},\textbf{w})$. In contrast, there is only one coefficient for each of $p(\textbf{Y}_b^{2}=\{1\}|\textbf{X}_b,\textbf{w})$ and $p(\textbf{Y}_b^{2}=\{3\}|\textbf{X}_b,\textbf{w})$ in \eqref{e:example1} indicated by thin arrows. Shaded boxes represent the probabilities computed to obtain $p(\textbf{Y}_b^{3}=\{1, 3\}|\textbf{X}_b,\textbf{w})$. Dotted boxes represent nonzero probabilities. In contrast, non-dotted boxes indicate zero probabilities. For instance, $p(\textbf{Y}_b^{1}=\{1,3\}|\textbf{X}_b,\textbf{w})=0$  since $\textbf{Y}_b^{1}$ can only contain one instance label.}
\label{fig:Dynamic}
\end{figure}

To understand the recursion in Proposition 1, we present an example of a bag with bag label $\textbf{Y}_b=\{1, 3, 4\}$ in Figure \ref{fig:Dynamic}. The process of computing $p(\textbf{Y}_b^{3}=\{1,3\}|\textbf{X}_b,\textbf{w})$ is as follows:\\

\noindent $\bullet$ First, $p(\textbf{Y}_b^{1}=\{1\}|\textbf{X}_b,\textbf{w})$ and $p(\textbf{Y}_b^{1}=\{3\}|\textbf{X}_b,\textbf{w})$ are computed by \eqref{e:initdp} as

\begin{align}
p(\textbf{Y}_b^{1}&=\{1\}|\textbf{X}_b,\textbf{w})=p(y_{b1}=1|\textbf{x}_{b1},\textbf{w}),\nonumber\\
p(\textbf{Y}_b^{1}&=\{3\}|\textbf{X}_b,\textbf{w})=p(y_{b1}=3|\textbf{x}_{b1},\textbf{w}).\label{e:example3}
\end{align}

\noindent Note that since $\textbf{Y}_b^{1}$ only contains the label of the first instance in the bag, it cannot have more than one label and therefore $p(\textbf{Y}_b^{1}=\{1,3\}|\textbf{X}_b,\textbf{w})=0$.\\

\noindent $\bullet$ Second, using \eqref{e:example3}, $p(\textbf{Y}_b^{2}=\{1\}|\textbf{X}_b,\textbf{w})$, $p(\textbf{Y}_b^{2}=\{3\}|\textbf{X}_b,\textbf{w})$, and $p(\textbf{Y}_b^{2}=\{1,3\}|\textbf{X}_b,\textbf{w})$ are computed from $p(\textbf{Y}_b^{1}=\{1\}|\textbf{X}_b,\textbf{w})$ and $p(\textbf{Y}_b^{1}=\{3\}|\textbf{X}_b,\textbf{w})$ as

\begin{alignat}{2}
&p(\textbf{Y}_b^{2}=\{1\}|\textbf{X}_b,\textbf{w})&=&p(\textbf{Y}_b^{1}=\{1\}|\textbf{X}_b,\textbf{w})\cdot p(y_{b2}=1|\textbf{x}_{b2},\textbf{w}),\nonumber\\
&p(\textbf{Y}_b^{2}=\{3\}|\textbf{X}_b,\textbf{w})&=&p(\textbf{Y}_b^{1}=\{3\}|\textbf{X}_b,\textbf{w})\cdot p(y_{b2}=3|\textbf{x}_{b2},\textbf{w}),\nonumber\\
&p(\textbf{Y}_b^{2}=\{1,3\}|\textbf{X}_b,\textbf{w})&=&p(\textbf{Y}_b^{1}=\{1\}|\textbf{X}_b,\textbf{w})\cdot p(y_{b2}=3|\textbf{x}_{b2},\textbf{w})\nonumber\\
&&&+p(\textbf{Y}_b^{1}=\{3\}|\textbf{X}_b,\textbf{w})\cdot p(y_{b2}=1|\textbf{x}_{b2},\textbf{w}).\label{e:example2}
\end{alignat}

\noindent $\bullet$ Finally, $p(\textbf{Y}_b^{3}=\{1,3\}|\textbf{X}_b,\textbf{w})$ is obtained by replacing the probabilities obtained in \eqref{e:example2} into \eqref{e:6} with $i=2$ and $\textbf{\L}=\{1, 3\}$ as follows
\begin{align}
p(\textbf{Y}_b^{3}=\{1,3\}|\textbf{X}_b,\textbf{w})=&p(\textbf{Y}_b^{2}=\{1\}|\textbf{X}_b,\textbf{w})\cdot p(y_{b3}=3|\textbf{x}_{b3},\textbf{w})\label{e:example1}\\
&+p(\textbf{Y}_b^{2}=\{3\}|\textbf{X}_b,\textbf{w})\cdot p(y_{b3}=1|\textbf{x}_{b3},\textbf{w})\nonumber\\
&+p(\textbf{Y}_b^{2}=\{1,3\}|\textbf{X}_b,\textbf{w})\cdot [p(y_{b3}=1|\textbf{x}_{b3},\textbf{w})+p(y_{b3}=3|\textbf{x}_{b3},\textbf{w})].\nonumber
\end{align}

\noindent Note that in Figure \ref{fig:Dynamic}, we only compute the probabilities stored in the boxes of the $i$th column, which are $p(\textbf{Y}_b^{i}=\textbf{\L}|\textbf{X}_b,\textbf{w})$ for \textbf{\L} in the power set of \{1, 3, 4\} (excluding the empty set), one time. The probabilities stored in the $(i+1)$th column depend only on the probabilities stored in the $i$th column, as illustrated in \crefrange{e:example3}{e:example1}.

The computational complexity of Step 1 can be derived as follows. Since $\textbf{Y}_b^{i+1} \subseteq \textbf{Y}_b$ and $\textbf{Y}_b^{i+1}$ contains at least one instance, there are at most $2^{|\textbf{Y}_b|}-1$ possible bag labels $\textbf{\L}$ for $\textbf{Y}_b^{i+1}$ in \eqref{e:6}. Moreover, computing the probability of each bag label $\textbf{Y}_b^{i+1}$ using \eqref{e:6} involves the computation of at most $2\times|\textbf{Y}_b|$ terms in the sum. As a result, the computational complexity of computing $p(\textbf{Y}_b^{i+1}|\textbf{X}_b,\textbf{w})$ from $p(\textbf{Y}_b^{i}|\textbf{X}_b,\textbf{w})$ is $O(|\textbf{Y}_b|2^{|\textbf{Y}_b|})$. To obtain $p(\textbf{Y}_b^{n_b-1}|\textbf{X}_b,\textbf{w})$, the aforementioned recursion is applied for $i=0,\dots,n_b-2$ and hence the computational complexity is $O(|\textbf{Y}_b|2^{|\textbf{Y}_b|}n_b)$.\\

\noindent \underline{\textbf{Step 2.}} Next, we compute $p(y_{bn_b}=c,\textbf{Y}_b|\textbf{X}_b,\textbf{w})$ for all $c\in\textbf{Y}_b$ from $p(\textbf{Y}_b^{n_b-1}|\textbf{X}_b,\textbf{w})$ and $p(y_{bn_b}=c|\textbf{x}_{bn_b},\textbf{w})$ using the following proposition.

\begin{proposition} {\em The probability} $p(y_{bn_b}=c,\textbf{Y}_b=\textbf{\L}|\textbf{X}_b,\textbf{w})$ {\em for all} $c\in\textbf{\L}$ {\em can be computed using}
\label{p:2}
\begin{align}
p(y_{bn_b}=c,\textbf{Y}_b=\textbf{\L}|\textbf{X}_b,\textbf{w})=&p(y_{bn_b}=c|\textbf{x}_{bn_b},\textbf{w})\nonumber\times\\
&[p(\textbf{Y}_b^{n_b-1}=\textbf{\L}|\textbf{X}_b,\textbf{w})+p(\textbf{Y}_b^{n_b-1}=\textbf{\L}_{\backslash c}|\textbf{X}_b,\textbf{w})] \label{e:last}.
\end{align}
\end{proposition}

\noindent For a detailed proof, we refer the reader to Appendix B.

The computational complexity required to obtain $p(y_{bn_b}=c,\textbf{Y}_b|\textbf{X}_b,\textbf{w})$ for all $c\in\textbf{Y}_b$ using Proposition 2 is $O(|\textbf{Y}_b|)$. Combining the computational complexity in Proposition \ref{p:1} and \ref{p:2}, the overall computational complexity for computing $p(y_{bn_b}=c,\textbf{Y}_b|\textbf{X}_b,\textbf{w})$ is $O(|\textbf{Y}_b|2^{|\textbf{Y}_b|}n_b)$.\\

\begin{figure}[H]
\center
\includegraphics[width=6in]{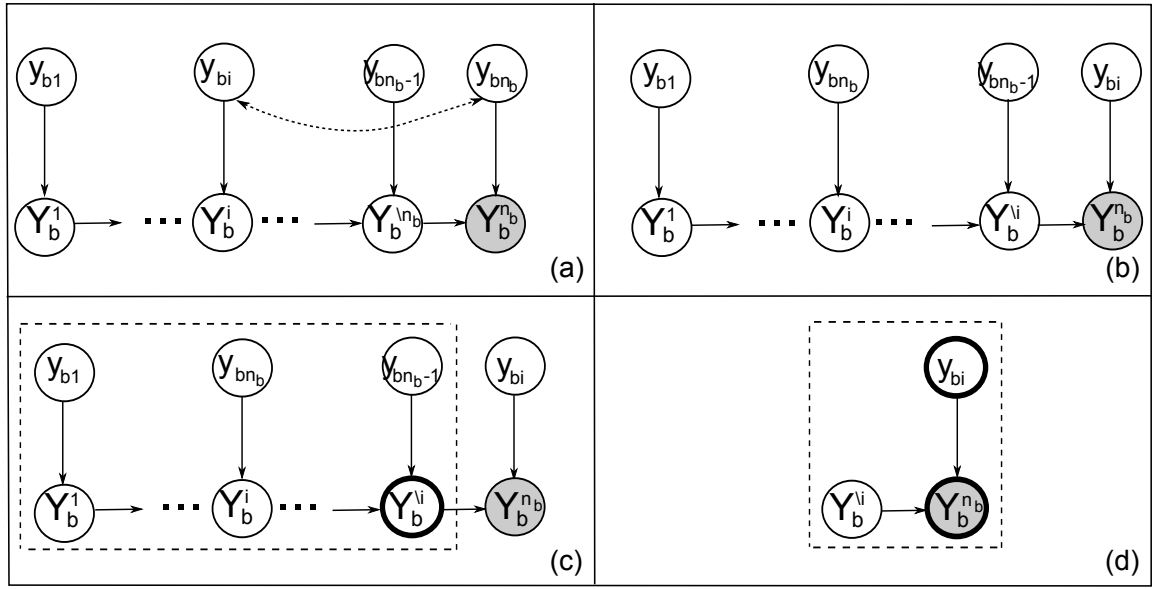}
\caption{Example of computing $p(y_{bi}=c,\textbf{Y}_b|\textbf{X}_b,\textbf{w})$. Nodes that are currently computed on are bolded. (a) The original order of instances. (b) The structure after swapping the $i$th and $n_b$th instances. (c) Compute $p(\textbf{Y}_b^{\backslash i}|\textbf{X}_b,\textbf{w})$ using Proposition \ref{p:1}. (d) Compute $p(y_{bi}=c,\textbf{Y}_b|\textbf{X}_b,\textbf{w})$ using Proposition \ref{p:2}. Note that for $i=n_b$, the process from (b) to (d) is the same as the process described from Figures \ref{fig:Dynamic4}(b) to (f). }
\label{fig:Dynamic6}
\end{figure}

\noindent \underline{\textbf{Step 3.}} Finally, we compute $p(y_{bi}=c,\textbf{Y}_b|\textbf{X}_b,\textbf{w}),$ $\forall i \neq n_b$. Note that $p(y_{bi}=c,\textbf{Y}_b|\textbf{X}_b,\textbf{w})$ is independent of the position of the $i$th instance in the $b$th bag. Define $\textbf{Y}_b^{\backslash i}=\bigcup_{\substack{j=1;j\neq i}}^{n_b}y_{bj}$ the union of the instance label of the $b$th bag excluding label $y_{bi}$. To compute $p(y_{bi}=c,\textbf{Y}_b|\textbf{X}_b,\textbf{w})$, we swap the $i$th instance with the $n_b$th instance. This process is depicted in Figures \ref{fig:Dynamic6}(a) and (b). We then compute $p(\textbf{Y}_b^{\backslash i}|\textbf{X}_b,\textbf{w})$ by applying Proposition \ref{p:1} on the newly ordered instances, as illustrated in Figure \ref{fig:Dynamic6}(c). Finally, the probability $p(y_{bi}=c,\textbf{Y}_b|\textbf{X}_b,\textbf{w})$ is computed from $p(\textbf{Y}_b^{\backslash i}|\textbf{X}_b,\textbf{w})$ and $p(y_{bi}=c|\textbf{x}_{bi},\textbf{w})$ using Proposition \ref{p:2}, which is shown in Figure \ref{fig:Dynamic6}(d).

The forward algorithm is presented in Algorithm \ref{a:1}. For each $i$, the computational complexity for computing $p(y_{bi}=c,\textbf{Y}_b|\textbf{X}_b,\textbf{w})$ is $O(|\textbf{Y}_b|2^{|\textbf{Y}_b|}n_b)$. As a result, the computational complexity for computing $p(y_{bi}=c,\textbf{Y}_b|\textbf{X}_b,\textbf{w})$, for $1 \le i \le n_b$, is $O(|\textbf{Y}_b|2^{|\textbf{Y}_b|}n_b^2)$. The limitation of the forward approach is that in order to compute $p(y_{bi}=c,\textbf{Y}_b|\textbf{X}_b,\textbf{w})$, we swap the $i$th and $n_b$th instances, and repeat the Step 1. Recomputing the Step 1 $n_b$ times for each bag is inefficient. That leads us to the forward and substitution algorithm in the following section.

\renewcommand{\algorithmicrequire}{\textbf{Input:}}
\renewcommand{\algorithmicensure}{\textbf{Output:}}
\newcommand{\algrule}[1][.2pt]{\par\vskip.5\baselineskip\hrule height #1\par\vskip.5\baselineskip}

\begin{algorithm}[H]
\caption{Forward algorithm to compute $p(y_{bi}=c,\textbf{Y}_b=\textbf{\L}|\textbf{X}_b,\textbf{w})$, $\forall i$ s.t.~$1\le i \le n_b$}
\begin{algorithmic}
\REQUIRE $\textbf{\L},\textbf{X}_b,\textbf{Y}_b,\textbf{w},c$
\\\algrule
\FOR{$i=1$ to $n_b$}
\STATE $\textit{Step 1:}$
\STATE Swap \{$\textbf{x}_{bi},y_{bi}$\} and \{$\textbf{x}_{bn_b},y_{bn_b}$\};
\STATE Initialize $p(\textbf{Y}_b^{1}=\textbf{\l}|\textbf{X}_b,\textbf{w})=0,$ $\forall \textbf{\l}$ in the power set of $\textbf{\L}$ excluding the empty set;
\FOR{$l$ $\in$ $\textbf{\L}$}
\STATE $p(\textbf{Y}_b^{1}=\{l\}|\textbf{X}_b,\textbf{w})=p(y_{b1}=l|\textbf{x}_{b1},\textbf{w})$;
\ENDFOR
\FOR{$k=1$ to $n_b-2$}
\FOR{$u=1$ to $2^{|\textbf{\L}|}-1$}
\STATE $p(\textbf{Y}_b^{k+1}=\textbf{\L}^u|\textbf{X}_b,\textbf{w})=\sum_{l\in \textbf{\L}^u}p(y_{b(k+1)}=l|\textbf{x}_{b(k+1)},\textbf{w})[p(\textbf{Y}_b^{k}=\textbf{\L}^u|\textbf{X}_b,\textbf{w})+p(\textbf{Y}_b^{k}=\textbf{\L}_{\backslash l}^u|\textbf{X}_b,\textbf{w})]$;
\ENDFOR
\ENDFOR
\STATE $\textit{Step 2:}$
\STATE Return $p(y_{bi}=c,\textbf{Y}_b=\textbf{\L}|\textbf{X}_b,\textbf{w})=p(y_{bn_b}=c|\textbf{x}_{bn_b},\textbf{w})[p(\textbf{Y}_b^{n_b-1}=\textbf{\L}|\textbf{X}_b,\textbf{w})+p(\textbf{Y}_b^{n_b-1}=\textbf{\L}_{\backslash c}|\textbf{X}_b,\textbf{w})]$;
\STATE Swap \{$\textbf{x}_{bi},y_{bi}$\} and \{$\textbf{x}_{bn_b},y_{bn_b}$\};
\ENDFOR
\end{algorithmic}
\label{a:1}
\end{algorithm}
\vspace{30mm}

\subsubsection{$\textbf{F}$orward $\textbf{A}$nd $\textbf{S}$ubs$\textbf{T}$itution (FAST) algorithm for the E-step}
\label{ss:farfe}

\begin{figure}[H]
\center
\includegraphics[width=6in]{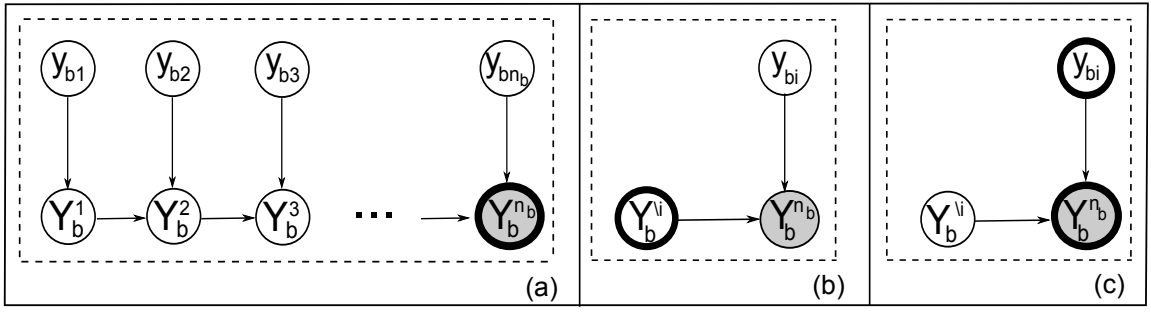}
\caption{The forward and substitution algorithm. Nodes that are currently computed on are bolded. (a) Compute $p(\textbf{Y}_b|\textbf{X}_b,\textbf{w})$ using Proposition \ref{p:1}. (b) Efficiently compute $p(\textbf{Y}_b^{\backslash i}|\textbf{X}_b,\textbf{w})$ from $p(\textbf{Y}_b|\textbf{X}_b,\textbf{w})$ using Proposition \ref{p:4}, $\forall 1 \le i \le n_b$. (c) Compute $p(y_{bi}=c,\textbf{Y}_b|\textbf{X}_b,\textbf{w})$ using Proposition \ref{p:2}, $\forall 1 \le i \le n_b$. }
\label{fig:Dynamic5}
\end{figure}

Recall in the forward algorithm, to compute $p(y_{bi}=c,\textbf{Y}_b|\textbf{X}_b,\textbf{w})$, for $i=1, 2,\dots,n_b-1$, we swap the $i$th and $n_b$th instances, and sequentially recompute $p(\textbf{Y}_b^{\backslash i}|\textbf{X}_b,\textbf{w})$. The computational complexity for the overall process is $O(|\textbf{Y}_b|2^{|\textbf{Y}_b|}n_b^2)$. In this section, we propose an efficient method for computing $p(y_{bi}=c,\textbf{Y}_b|\textbf{X}_b,\textbf{w})$ for $i=1, 2,\dots,n_b$. The computational complexity of this method is only $O(|\textbf{Y}_b|2^{|\textbf{Y}_b|}n_b)$. A conceptual illustration of the forward and substitution algorithm is presented in Figure \ref{fig:Dynamic5}. The overview of the forward and substitution algorithm is presented as follows.\\

\noindent \underline{$\textbf{Step 1.}$} Compute $p(\textbf{Y}_b^{n_b}|\textbf{X}_b,\textbf{w})$ using Proposition \ref{p:1}. From Proposition \ref{p:1}, the computational complexity for Step 1 is $O(|\textbf{Y}_b|2^{|\textbf{Y}_b|}n_b)$.\\
\vspace{3mm}\\
\noindent \underline{$\textbf{Step 2.}$} For each $i \in\{1, 2,\dots,n_b\}$, efficiently compute $p(\textbf{Y}_b^{\backslash i}|\textbf{X}_b,\textbf{w})$ from $p(\textbf{Y}_b^{n_b}|\textbf{X}_b,\textbf{w})$ using the following proposition.

\begin{proposition}
\label{p:4}
{\em Define} \{${\textbf{\L}^1, \dots, \textbf{\L}^{2^{|\textbf{Y}_b|}-1}}$\} {\em as the power set of} $\textbf{Y}_b$ {\em excluding the empty set. Define vectors} $\textbf{u}, \textbf{v} \in \mathbb{R}^{(2^{|\textbf{Y}_b|}-1)\times 1}$ {\em such that} $\textbf{u}(m)=p(\textbf{Y}_b=\textbf{\L}^m|\textbf{X}_b,\textbf{w})$ {\em and} $\textbf{v}(m)=p(\textbf{Y}_b^{\backslash i}=\textbf{\L}^m|\textbf{X}_b,\textbf{w})$, {\em for} $1 \le m \le 2^{|\textbf{Y}_b|}-1$. {\em Define a matrix} $\textbf{A}\in\mathbb{R}^{(2^{|\textbf{Y}_b|}-1)\times (2^{|\textbf{Y}_b|}-1)}$ {\em such that}
\begin{subnumcases}{\textbf{A}(r,s)=}
p(y_{bi}=c|\textbf{x}_{bi},\textbf{w}) &$\text{{\em if} } \textbf{\L}^s \subset \textbf{\L}^r \text{{\em and} } \textbf{\L}^r \backslash \textbf{\L}^s =\{c\}$\label{case1}\\
\sum_{l\in\textbf{\L}^r}p(y_{bi}=l|\textbf{x}_{bi},\textbf{w}) &\text{{\em if} $\textbf{\L}^s=\textbf{\L}^r$}\label{case2}\\
0 &\text{{\em otherwise, }}\label{case3}
\end{subnumcases}
{\em where} $1 \le r,s \le 2^{|\textbf{Y}_b|}-1$. {\em Then,} $p(\textbf{Y}_b^{\backslash i}|\textbf{X}_b,\textbf{w})$ {\em can be computed from} $p(\textbf{Y}_b|\textbf{X}_b,\textbf{w})$ {\em by solving} $\textbf{u}=\textbf{A}\textbf{v}$.

\end{proposition}
\noindent For a proof, we refer the reader to Appendix C.

\begin{figure}[H]
\center
\includegraphics[width=5.4in]{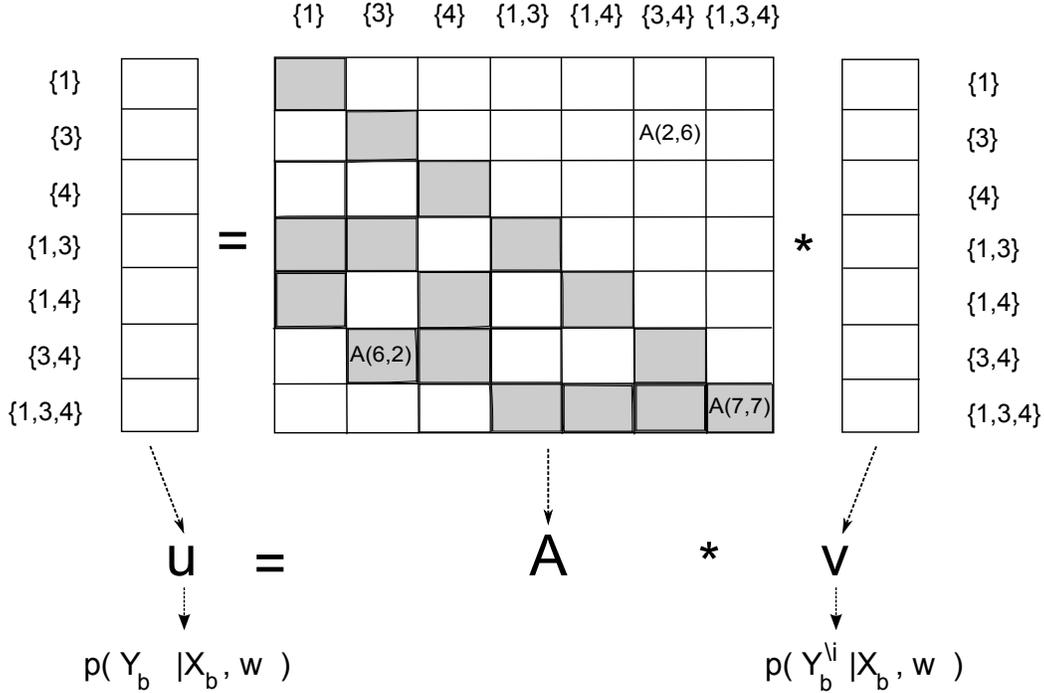}
\caption{Example of the lower triangular matrix \textbf{A} for a bag with three labels as in Figure \ref{fig:Dynamic}. Nonzero elements in \textbf{A} are shaded. Sets in the power set of \{1,3,4\}, excluding the empty set, are sorted in an ascending order of their cardinality. If two sets are equal in cardinality, they are sorted based on the lexicographical order. Consequently, the sets are sorted as \{1\}, \{3\}, \{4\}, \{1,3\}, \{1,4\}, \{3,4\}, \{1,3,4\} which are denoted by $\textbf{\L}^1, \textbf{\L}^2, \textbf{\L}^3, \textbf{\L}^4, \textbf{\L}^5, \textbf{\L}^6, \textbf{\L}^7$, respectively. For this order, \textbf{A} is constructed as in Proposition \ref{p:4}. For example, from \eqref{case1}, $\textbf{A}(6,2)=p(y_{bi}=4|\textbf{x}_{bi},\textbf{w})$ since $\textbf{\L}^2$=\{3\} and $\textbf{\L}^6$=\{3,4\} leading to $\textbf{\L}^2 \subset \textbf{\L}^6$ and $\textbf{\L}^6 \backslash \textbf{\L}^2$=\{4\}. Moreover, from \eqref{case2}, $\textbf{A}(7,7)=\sum_{l\in\{1,3,4\}}p(y_{bi}=l|\textbf{x}_{bi},\textbf{w})$ since $\textbf{\L}^7=\{1,3,4\}$. Additionally, from \eqref{case3}, $\textbf{A}(2,6)=0$ since $\textbf{\L}^6 \not\subseteq \textbf{\L}^2$.}
\label{fig:Dynamic7}
\end{figure}

We assume that $\textbf{\L}^1$, $\textbf{\L}^2$, $\dots$, $\textbf{\L}^{2^{|\textbf{Y}_b|}-1}$ are sorted in an ascending order of their cardinality. Moreover, if two sets are equal in cardinality, they are sorted based on the lexicographical order. For this order, the matrix $\textbf{A}$ constructed from Proposition \ref{p:4} is a lower triangular matrix. We prove this by contradiction. Assume that $\textbf{A}$ is not a lower triangular matrix then there exist $(r,s)$ such that $r<s$ and $\textbf{A}(r,s)\neq 0$. From Proposition \ref{p:4}, if $\textbf{A}(r,s)\neq 0$ then $\textbf{\L}^s=\textbf{\L}^r$ or $\textbf{\L}^s \subset \textbf{\L}^r$ which means that $s\le r$ leading to a contradiction with $r<s$. The structure of $\textbf{A}$ for the example of Figure \ref{fig:Dynamic} is presented in Figure \ref{fig:Dynamic7}. Since $\textbf{A}$ is lower triangular, $\textbf{v}$ can be obtained from $\textbf{A}$ and $\textbf{u}$ using the forward substitution method \citep{golub2012matrix} with the time complexity proportional to the number of nonzero elements in $\textbf{A}$ which is $O(|\textbf{Y}_b|2^{|\textbf{Y}_b|})$. As a result, the computational complexity for Step 2 is $O(|\textbf{Y}_b|2^{|\textbf{Y}_b|}n_b)$. Computing $p(\textbf{Y}_b^{\backslash i}|\textbf{X}_b,\textbf{w})$ by Proposition \ref{p:4} in the forward and substitution algorithm is faster by an order of $O(n_b)$ compared to swapping the $i$th and $n_b$th instances then using Proposition \ref{p:1} as in the forward algorithm.\\

\noindent \underline{$\textbf{Step 3.}$} For each $i \in\{1, 2,\dots,n_b\}$, compute $p(y_{bi}=c,\textbf{Y}_b|\textbf{X}_b,\textbf{w})$ from $p(y_{bi}=c|\textbf{x}_{bi},\textbf{w})$ and $p(\textbf{Y}_b^{\backslash i}|\textbf{X}_b,\textbf{w})$, for all $c\in\textbf{Y}_b$, using Proposition \ref{p:2}. From Proposition \ref{p:2}, the computational complexity for Step 3 is $O(n_b|\textbf{Y}_b|)$.

\vspace{5mm}

Combining the computational complexities in Step 1, 2, and 3, the computational complexity of the forward and substitution algorithm is $O(|\textbf{Y}_b|2^{|\textbf{Y}_b|}n_b)$. A detailed description of the forward and substitution algorithm is presented in Algorithm \ref{a:3}. From Section \ref{s:esaoc}, the computational complexity for obtaining the prior probabilities is $O(n_bCd)$. Consequently, the computational complexity of the E-step implementation per bag using the forward and substitution algorithm is $O(|\textbf{Y}_b|2^{|\textbf{Y}_b|}n_b+n_bCd)$. As a result, the simple structure of our ORed-logistic regression model allows to perform an efficient and exact inference instead of using approximation techniques which may then degrade the accuracy.

\begin{algorithm}[!h]
\caption{$\textbf{F}$orward $\textbf{A}$nd $\textbf{S}$ubs$\textbf{T}$itution algorithm (FAST) to compute $p(y_{bi}=c,\textbf{Y}_b=\textbf{\L}|\textbf{X}_b,\textbf{w})$, $\forall i$ s.t.~$1\le i \le n_b$}
\begin{algorithmic}
\REQUIRE $\textbf{\L},\textbf{X}_b,\textbf{Y}_b,\textbf{w},c$
\\\algrule
\STATE $\textit{Step 1:}$
\STATE Initialize $p(\textbf{Y}_b^{1}=\textbf{\l}|\textbf{X}_b,\textbf{w})=0,$ $\forall \textbf{\l}$ in the power set of $\textbf{\L}$ excluding the empty set;
\FOR{$l \in \textbf{\L}$}
\STATE $p(\textbf{Y}_b^{1}=\{l\}|\textbf{X}_b,\textbf{w})=p(y_{b1}=l|\textbf{x}_{b1},\textbf{w})$;
\ENDFOR
\FOR{$i=1$ to $n_b-1$}
\FOR{$m=1$ to $2^{|\textbf{\L}|}-1$}
\STATE $p(\textbf{Y}_b^{i+1}=\textbf{\L}^m|\textbf{X}_b,\textbf{w})=\sum_{l\in \textbf{\L}^m}p(y_{b(i+1)}=l|\textbf{x}_{b(i+1)},\textbf{w})[p(\textbf{Y}_b^{i}=\textbf{\L}^m|\textbf{X}_b,\textbf{w})+p(\textbf{Y}_b^{i}=\textbf{\L}_{\backslash l}^m|\textbf{X}_b,\textbf{w})]$;
\ENDFOR
\ENDFOR
\STATE $\textit{Step 2:}$
\FOR{$i=1$ to $n_b$}
\STATE Construct the $\textbf{u}$ vector in Proposition \ref{p:4}, from $p(\textbf{Y}_b=\textbf{\L}^{m}|\textbf{X}_b,\textbf{w}), \forall 1 \le m \le 2^{|\textbf{\L}|}-1$;
\STATE Construct the $\textbf{A}$ matrix in Proposition \ref{p:4}, from $p(y_{bi}|\textbf{x}_{bi},\textbf{w})$;
\STATE Solve $\textbf{u}=\textbf{A}\textbf{v}$ in order to obtain $\textbf{v}$ containing $p(\textbf{Y}_b^{\backslash i}=\textbf{\L}^{m}|\textbf{X}_b,\textbf{w}), \forall 1 \le m \le 2^{|\textbf{\L}|}-1$;
\ENDFOR
\STATE $\textit{Step 3:}$
\FOR{$i=1$ to $n_b$}
\STATE Return $p(y_{bi}=c,\textbf{Y}_b=\textbf{\L}|\textbf{X}_b,\textbf{w})=p(y_{bi}=c|\textbf{x}_{bi},\textbf{w})[p(\textbf{Y}_b^{\backslash i}=\textbf{\L}|\textbf{X}_b,\textbf{w})+p(\textbf{Y}_b^{\backslash i}=\textbf{\L}_{\backslash c}|\textbf{X}_b,\textbf{w})]$;
\ENDFOR
\end{algorithmic}
\label{a:3}
\end{algorithm}

\subsubsection{Maximization step}
We use gradient ascent to increase the objective function in \eqref{e:sllh} as follows
\begin{equation}
\textbf{w}_c^{(k+1)}=\textbf{w}_c^{(k)}+ \frac{\partial g(\textbf{w},\textbf{w}^{(k)})}{\partial\textbf{w}_c}\bigg|_{\textbf{w}=\textbf{w}^{(k)}} \times \eta,
\label{e:12}
\end{equation}
where the first derivative of $g(\textbf{w},\textbf{w}^{(k)})$ w.r.t.~$\textbf{w}_c$ is computed as follows
\begin{align}
&\frac{\partial  g(\textbf{w},\textbf{w}^{(k)})}{\partial\textbf{w}_c}=\sum_{b=1}^B\sum_{i=1}^{n_b}[p(y_{bi}=c|\textbf{Y}_b,\textbf{X}_b,\textbf{w}^{(k)})\textbf{x}_{bi}-\frac{e^{\textbf{w}_c^T\textbf{x}_{bi}}\textbf{x}_{bi}}{\sum_{c=1}^C{e^{\textbf{w}_c^T\textbf{x}_{bi}}}}]\nonumber\\
&=\sum_{b=1}^B\sum_{i=1}^{n_b}[p(y_{bi}=c|\textbf{Y}_b,\textbf{X}_b,\textbf{w}^{(k)})-p(y_{bi}=c|\textbf{x}_{bi},\textbf{w})]\textbf{x}_{bi}\label{e:13},
\end{align}
and $\eta$ in \eqref{e:12} is determined using the backtracking line search method \citep{boyd2004convex}. In each backtracking step, the surrogate function in \eqref{e:sllh} is computed with the time complexity $O(\sum_{b=1}^{B}n_bCd)$. As a result, the computational complexity of the maximization step is $O(\sum_{b=1}^{B}n_bCd\overline{M})$, where $\overline{M}$ is the average number of backtracking steps.

\subsection{Kernel-based OR-ed logistic regression model}
\label{ss:kbolrm}
In order to deal with linearly inseparable data, we introduce the kernel extension to our model. Kernel learning is a method that transforms the data from the original feature space to a higher dimensional space where the data can be separated \citep{muller2001introduction,rosipal2002kernel,erkan2010semi}. Recall the multi-class logistic regression function from \eqref{e:prior}. Applying $\phi$ to the $\textbf{x}_{bi}$ in \eqref{e:prior}, where $\phi$ is the function transforming from feature space to kernel space yields
\begin{equation}
p(y_{bi}|\textbf{x}_{bi},\textbf{w})=\frac{\prod_{c=1}^{C}e^{I(y_{bi}=c)\textbf{w}_c^T\phi(\textbf{x}_{bi})}}{\sum_{c=1}^C{e^{\textbf{w}_c^T\phi(\textbf{x}_{bi})}}}.
\label{e:kernel1}
\end{equation}
Replacing $\textbf{w}_c=\sum_{b=1}^B\sum_{j=1}^{n_b}\boldsymbol\alpha_{c_{bj}}\phi(\textbf{x}_{bj})$ and $\phi(\textbf{x})^{T}\phi(\textbf{x}_{bj})=K(\textbf{x},\textbf{x}_{bj})$ into \eqref{e:kernel1}, yields
\begin{equation}
p(y_{bi}|\textbf{x}_{bi},\boldsymbol\alpha)=\frac{\prod_{c=1}^{C}e^{I(y_{bi}=c)\boldsymbol\alpha_c^T\textbf{k}(\textbf{x}_{bi})}}{\sum_{c=1}^C{e^{\boldsymbol\alpha_c^T\textbf{k}(\textbf{x}_{bi})}}},
\label{e:kernel2}
\end{equation}
where $\textbf{k}(\textbf{x})$ is $[K(\textbf{x},\textbf{x}_{11}), K(\textbf{x},\allowbreak\textbf{x}_{12}),\dots, K(\textbf{x},\textbf{x}_{Bn_B})]^T$. The kernel logistic regression does not require to compute the high dimensional $\phi(\textbf{x})$. Instead, only the dot product $K(\cdot,\cdot)$ is computed. In our paper, we consider an RBF kernel between instances $\textbf{x}$ and $\textbf{x}'$ as follows
\begin{equation}
K(\textbf{x},\textbf{x}')=e^{\frac{{-\lVert\textbf{x}-\textbf{x}'\rVert}_{2}^{2}}{\delta}},
\label{e:kernele}
\end{equation}
where $\delta$ is a tuning parameter to control the kernel width. Note that in kernel learning, the model parameter vector is $\boldsymbol\alpha$ instead of $\textbf{w}$. Thus, wherever encountering $\textbf{w}$, replace $\textbf{w}$ by $\boldsymbol\alpha$. Specifically, to implement the kernel version of our model, the following changes are made.
\begin{itemize}[leftmargin=30pt]
  \item E-step: Compute $p(y_{bi}=c|\textbf{Y}_b,\textbf{X}_b,\boldsymbol\alpha^{(k)})$ for $b=1,\dots,B$ and $i=1,\dots,n_b$ using Section \ref{ss:farfe}. Note that $p(y_{bi}|\textbf{x}_{bi},\textbf{w}^{(k)})$ is replaced by $p(y_{bi}|\textbf{x}_{bi},\boldsymbol\alpha^{(k)})$ given in \eqref{e:kernel2}.
  \item M-step: Compute $\boldsymbol\alpha^{(k+1)}$ using \eqref{e:12} and \eqref{e:13}. In \eqref{e:12}, $\textbf{w}$, $\textbf{w}^{(k)}$, and $\textbf{w}^{(k+1)}$ are replaced by $\boldsymbol\alpha$, $\boldsymbol\alpha^{(k)}$, and $\boldsymbol\alpha^{(k+1)}$, respectively. In \eqref{e:13}, $p(y_{bi}=c|\textbf{Y}_b,\textbf{X}_b,\textbf{w}^{(k)})$ is replaced by $p(y_{bi}=c|\textbf{Y}_b,\textbf{X}_b,\boldsymbol\alpha^{(k)})$ computed in the E-step, $p(y_{bi}|\textbf{x}_{bi},\textbf{w})$ is replaced by $p(y_{bi}|\textbf{x}_{bi},\boldsymbol\alpha)$ given in \eqref{e:kernel2}, and $\textbf{x}_{bi}$ is replaced by $\textbf{k}(\textbf{x}_{bi})$.

\end{itemize}

\section{Prediction}
We consider both instance level label prediction and bag level label prediction.
\subsection{Instance label prediction}
In instance label prediction, we consider two cases: predicting instance labels without knowing their bag label (inductive mode) and predicting instance labels when knowing their bag label (transductive mode). Consider the $i$th instance in the $t$th test bag. In the inductive setting, the predicted label maximizes the instance label probability given feature vector $\textbf{x}_{ti}$ and parameter $\textbf{w}$ is
\begin{equation}
\hat{y}_{ti}=\argmax_k{p(y_{ti}=k|\textbf{x}_{ti},\textbf{w})},
\label{e:inductive}
\end{equation}
where $p(y_{ti}=k|\textbf{x}_{ti},\textbf{w})$ is given in \eqref{e:prior}. Note that in the absence of bag level label, $\hat{y}_{ti}$ can be predicted independently using $\hat{y}_{ti}=\argmax_k{\textbf{w}_k^{T}\textbf{x}_{ti}}$, without the need for dynamic programming. Therefore, the computational complexity of inductive prediction for the $t$th test bag is $O(n_tCd)$.

In the transductive setting, the predicted label maximizes the joint probability of the instance label and its bag label as follows
\begin{equation}
\hat{y}_{ti}=\argmax_k{p(y_{ti}=k,\textbf{Y}_t|\textbf{X}_t,\textbf{w})},
\end{equation}
where $p(y_{ti}=k,\textbf{Y}_t|\textbf{X}_t,\textbf{w})$ is computed as described in Section \ref{ss:farfe}. As a result, the dynamic programming in Section \ref{ss:farfe} is used to obtain $p(y_{ti}=k,\textbf{Y}_t|\textbf{X}_t,\textbf{w})$ and find $\hat{y}_{ti}$. The computational complexity of transductive prediction for the $t$th test bag is $O(n_t|\textbf{Y}_t|2^{|\textbf{Y}_t|})+O(n_t|\textbf{Y}_t|d)$.

\subsection{Bag label prediction}
\label{ss:blp}
For the $t$th test bag, the predicted bag label $\hat{\textbf{Y}}_t$ is obtained by taking the union of its instance labels $\hat{y}_{ti}$ in \eqref{e:inductive} as follows
\begin{equation}
\hat{\textbf{Y}}_t=\bigcup_{i=1}^{n_t}\hat{y}_{ti}.
\end{equation}
\section{Experiments}
In this section, we evaluate and compare the proposed framework with several state-of-the-art approaches for MIML instance annotation and bag level prediction.

\subsection{General setting}
\label{ss:gs}

\noindent \textbf{Approaches.}
We compare the proposed approach of ORed-logistic regression model with maximum likelihood inference for instance annotation denoted by MLR with the following methods: SIM \citep{briggs2012}, MIMLfast (Mfast for short) \citep{huang2013fast}, and LSB-CMM (LSB for short) \citep{liu2012conditional}. For bag level predicton we include a comparison with MIMLSVM (M-SVM for short) \citep{zhou2007multi} and MIMLNN (M-NN for short) \citep{zhang2007multi,zhou2012multi}.

\noindent \textbf{Parameter tuning.} For MLR, by observation that the objective function in \eqref{e:sllh} stabilizes in around $50$ expectation maximization iterations for all datasets, we fix the number of iterations for the proposed EM framework to $50$. For SIM, we tune $\lambda$ over the set $\{10^{-4}, 10^{-5}, 10^{-6}, 10^{-7}, 10^{-8}, 10^{-9}\}$. For Mfast, following \cite{huang2013fast}, we search over the set $\{50, 100, 200\}$, $\{1, 5, 10\}$, $\{1, 5, 10, 15\}$, $\{10^{-4}, 5\cdot10^{-4}, 10^{-3}, 5\cdot10^{-3}\}$, $\{10^{-5}, 10^{-6}\}$ for parameters $m$, $C$, $K$, $\gamma_0$, $\eta$, respectively. Additionally, to satisfy the norm constraint in Mfast, we divide each feature vector by a constant in the set $\{10^{-2}, 3\cdot10^{-2}, 10^{-1}, 3\cdot10^{-1}, 1, 3, 10, 30, 10^{2}, 3\cdot10^{2}, 10^{3}\}$. Note that in the training phase, Mfast learns a parameter $\textbf{W}$ which can be used to compute the score $f_{Mfast}(\textbf{x}_{ti},c)$ indicating the confidence of instance $\textbf{x}_{ti}$ having class $c$. However, Mfast does not predict the label $\hat{\textbf{y}}_{ti}$ for each instance $\textbf{x}_{ti}$. It indirectly uses $f_{Mfast}(\textbf{x}_{ti},c)$ to compute $f_{Mfast}(\textbf{X}_{t},c)$ indicating the confidence of the $t$th test bag having class $c$. In the experiments, we access $f_{Mfast}(\textbf{x}_{ti},c)$ and predict the label $\hat{\textbf{y}}_{ti}$ as $\hat{\textbf{y}}_{ti}=\argmax_{c}f_{Mfast}(\textbf{x}_{ti},c)$. For LSB, we search $\sigma^{2}$ over the set $\{10^{-2}, 10^{-1},\allowbreak 1, 10, 10^{2}, 10^{3}, 10^{4}, 10^{5}\}$ as suggested by the authors. Moreover, we set the parameter $\alpha$ as 0.05 and set $K$ to 5 $\times$ (the number of classes). For M-SVM, we use RBF kernel, with $\gamma$ in \{0.1, 0.2, 0.4, 0.8\} and the parameter ratio is searched over the set \{0.2, 0.4, 0.6, 0.8, 1\}. For M-NN, we search over the set \{0.2, 0.4, 0.6, 0.8, 1\} for the parameter ratio, and the parameter $\lambda$ over the set $\{10^{-2}, 10^{-1}, 1, 10\}$.

\noindent \textbf{Datasets.} We compare the aforementioned approaches on six datasets: HJA birdsong, MSCV2, Letter Frost, Letter Carroll, Voc12 \citep{briggs2012}, and 50Salad \citep{Stein_2013b}. For the HJA birdsong, there are a few bags where the bag label is not equal the union of its instance labels. Therefore, we create an additional version of the HJA birdsong dataset namely HJA union in which the label of each bag is the union of its instance labels. For the Letter Carroll and Letter Frost datasets, bags are created from words of two poems, and instances of each bag are their letters which are sampled from the Letter Recognition dataset \citep{frey1991letter} on the UCI Machine Learning repository. However, in each of these two poems, two letters are missing. Therefore, we only consider 24 classes in each dataset. Detailed information of all the aforementioned datasets is shown in Table \ref{table:1}.\\

\begin{table}[H]
\centering
    \begin{tabular}{ | l || l | l | l | l | l| l |}
    \hline
    \multirow{2}{*} {Dataset} & classes & bags &instances& dimension & classes per & instances per\\
    {} & $\textit{(C)}$& $\textit{(B)}$&  $\textit{(N)}$ &  $\textit{(d)}$ & bag $\overline{\textbf{Y}}_b$ & bag $\overline{n}_b$\\
    \hline\hline
    HJA bird & 13 & 548 & 4,998 & 38 &2.1 &9.1\\ \hline
    MSCV2 & 23 & 591 & 1,758 & 48 &2.5 &3.0\\ \hline
    Letter Carroll & 24 & 166 & 717 & 16 &3.9 &4.3\\ \hline
    Letter Frost & 24 & 144 & 565 & 16 &3.6 &3.9\\ \hline
    Voc12 &20 &1,053 &4,142 &48 &2.3 &3.9\\ \hline
    50Salad &6 &124 &2,020  &57 &2.3 &16.3\\ \hline
    \end{tabular}
\caption{Statistics of datasets in our experiments}
\label{table:1}
\end{table}

\noindent \textbf{Evaluation measures.} We consider different metrics for evaluation of instance annotation and bag label prediction. For instance annotation, we consider instance level accuracy by dividing the correctly predicted label instances by the total number of predicted instances. For bag label prediction, we consider Hamming loss, ranking loss, average precision, one error, and coverage as in \cite{zhou2007multi,zhang2008m3miml,zhou2012multi,huang2013fast}.

\subsection{Instance annotation experiments}
\textbf{Setting.} In this section, we compare the proposed MLR approach with Mfast, SIM, and LSB methods. We also consider the logistic regression classifier in the single instance single label setting (SLR) in which all instance labels are provided. The accuracy of SLR is considered as an upper bound for those of all methods. Furthermore, we consider the dummy classifier that classifies every instance in the test dataset with the most common class in the training dataset without using any information from the feature vectors. The accuracy of the dummy classifier is considered as a lower bound for those of all methods. For the inductive setting, we use 10-fold cross validation to evaluate the performance. For both inductive and transductive settings, there are two methods, namely post-hoc parameter tuning, for selecting the optimal parameters: (1) using an instance level metric and (2) using a bag level metric. Note that the proposed MLR approach is free of parameter tuning hence we only consider these tuning schemes for baseline methods. In that way, there is no unfair advantage for MLR from selecting the best parameter settings. The inductive results with post-hoc tuning using instance level accuracy and bag level accuracy are presented in Tables \ref{table:2} and \ref{table:3}, respectively. The transductive results with post-hoc tuning using instance level accuracy and bag level accuracy are presented in Tables \ref{table:4} and \ref{table:5}, respectively.

\noindent \textbf{Inductive results.} From Tables \ref{table:2} and \ref{table:3}, the MLR approach outperforms and in some cases significantly outperforms other methods on the datasets considered. On MSCV2, Letter Carroll, and Letter Frost, the results of MLR are from 7\% to 12\% higher than those of Mfast, SIM, and LSB, whose performances are comparable. For HJA bird, Voc12, and 50Salad, LSB is comparable or achieves a higher accuracy than Mfast which outperforms SIM. For these three datasets, the proposed MLR approach outperforms LSB 3\% on HJA bird and is comparable with LSB on Voc12 and 50Salad. Furthermore, the accuracy obtained by MLR is only 5-6\% lower than that of SLR on HJA bird and MSCV2 datsets, and 1-3\% on other datasets.

\noindent \textbf{Analysis.} The accuracy of all methods on HJA dataset is lower than on HJA union dataset. This is due to the violation of the union assumption in the HJA dataset. The accuracy of SLR approach on both datasets is the same since SLR relies only on the instance labels. MLR significantly outperforms other methods in MSCV2, Letter Carroll, and Letter Frost. The reason is that for those datasets, the average number of classes per bag is high. By carefully considering all possibilities for instance labels and avoiding any approximation in inference, MLR uses all the instances effectively. In contrast, Mfast and SIM use the max or softmax principle which may ignore useful information from most of the instances. Moreover, LSB uses an approximate inference method, MCMC sampling. In general, sampling methods are slow to converge and due to randomness present a challenge in establishing a stopping criteria. Furthermore, in LSB, each instance has a set of possible labels which is the label of its bag. As a result, LSB may ignore a useful constraint, that the union of instance labels equals their bag label, which is preserved in our dynamic programming approach. For the Voc12 dataset, the dummy classifier works well since there exists a dominant class consisting of more than 30\% instances of the dataset. In Table \ref{table:3}, the accuracy of MLR is similar to that in Table \ref{table:2} since MLR has no tuning parameter. For LSB and SIM, the accuracy in Table \ref{table:3} is slightly smaller than in Table \ref{table:2} since the tuning is performed indirectly using the bag level measurement instead of directly using the instance level accuracy. For Mfast, while we follow the parameter tuning scheme proposed in \cite{huang2013fast}, we understand that post-hoc tuning benefits methods with a high number of parameter settings. Since we consider a large number of choices (a total of 3,168) for the parameters, the accuracy drops significantly from the scenario when we both tune and test on instance level metric, as in Table \ref{table:2}, to the case when we tune on bag level metric and test on instance level metric, as in Table \ref{table:3}.

\noindent \textbf{Transductive results and analysis.} In Tables \ref{table:4} and \ref{table:5}, there is no standard deviation reported for the transductive setting. The reason is that since bag level labels are known in both training and test phases, approaches are trained and tested on the whole dataset instead of dividing into 10 folds as in the inductive setting. MLR outperforms all other methods on all datasets, especially on MSCV2, Letter Carroll, and Letter Frost. For example, on MSCV2, the accuracy of MLR is 14\%, 15\%, and 23\% higher than those of Mfast, SIM, and LSB, respectively. For datasets HJA union and Voc12, MLR outperforms LSB even though they are comparable in the inductive setting with instance level tuning. Similarly, even though SIM, LSB, and Mfast perform comparably on MSCV2 in inductive setting, their reported accuracies are noticeably different in transductive setting.

\begin{table}
\fontsize{10}{13}\selectfont
\centering
    \begin{tabular}{ | p{11mm} || p{15mm} | p{17mm} | p{14mm} | l | l | l | l |}
    \hline
    Dataset       & HJA bird               & HJA union             & MSCV2                     & Carroll                      & Frost             &Voc12                &50Salad\\ \hline\hline
    {\bf MLR}     & {\bf69.1$\pm$4.7}      & {\bf72.3$\pm$3.8}     & {\bf54.8$\pm$4.7}         & {\bf67.7$\pm$5.2}            &{\bf71.3$\pm$7.6}  &{\bf43.2$\pm$3.2}    &{\bf76.0$\pm$5.3}\\ \hline\hline
    Mfast         & 60.5$\pm$5.5           & 61.3$\pm$3.6          & 47.6$\pm$5.1              & 56.1$\pm$5.9                 &58.9$\pm$5.3       &{\bf42.5$\pm$3.3}    &{\bf72.4$\pm$7.5}\\ \hline
    SIM           & 61.9$\pm$4.2           & 62.7$\pm$4.1          & 46.7$\pm$4.4              & 54.2$\pm$6.5                 &58.1$\pm$5.7       &38.0$\pm$3.4         &66.9$\pm$7.3\\ \hline
    LSB           & 66.3$\pm$4.3           & {\bf72.0$\pm$4.0}     & 45.4$\pm$4.8              & 55.9$\pm$2.6                 &55.5$\pm$3.6       &{\bf42.8$\pm$4.5}    &{\bf76.8$\pm$7.8}\\ \hline\hline
    SLR           & 74.6$\pm$3.1           & 74.6$\pm$3.1          & 60.2$\pm$4.8              & 70.5$\pm$4.8                 &72.7$\pm$4.0       &45.2$\pm$3.9         &77.6$\pm$4.7\\ \hline
    Dummy         & 12.1$\pm$3.4           & 12.1$\pm$3.4          & 14.5$\pm$3.7              & 11.2$\pm$3.7                 &09.3$\pm$2.4       &36.1$\pm$4.1         &16.7$\pm$7.7\\ \hline
    \end{tabular}
\caption{Accuracy results (percentage) for instance label prediction in inductive mode for MLR, Mfast, SIM, LSB, SLR, and Dummy approaches using instance level accuracy parameter tuning. The proposed approach and values that are statistically indistinguishable using two-tailed paired $\textit{t}$-tests at 95\% confidence level with the optimal performances are bolded.}
\label{table:2}
\end{table}

\begin{table}
\fontsize{10}{13}\selectfont
\centering
    \begin{tabular}{ | p{11mm} || p{15mm} | p{17mm} | p{14mm} | l | l | l | l |}
    \hline
    Dataset       & HJA bird                    & HJA union             & MSCV2                     & Carroll                      & Frost                      &Voc12                      &50Salad\\ \hline\hline
    {\bf MLR}     & {\bf69.1$\pm$4.7}           & {\bf72.3$\pm$3.8}     & {\bf54.8$\pm$4.7}         & {\bf67.7$\pm$5.2}            &{\bf71.3$\pm$7.6}           &{\bf43.2$\pm$3.2}          &{\bf76.0$\pm$5.3}\\ \hline\hline
    Mfast         & 60.0$\pm$3.6                & 59.8$\pm$4.9          & 41.5$\pm$5.8              & 51.5$\pm$6.5                 &45.4$\pm$7.1                &41.6$\pm$2.8               &{\bf72.4$\pm$7.5}\\ \hline
    SIM           & 61.7$\pm$4.3                & 62.7$\pm$4.1          & 46.7$\pm$4.4              & 54.2$\pm$6.5                 &58.1$\pm$5.7                &38.0$\pm$3.4               &66.9$\pm$7.3\\ \hline
    LSB           & 66.1$\pm$4.4                & {\bf71.4$\pm$4.3}     & 45.2$\pm$4.0              & 55.9$\pm$2.6                 &55.5$\pm$3.6                &{\bf42.0$\pm$4.3}          &{\bf75.9$\pm$6.7}\\ \hline\hline
    SLR           & 74.6$\pm$3.1                & 74.6$\pm$3.1          & 60.2$\pm$4.8              & 70.5$\pm$4.8                 &72.7$\pm$4.0                &45.2$\pm$3.9               &77.6$\pm$4.7\\ \hline
    Dummy         & 12.1$\pm$3.4                & 12.1$\pm$3.4          & 14.5$\pm$3.7              & 11.2$\pm$3.7                 &09.3$\pm$2.4                &36.1$\pm$4.1               &16.7$\pm$7.7\\ \hline
    \end{tabular}
\caption{Accuracy results (percentage) for instance label prediction in inductive mode for MLR, Mfast, SIM, LSB, SLR, and Dummy approaches using Hamming loss parameter tuning. The proposed approach and values that are statistically indistinguishable using two-tailed paired $\textit{t}$-tests at 95\% confidence level with the optimal performances are bolded.}
\label{table:3}
\end{table}

\begin{table}
\small
\centering
    \begin{tabular}{| l || l | l | l | l | l | l | l |}
    \hline
    Dataset       & HJA bird             & HJA union             & MSCV2                      & Carroll                     & Frost                 &Voc12                          &50Salad\\ \hline\hline
    {\bf MLR}     & 80.8         & {\bf88.0}     & {\bf84.9}          & {\bf91.5}           & {\bf91.5}     &{\bf67.9}             &86.8\\ \hline\hline
    Mfast         & 78.3         & 81.8          & 70.6               & 72.7                & 75.8          &61.0                  &84.7\\ \hline
    SIM           & {\bf81.7}    & 83.5          & 69.7               & 72.2                & 77.3          &63.1                  &83.1\\ \hline
    LSB           & 77.0         & 85.7          & 61.1               & 70.0                & 66.2          &57.8                  &{\bf88.3}\\ \hline\hline
    Dummy         & 51.6         & 56.3          & 38.0               & 25.2                & 27.8          &50.9                  &46.9\\ \hline
    \end{tabular}
\caption{Accuracy results (percentage) for instance level prediction in transductive mode for MLR, Mfast, SIM, LSB, and Dummy approaches using instance level accuracy parameter tuning. The proposed approach and optimal performances are bolded.}
\label{table:4}
\end{table}

\begin{table}
\small
\centering
    \begin{tabular}{| l || l | l | l | l | l | l | l |}
    \hline
    Dataset       & HJA bird             & HJA union             & MSCV2                      & Carroll                     & Frost                   &Voc12                    &50Salad\\ \hline\hline
    {\bf MLR}     & 80.8         & {\bf88.0}     & {\bf84.9}          & {\bf91.5}           & {\bf91.5}       &{\bf67.9}        &86.8\\ \hline\hline
    Mfast         & 77.5         & 80.0          & 65.1               & 67.3                & 70.2            &58.4             &84.7\\ \hline
    SIM           & {\bf81.7}    & 83.5          & 69.7               & 71.7                & 76.5            &63.1             &82.0\\ \hline
    LSB           & 75.4         & 81.0          & 59.0               & 68.3                & 65.7            &54.8             &{\bf87.5}\\ \hline\hline
    Dummy         & 51.6         & 56.3          & 38.0               & 25.2                & 27.8            &50.9             &46.9\\ \hline
    \end{tabular}
\caption{Accuracy results (percentage) for instance level prediction in transductive mode for MLR, Mfast, SIM, LSB, and Dummy approaches using Hamming loss parameter tuning. The proposed approach and optimal performances are bolded.}
\label{table:5}
\end{table}

\subsection{Instance annotation accuracy vs.~the number of training bags}

\begin{figure}
\begin{minipage}[b]{0.48\linewidth}
  \centering
  \centerline{\includegraphics[width=7.8cm]{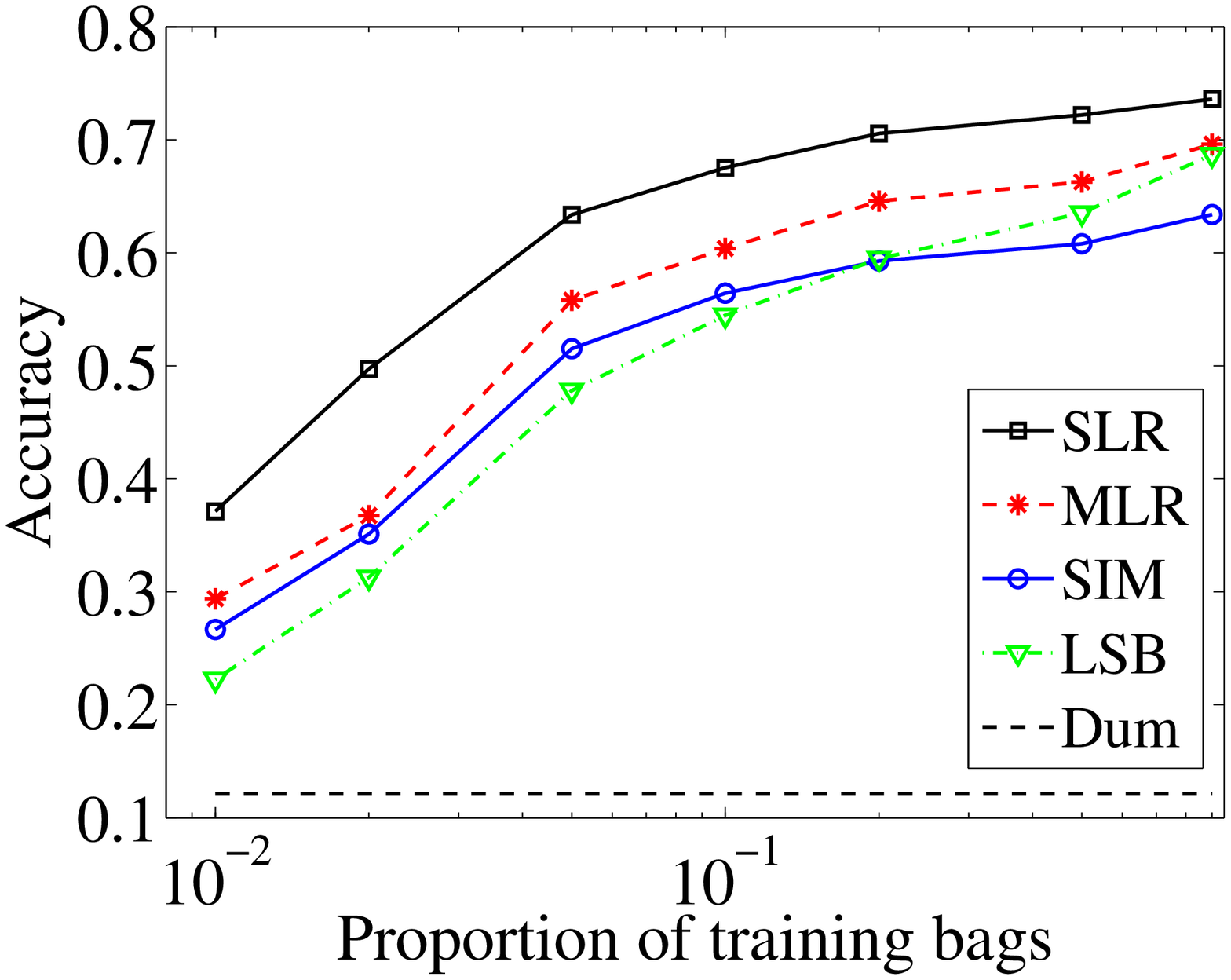}}
  \centerline{HJA bird}\medskip
\end{minipage}
\hfill
\begin{minipage}[b]{0.48\linewidth}
  \centering
  \centerline{\includegraphics[width=7.8cm]{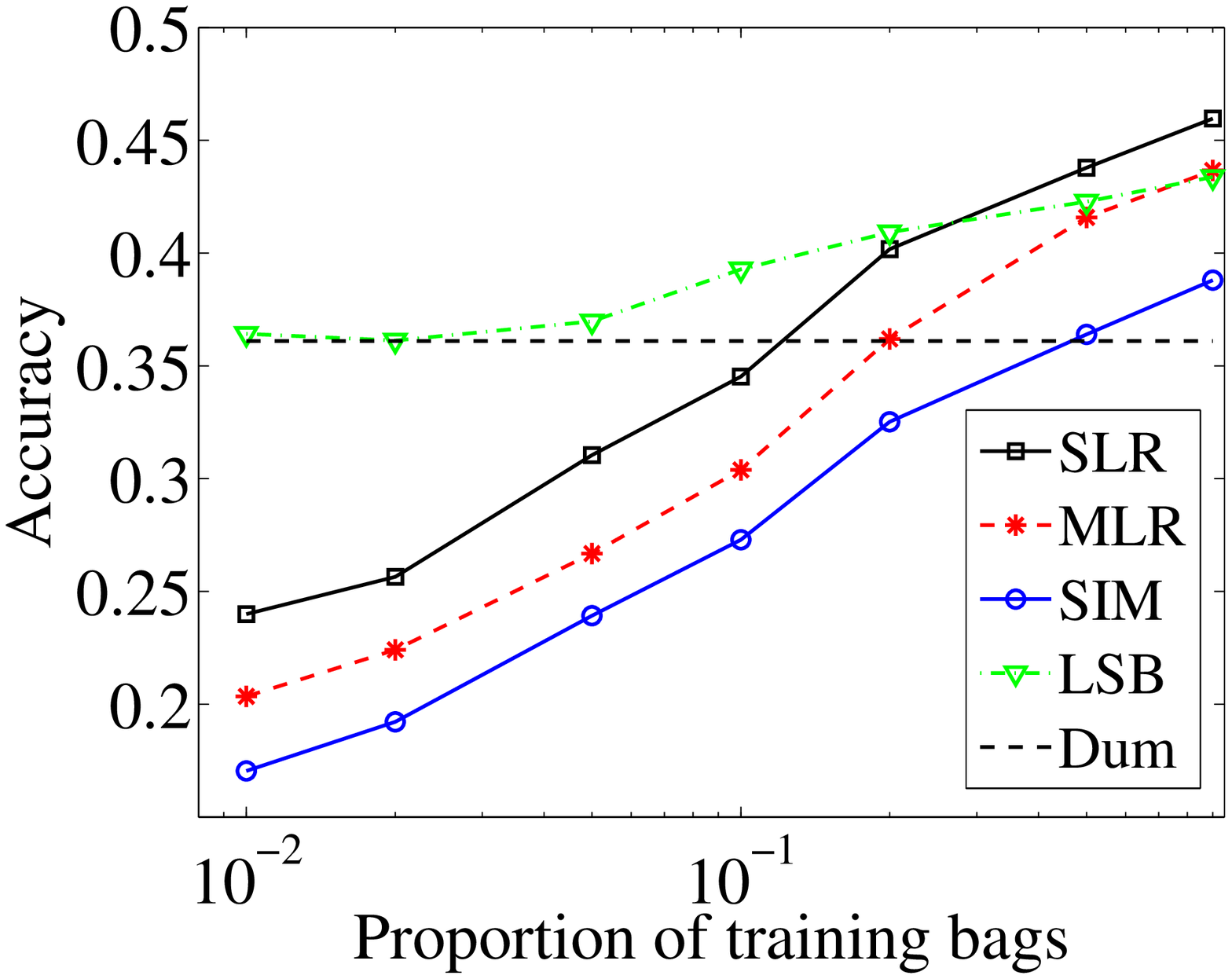}}
  \centerline{Voc12}\medskip
\end{minipage}

\begin{minipage}[b]{0.48\linewidth}
  \centering
  \centerline{\includegraphics[width=7.8cm]{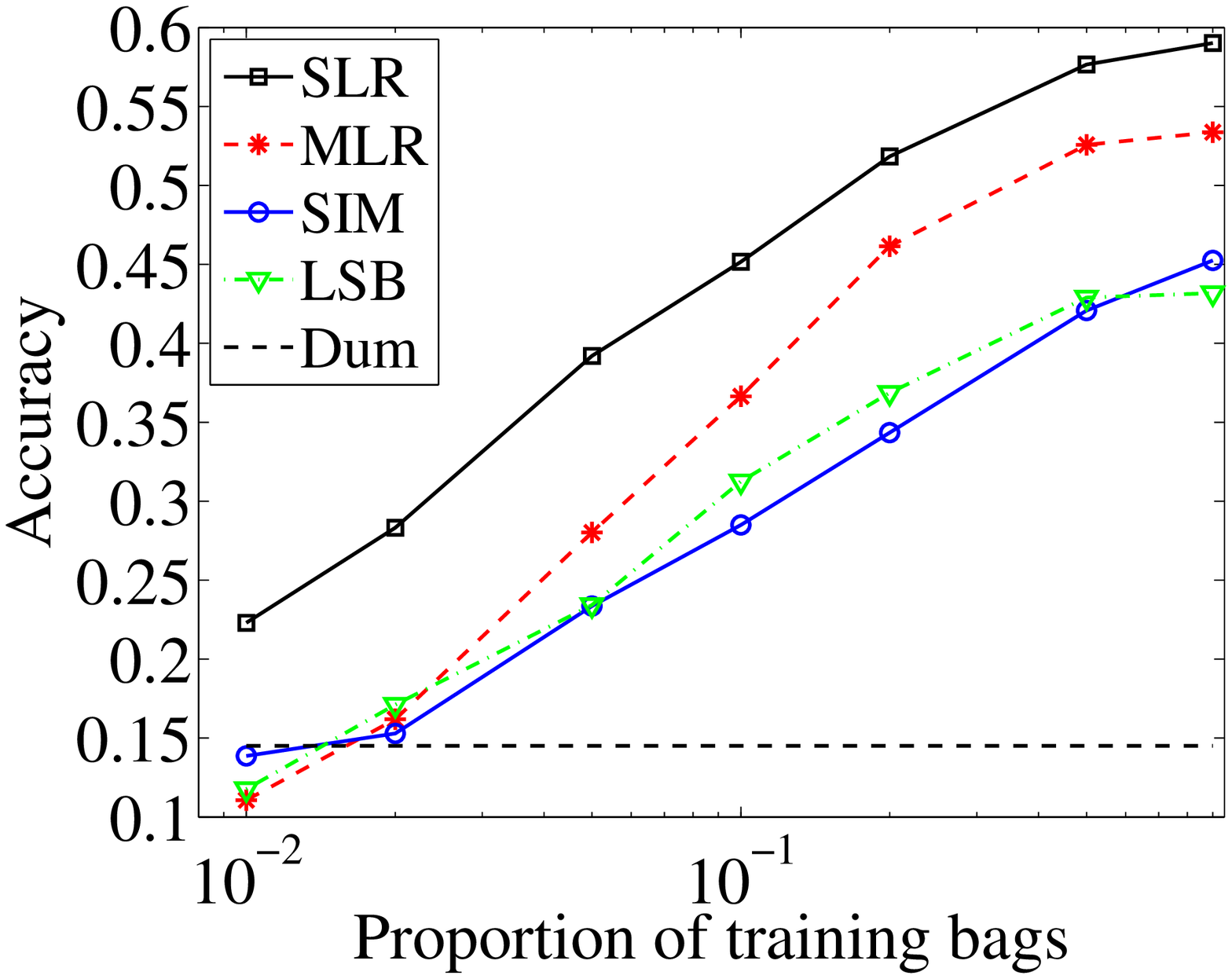}}
  \centerline{MSCV2}\medskip
\end{minipage}
\hfill
\begin{minipage}[b]{0.48\linewidth}
  \centering
  \centerline{\includegraphics[width=7.8cm]{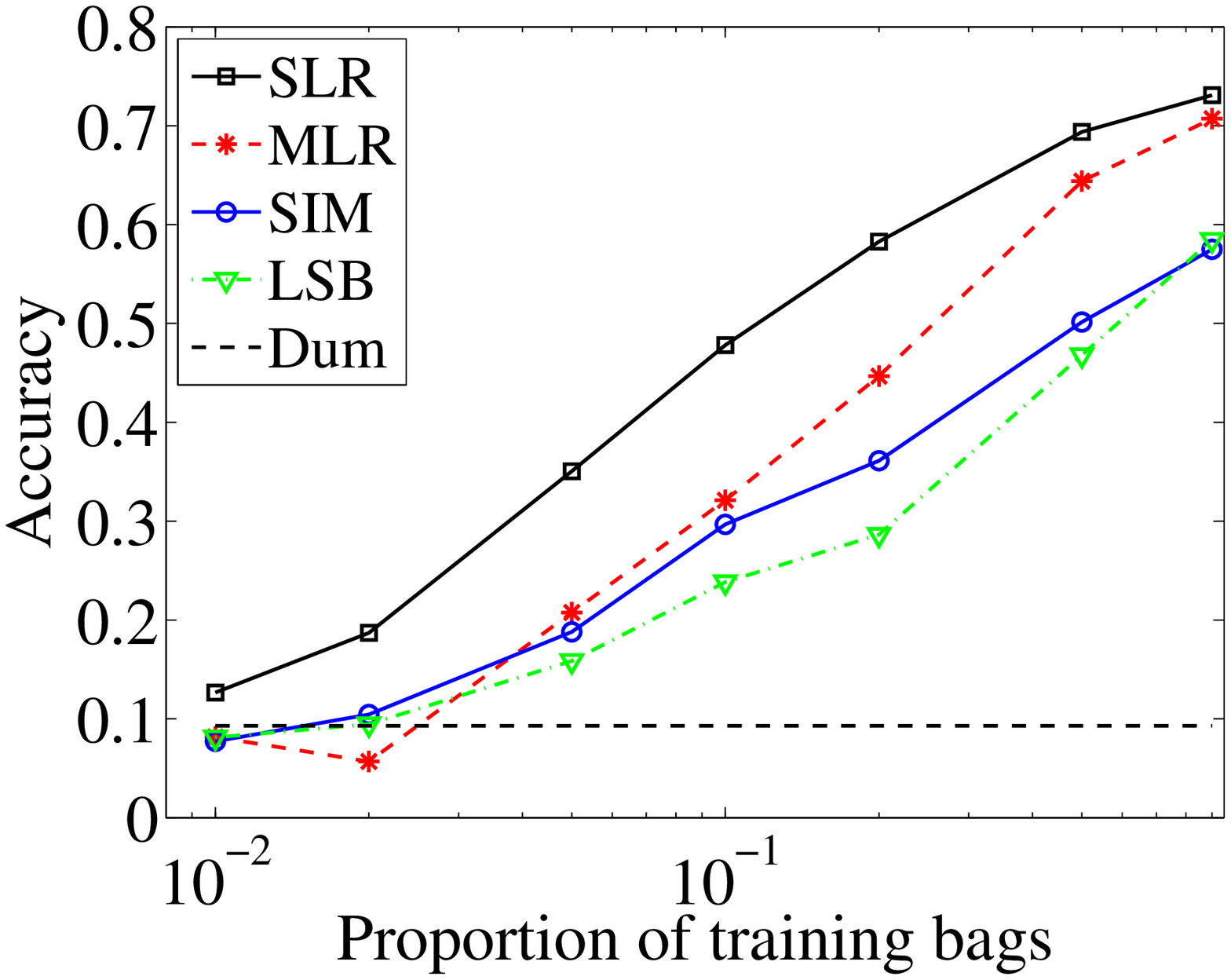}}
  \centerline{Letter Frost}\medskip
\end{minipage}
\begin{minipage}[b]{0.48\linewidth}
  \centering
  \centerline{\includegraphics[width=7.8cm]{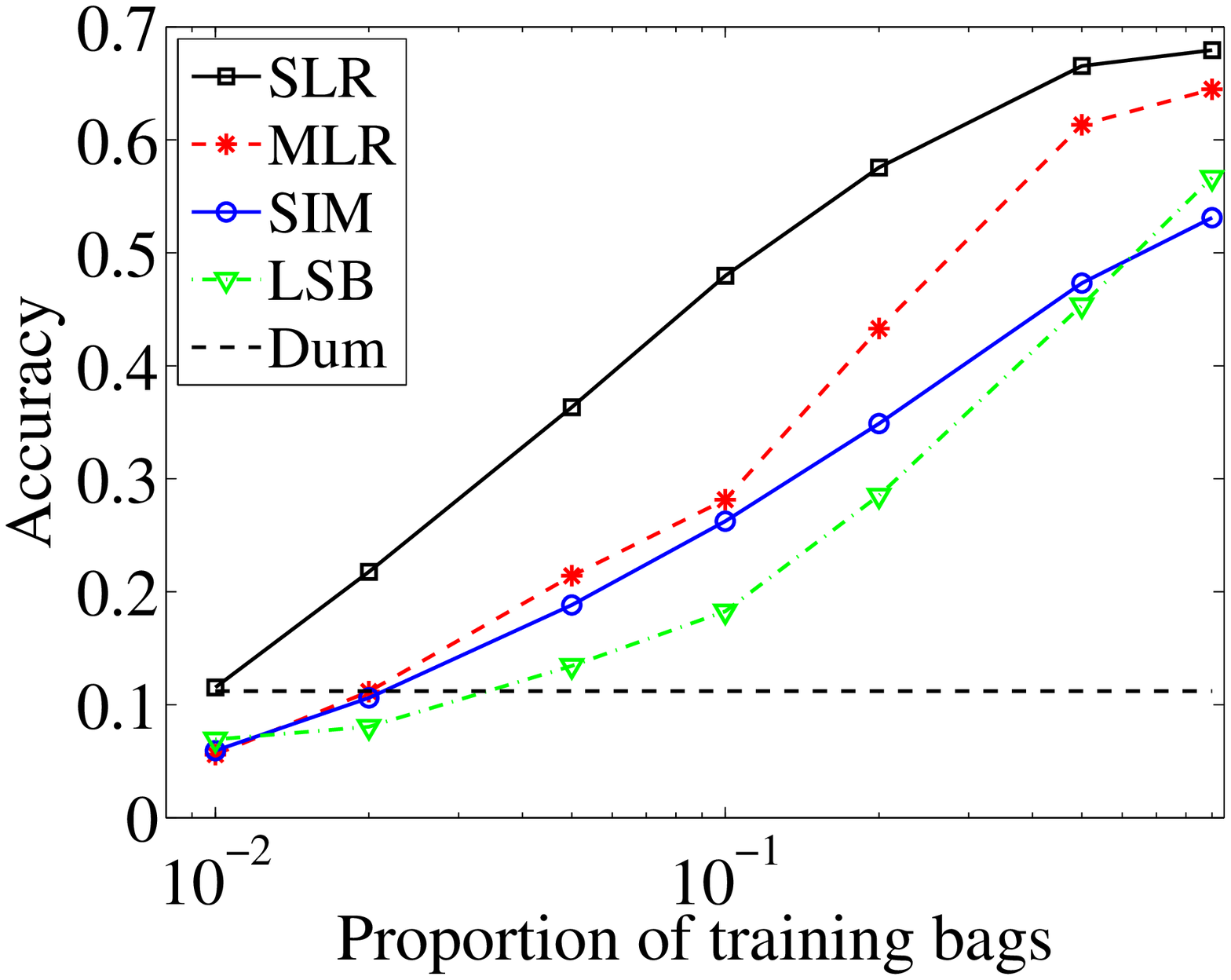}}
  \centerline{Letter Carroll}\medskip
\end{minipage}
\hfill
\begin{minipage}[b]{0.48\linewidth}
  \centering
  \centerline{\includegraphics[width=7.8cm]{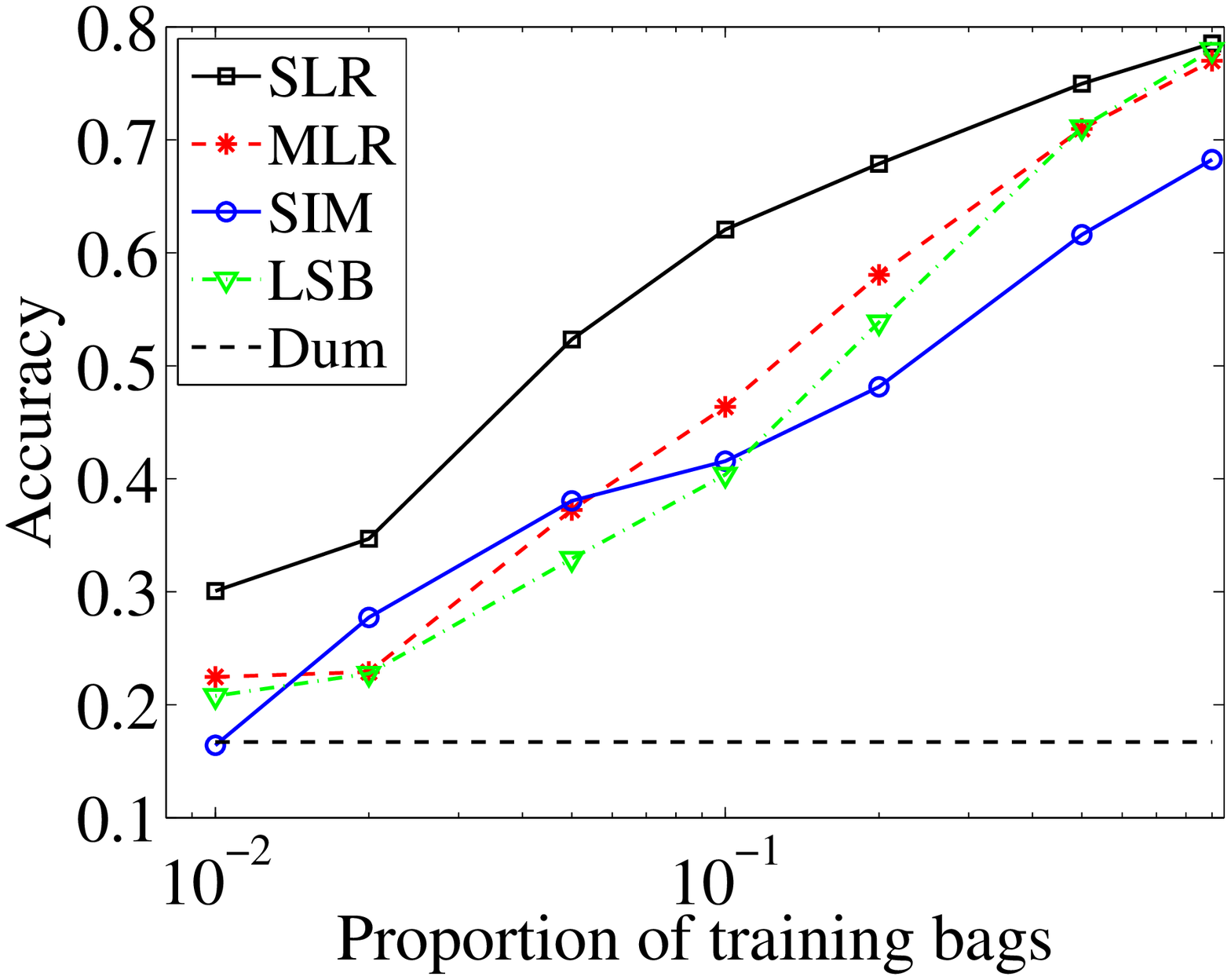}}
  \centerline{50Salad}\medskip
\end{minipage}
\caption{Instance annotation accuracy vs.~percentage of training bags for MLR, SIM, LSB, SLR, and Dummy approaches in inductive mode}
\label{fig:accvsbag}
\end{figure}

\textbf{Setting.} In this section, we examine the ability of the proposed approach to effectively use the information from all instances compared to LSB and SIM. We exclude Mfast since its post-hoc instance level tuning accuracy is far different from its bag level tuning accuracy. In addition, adding one parameter for the percentage of training bags leading to a larger number of parameter settings which is already high for Mfast, and longer runtime. We perform the experiments on HJA, MSCV2, Letter Carroll, Letter Frost, Voc12, and 50Salad datasets. For each dataset we randomly select \{1\%, 2\%, 5\%, 10\%, 20\%, 50\%, 90\%\} of the data for training and the remaining data for testing. For each of these sampling percentage values, we perform the experiment 10 times and report the average accuracy. The accuracy as a function of the sampling percentage is depicted in Figure \ref{fig:accvsbag}.

\noindent \textbf{Results.} From Figure \ref{fig:accvsbag}, we observe that the accuracy for each method increases with the number of training bags. Furthermore, in order to achieve similar accuracy, our MLR approach uses fewer bags. This has significant implications in practice since the data labeling process is costly. In addition, the time required for training could also be saved. For example, in MSCV2, MLR can achieve similar level of accuracy to that of SIM and LSB using only 17\% of training data SIM and LSB use. With a small amount of data, there is a clear gap between MLR and the ideal upper bound SLR. However, with enough training data, the performance of MLR is close to SLR. For example, in 50Salad, at 10\% of training data, the performance of MLR is 20\% less than that of SLR. However, with 90\% data for training, the gap is only 1-2\%. From the figure, we could observe how efficiently each approach uses information from instance labels. SLR achieves high accuracy even with a small number of training bags since it is directly given instance labels. MLR achieves a similar level of accuracy using a smaller number of bags, compared to SIM and LSB, indicating that MLR efficiently uses information from bag labels. In SIM, the softmax function for computing the weight of each instance in bags is heuristically defined instead of directly coming from the objective function. As a result, SIM may not efficiently use information from all instances. For LSB, a possible explanation for its lower accuracy compared to that of MLR is that approximation methods including variational EM and MCMC sampling are used to inference. Another possible explanation is that LSB does not maintain the constraint that the union of instance labels in each bag is equal to their bag label. We also observe that the accuracy of SIM and LSB are quite similar.

\begin{figure}
\center
\includegraphics[width=7.4cm]{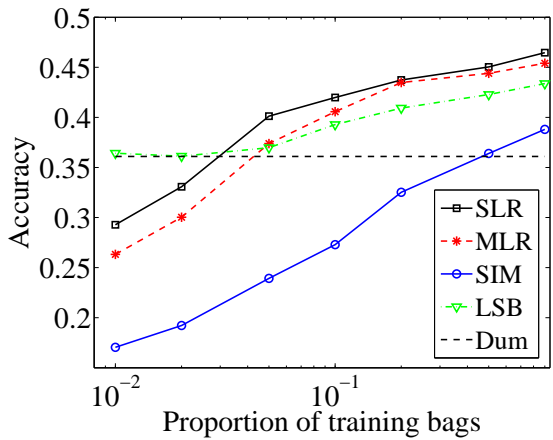}
\caption{Instance annotation accuracy vs.~percentage of training bags for MLR and SLR (with $l_2$-norm regularization), SIM, LSB, and Dummy approaches on Voc12 in inductive mode}
\label{fig:Voc_reg}
\end{figure}

\noindent \textbf{Special case for Voc12.} For small training data, if we have a high number of classes, over-fitting problem may occur, as in Voc12. Specifically, MLR and SLR are outperformed by the dummy classifier, whose accuracy is around 36\%, since there is a dominant class in the dataset such that more than 30\% of the instances in that class. We can mitigate these problems by adding an $l_2$-norm regularization term to the objective function for both MLR and SLR. LSB is less susceptible to over-fitting problem since LSB uses a parameter $\sigma^{2}$ for regularization which is searched over the set $\{10^{-2}, 10^{-1},\dots,10^{5}\}$ as in Section \ref{ss:gs}. The results are presented in Figure \ref{fig:Voc_reg}, indicating that if we use more than 5\% of data for training, SLR and MLR can obtain a higher accuracy compared to the dummy classifier.

\subsection{Bag label prediction}

\begin{table}
\fontsize{9}{11}\selectfont
\centering
\begin{tabular}{| l|| p{13mm}||p{14mm}|p{13mm}| p{13mm}|p{13mm}|| p{13mm}|p{13mm}|| p{13mm}|}
    \hline
                            &SLR                &{\bf MLR}                  &SIM                    &Mfast                  &LSB                        &M-SVM                  &M-NN                    &Dummy\\ \hline
    \multicolumn{9}{|l|}{HJA bird} \\\hline
    $\downarrow$ hl        &11.2$\pm$0.9    &09.6$\pm$1.0            &15.9$\pm$1.5           &05.5$\pm$1.1           &10.6$\pm$1.5               &{\bf04.5$\pm$0.6}        &{\bf04.7$\pm$1.1}          &18.7$\pm$0.8\\ \hline
    $\downarrow$ rl        &03.5$\pm$0.4    &{\bf02.7$\pm$0.6}       &{\bf02.2$\pm$0.8}      &{\bf02.5$\pm$0.7}      &06.9$\pm$1.8               &{\bf02.7$\pm$1.1}        &{\bf02.7$\pm$1.1}          &32.3$\pm$3.1\\ \hline
    $\uparrow$ ap          &92.5$\pm$1.0    &{\bf94.2$\pm$1.2}       &{\bf94.1$\pm$1.8}      &{\bf94.1$\pm$1.4}      &89.7$\pm$2.6               &{\bf94.0$\pm$2.0}        &{\bf93.9$\pm$2.8}          &44.1$\pm$3.2\\ \hline
    $\downarrow$ oe        &05.7$\pm$3.3    &{\bf03.8$\pm$1.8}       &{\bf05.1$\pm$3.1}      &{\bf03.7$\pm$2.4}      &{\bf03.7$\pm$1.7}          &{\bf04.6$\pm$2.6}        &{\bf05.3$\pm$4.4}               &64.0$\pm$6.2\\ \hline
    $\downarrow$ co        &15.2$\pm$1.6    &13.9$\pm$1.6            &{\bf12.4$\pm$1.6}      &{\bf13.4$\pm$1.6}      &21.7$\pm$3.6               &{\bf13.2$\pm$1.6}        &13.4$\pm$1.3               &45.9$\pm$2.7\\ \hline
    \multicolumn{9}{|l|}{HJA union} \\\hline
    $\downarrow$ hl       &11.2$\pm$1.0    &11.1$\pm$1.2            &16.1$\pm$1.9           &05.5$\pm$1.0           &10.1$\pm$1.5               &{\bf04.5$\pm$0.5}        &{\bf04.7$\pm$0.7}          &18.7$\pm$0.8\\ \hline
    $\downarrow$ rl       &02.8$\pm$0.5    &{\bf02.7$\pm$0.6}       &{\bf02.4$\pm$0.7}      &{\bf02.6$\pm$0.6}      &04.1$\pm$1.3               &{\bf02.6$\pm$1.0}        &{\bf02.6$\pm$0.9}          &33.8$\pm$2.9\\ \hline
    $\uparrow$ ap         &93.3$\pm$1.5    &{\bf93.7$\pm$1.7}       &{\bf93.6$\pm$1.9}      &{\bf93.9$\pm$1.5}      &{\bf93.0$\pm$2.3}          &{\bf93.7$\pm$2.1}        &{\bf94.4$\pm$2.0}          &40.5$\pm$2.7\\ \hline
    $\downarrow$ oe       &06.0$\pm$3.7    &{\bf05.7$\pm$3.7}            &{\bf06.0$\pm$2.6}      &{\bf04.2$\pm$2.6}      &{\bf03.8$\pm$3.1}          &{\bf05.3$\pm$2.8}        &{\bf04.0$\pm$2.8}          &67.8$\pm$10.0\\ \hline
    $\downarrow$ co       &12.5$\pm$1.2    &12.7$\pm$1.7            &{\bf11.5$\pm$1.3}      &12.4$\pm$1.3           &15.7$\pm$2.5               &{\bf12.2$\pm$1.3}        &{\bf12.4$\pm$1.4}          &45.1$\pm$2.7\\ \hline
    \multicolumn{9}{|l|}{MSCV2} \\\hline
    $\downarrow$ hl       &07.3$\pm$1.0    &07.7$\pm$1.0            &08.9$\pm$0.8           &09.0$\pm$0.9           &07.9$\pm$0.8               &07.2$\pm$1.1             &{\bf06.4$\pm$0.8}          &13.0$\pm$0.4\\ \hline
    $\downarrow$ rl       &08.7$\pm$1.5    &08.7$\pm$1.6            &10.5$\pm$2.2           &10.7$\pm$1.6           &17.0$\pm$2.8               &08.1$\pm$1.6             &{\bf07.3$\pm$2.0}          &34.2$\pm$3.5\\ \hline
    $\uparrow$   ap       &73.2$\pm$5.1    &73.1$\pm$4.9            &70.3$\pm$3.8           &69.3$\pm$4.1           &65.2$\pm$4.4               &74.6$\pm$5.1             &{\bf78.5$\pm$5.6}          &40.2$\pm$3.0\\ \hline
    $\downarrow$ oe       &26.6$\pm$7.4    &27.1$\pm$6.1            &27.1$\pm$6.6           &27.9$\pm$6.8           &25.4$\pm$5.9               &25.1$\pm$7.7             &{\bf20.3$\pm$6.2}          &64.0$\pm$7.7\\ \hline
    $\downarrow$ co       &20.6$\pm$2.6    &20.5$\pm$2.9            &23.3$\pm$3.3           &23.3$\pm$3.1           &36.3$\pm$3.9               &19.8$\pm$2.3             &{\bf18.3$\pm$2.8}          &53.0$\pm$3.3\\ \hline
    \multicolumn{9}{|l|}{Letter Carroll} \\\hline
    $\downarrow$ hl       &07.8$\pm$1.4    &{\bf09.0$\pm$1.3}       &11.9$\pm$1.5           &11.9$\pm$1.2           &10.8$\pm$1.0               &14.9$\pm$1.2             &14.2$\pm$1.8               &18.9$\pm$1.4\\ \hline
    $\downarrow$ rl       &06.6$\pm$2.6    &{\bf07.4$\pm$2.8}       &13.5$\pm$4.0           &12.2$\pm$3.2           &13.9$\pm$2.3               &20.5$\pm$3.3             &17.7$\pm$2.3               &25.7$\pm$3.3\\ \hline
    $\uparrow$   ap       &84.5$\pm$5.0    &{\bf83.1$\pm$4.6}       &73.0$\pm$5.4           &75.6$\pm$5.1           &74.9$\pm$2.7               &58.6$\pm$8.4             &62.2$\pm$4.5               &47.1$\pm$5.3\\ \hline
    $\downarrow$ oe       &06.7$\pm$7.4    &{\bf05.5$\pm$6.1}       &{\bf12.0$\pm$9.1}      &{\bf06.6$\pm$5.9}      &{\bf09.1$\pm$3.3}          &38.1$\pm$20.6            &28.3$\pm$10.3              &56.1$\pm$15.4\\ \hline
    $\downarrow$ co       &28.1$\pm$5.4    &{\bf29.8$\pm$5.2}       &39.5$\pm$5.4           &36.8$\pm$4.6           &41.6$\pm$4.6               &44.3$\pm$2.6             &41.4$\pm$3.8               &48.4$\pm$4.9\\ \hline
    \multicolumn{9}{|l|}{Letter Frost} \\\hline
    $\downarrow$ hl       &06.7$\pm$1.1    &{\bf07.5$\pm$2.2}       &10.7$\pm$1.7           &11.2$\pm$1.5           &10.2$\pm$1.3               &12.3$\pm$2.5             &11.9$\pm$2.2               &17.2$\pm$2.0\\ \hline
    $\downarrow$ rl       &05.7$\pm$2.0    &{\bf06.1$\pm$1.8}       &12.0$\pm$2.1           &12.7$\pm$2.3           &15.4$\pm$1.9               &17.7$\pm$3.1             &15.7$\pm$1.6               &25.4$\pm$3.5\\ \hline
    $\uparrow$   ap       &85.6$\pm$4.6    &{\bf83.9$\pm$3.8}       &73.3$\pm$2.3           &73.6$\pm$4.0           &71.9$\pm$3.4               &64.3$\pm$9.1             &68.9$\pm$4.4               &44.5$\pm$5.0\\ \hline
    $\downarrow$ oe       &09.6$\pm$8.1    &{\bf11.1$\pm$7.6}       &{\bf14.6$\pm$9.9}      &{\bf09.6$\pm$9.2}      &{\bf11.0$\pm$9.3}          &30.0$\pm$27.3            &20.7$\pm$13.0              &64.5$\pm$9.2\\ \hline
    $\downarrow$ co       &24.0$\pm$4.9    &{\bf24.8$\pm$6.0}       &34.6$\pm$5.8           &37.2$\pm$6.5           &42.3$\pm$6.3               &41.7$\pm$3.7             &39.7$\pm$4.1               &49.0$\pm$3.2\\ \hline
    \multicolumn{9}{|l|}{Voc12} \\\hline
    $\downarrow$ hl       &10.5$\pm$0.8    &10.2$\pm$0.7            &12.5$\pm$0.9           &{\bf09.3$\pm$0.5}      &{\bf09.2$\pm$0.8}          &09.5$\pm$0.7             &{\bf09.2$\pm$0.7}          &09.5$\pm$0.7\\ \hline
    $\downarrow$ rl       &15.5$\pm$1.4    &{\bf15.8$\pm$1.1}       &18.3$\pm$1.6           &17.0$\pm$1.6           &22.7$\pm$1.6               &19.4$\pm$2.3             &{\bf15.6$\pm$1.9}          &23.6$\pm$1.8\\ \hline
    $\uparrow$ ap         &64.4$\pm$2.8    &{\bf63.8$\pm$1.3}       &60.7$\pm$1.8           &62.5$\pm$2.3           &58.9$\pm$2.6               &61.7$\pm$2.8             &{\bf65.0$\pm$2.6}          &55.4$\pm$2.4\\ \hline
    $\downarrow$ oe       &25.0$\pm$6.7    &{\bf26.1$\pm$4.1}       &30.4$\pm$3.1           &{\bf25.8$\pm$4.7}      &{\bf25.2$\pm$4.6}          &28.0$\pm$5.6             &{\bf25.5$\pm$4.6}          &28.6$\pm$5.3\\ \hline
    $\downarrow$ co       &31.9$\pm$2.2    &{\bf32.6$\pm$2.3}       &35.5$\pm$2.8           &34.1$\pm$2.9           &45.0$\pm$3.0               &37.8$\pm$2.6             &{\bf32.2$\pm$3.4}          &44.9$\pm$2.5\\ \hline
    \multicolumn{9}{|l|}{50Salad} \\\hline
    $\downarrow$ hl       &16.5$\pm$2.1    &{\bf18.0$\pm$3.4}       &{\bf20.3$\pm$4.2}      &{\bf19.0$\pm$5.0}      &{\bf17.6$\pm$5.0}          &23.6$\pm$4.8             &24.6$\pm$3.8               &41.0$\pm$2.8\\ \hline
    $\downarrow$ rl       &07.9$\pm$3.0    &{\bf07.6$\pm$4.4}       &11.1$\pm$5.7           &{\bf09.0$\pm$3.3}      &{\bf07.5$\pm$2.8}          &19.3$\pm$5.8             &20.5$\pm$5.2               &54.4$\pm$7.0\\ \hline
    $\uparrow$ ap         &91.9$\pm$5.3    &{\bf92.1$\pm$5.4}       &{\bf88.7$\pm$5.6}      &{\bf90.9$\pm$3.4}      &{\bf91.9$\pm$3.6}          &81.8$\pm$4.5             &80.6$\pm$5.0               &52.8$\pm$6.9\\ \hline
    $\downarrow$ oe       &11.0$\pm$12.8   &{\bf10.9$\pm$10.5}      &{\bf13.9$\pm$9.8}      &{\bf09.1$\pm$6.0}      &{\bf09.8$\pm$8.1}          &20.4$\pm$8.6             &22.9$\pm$7.7               &62.4$\pm$13.7\\ \hline
    $\downarrow$ co       &27.9$\pm$4.7    &{\bf27.7$\pm$5.5}       &31.5$\pm$5.0           &29.5$\pm$5.5           &{\bf28.1$\pm$4.6}          &37.6$\pm$3.9             &37.6$\pm$4.8               &61.1$\pm$4.1\\ \hline
    \end{tabular}
\caption{Bag level measurements (percentage) for methods on datasets with instance labels. $\downarrow$ ($\uparrow$) next to a metric indicates that the performance improves when the metric is decreased (increased). The proposed approach and values that are statistically indistinguishable using two-tailed paired $\textit{t}$-tests at 95\% confidence level with the optimal performances are bolded. Note that the SLR approach is only used as a reference since instance level labels are provided in training.}
\label{table:6}
\end{table}

\textbf{Setting.} Even though this paper focuses on instance label prediction, we also consider examining the application of the proposed method for bag label prediction. We perform experiments on datasets with instance level label such as HJA, MSCV2, Letter Carroll, Letter Frost, Voc12, and 50Salad following the method described in Section \ref{ss:blp}. We compare MLR with the following methods: SIM, Mfast, LSB, M-SVM, and M-NN. Moreover, we also compare MLR with the logistic regression trained in the SISL setting (SLR) and a dummy classifier designed to optimize the performance for each of the following evaluation metrics: Hamming loss, ranking loss, average precision, one error, and coverage. Note that the dummy classifier assigns the same output for any test bag. For Hamming loss, the dummy classifier outputs a bag label consisting of all classes which appear in more than 50\% of training bags. For ranking loss, average precision, one error, and coverage, with each class $c$, the dummy classifier outputs $f_{dummy}(\textbf{X}_t,c)=$ (the percentage of training bags having class $c$) indicating the confidence of the test bag $\textbf{X}_t$ having class $c$. Note that $f_{dummy}(\textbf{X}_t,c)$ is independent of $\textbf{X}_t$. To compute ranking loss, average precision, one error, and coverage from the confidence values, we follow \cite{zhou2012multi}. For MLR, SIM, and LSB, the confidence for the bag on each class is computed as the maximal value of the confidence of its instances w.r.t.~the class. Specifically, $f_{MLR}(\textbf{X}_t,c)=\max_{i}p(\textbf{y}_{ti}=c|\textbf{x}_{ti},\textbf{w})$, $f_{LSB}(\textbf{X}_t,c)=\max_{i}p(\textbf{y}_{ti}=c|\textbf{x}_{ti},\textbf{w},\hat{\alpha})$, and $f_{SIM}(\textbf{X}_t,c)=\max_{i}\textbf{w}_c^{T}\textbf{x}_{ti}$. For Mfast, M-SVM, and M-NN, the bag level metrics are computed as in \cite{huang2013fast,zhou2007multi,zhang2007multi}, respectively. The results are reported in Table \ref{table:6}.

\noindent \textbf{Results.} From Table \ref{table:6}, we observe that MLR matches or outperforms other methods in term of each evaluation measure in almost all datasets, except for the MSCV2 where the accuracy of MLR is lower than those of M-NN and M-SVM. The reason is that M-NN and M-SVM consider bag level information by encoding each bag with a vector containing the similarities of the bag with representative bags in the dataset. However, instance annotation methods including MLR, SIM, LSB, and Mfast only focus on instance level features without including additional bag level information. Consequently, approaches that rely on bag level information may outperform instance level approaches on datasets where each test bag information is helpful to predict its instance labels such as image datasets including MSCV2. For example, we are not sure whether a single white color pixel is from the sky or a blank paper. However, if we know that the overall image is about a scene, we would rather predict the pixel is from the sky. Similar to the instance annotation accuracy in Table \ref{table:2}, our approach performance is close to the SLR version. Among the considered instance annotation methods, MLR seems to outperform other methods in bag level measurements. Among the bag level methods, M-NN seems to outperform M-SVM.

\subsection{Kernel OR-LR}
\label{ss:kernel}
\textbf{Setting.} In kernel learning, we transform feature vector $\textbf{x}$ to $\textbf{k}(\textbf{x})$ as in Section \ref{ss:kbolrm}. We refer to the setting which involves $\textbf{x}$ (as in \eqref{e:prior}) as feature learning and to the setting which $\textbf{x}$ is replaced by $\textbf{k}(\textbf{x})$ as kernel learning. The parameter $\delta$ in \eqref{e:kernele} is selected as follows. We first compute the mean square distance among every pair of instances in the dataset $\overline{d^2}$. Then, $\delta$ is computed as $\delta=s\times \overline{d^2}$, where $s$ is in $\{0.1,0.2,0.5,1,2,5,10\}$.

We evaluate the instance level accuracy as a function of $\delta$ for SLR and MLR and report the highest accuracy for each method. Since the dimension of $\textbf{w}$ in kernel learning is higher than in feature learning, we set the number of expectation maximization iterations to $500$. For SIM and Mfast, we add $\delta$ to the parameters used for tuning in Section \ref{ss:gs}. In SIM, the number of iterations in each phase $T$ is set to 1,000 due to the higher dimension in kernel space. Similarly, in Mfast, the number of iterations is set to 100. Due to the new parameter $s$, the large number of parameter settings, and the higher number of iterations leading to a longer runtime compared to the feature learning case, the parameters of Mfast are selected as follows. Since the performance of Mfast depends mainly on $K$ \citep{huang2013fast}, $K$ is searched over the set $\{1, 5, 10, 15\}$ and $m,C,\gamma_0,\eta$ are set to $100,10,10^{-3},10^{-5}$, respectively. In this set of experiments, due to significant increase in runtime of the kernel learning setting, we do not include LSB. The results for kernel learning of MLR, SLR, SIM, and Mfast are reported in Table \ref{table:8}.

\noindent \textbf{Results.} From Table \ref{table:8}, on MSCV2, Letter Carroll, and Letter Frost, there is a significant improvement of accuracy in kernel case compared to the feature case in both SLR and MLR. Specifically,  on kernel learning, the instance annotation accuracy of MLR is 5\%, 4\%, and 3\% higher on MSCV2, Letter Carroll, and Letter Frost, compared to the feature case, respectively. Moreover, the accuracy of SLR is also significantly higher 5\%, 9\%, and 6\% on those datasets, compared to the feature case, respectively. Since the dummy classifier does not rely on instance feature vectors, the accuracy is unchanged compared to the feature case as presented in Table \ref{table:2}. In the kernel learning case, MLR outperforms Mfast which seems to outperform SIM. We note that due to the higher number of iterations and larger size parameters, running the kernel version of methods takes longer than the feature version. For instance, running the kernel version of MLR on Letter Carroll takes 7 times longer than the feature version whereas running the kernel version of SIM on Letter Carroll takes 15 times longer than the feature version.

\begin{table}
\fontsize{9}{12}\selectfont
\centering
    \begin{tabular}{ | l || l | l | l | l | l | l | l |}
    \hline
    Dataset       & HJA bird               & HJA union             & MSCV2                     & Carroll                      & Frost                   &Voc12                  &50Salad\\ \hline\hline
    {\bf K-MLR} & {\bf70.2$\pm$5.1}      & {\bf72.8$\pm$3.7}     & {\bf61.0$\pm$5.3}         & {\bf72.1$\pm$5.7}            &{\bf74.0$\pm$5.5}        &{\bf45.0$\pm$3.0}      &{\bf78.4$\pm$5.1}\\ \hline\hline
    K-Mfast     & 62.1$\pm$3.0           & 63.8$\pm$3.9          & 53.7$\pm$4.9              & 68.6$\pm$4.5                 &67.0$\pm$5.1             &37.7$\pm$3.6           &74.2$\pm$6.6\\ \hline
    K-SIM       & 63.4$\pm$3.0           & 63.8$\pm$3.2          & 51.2$\pm$6.6              & 65.9$\pm$3.9                 &64.9$\pm$7.8             &42.1$\pm$4.1           &71.9$\pm$6.2\\ \hline\hline
    K-SLR       & 77.6$\pm$3.6           & 77.6$\pm$3.6          & 65.1$\pm$4.1              & 79.5$\pm$5.2                 &79.5$\pm$3.5             &46.6$\pm$2.5           &83.7$\pm$3.5\\ \hline
    K-Dummy     & 12.1$\pm$3.4           & 12.1$\pm$3.4          & 14.5$\pm$3.7              & 11.2$\pm$3.7                 &9.3$\pm$2.4              &36.1$\pm$4.1           &16.7$\pm$7.7\\ \hline
    \end{tabular}
\caption{Accuracy results (percentage) for instance label prediction in inductive mode for kernel version of MLR, Mfast, SIM, LSB, and Dummy approaches denoted by K-MLR, K-Mfast, K-SIM, K-SLR, and K-Dummy, respectively. The proposed approach and values that are statistically indistinguishable using two-tailed paired $\textit{t}$-tests at 95\% confidence level with the optimal performances are bolded.}
\label{table:8}
\end{table}

\subsection{Speed up}
In this section, we study techniques to reduce the computational complexity of our approach. The main time-consuming part of MLR is in the E-step. Table \ref{table:9} shows the computational complexity per bag for each of our E-step implementation and the sum of each factor in that computational complexity over all bags on every dataset. Since all implementations have an exponential term regarding the number of classes per bag $2^{|\textbf{Y}_b|}$, speeding up is necessary for datasets such as Letter Carroll or Letter Frost, where the factor $\sum_{b=1}^{B}|\textbf{Y}_b|2^{|\textbf{Y}_b|}n_b^2$ significantly dominates the factor $\sum_{b=1}^{B}n_bCd$, as illustrated in Table \ref{table:9}, because of several bags having a large number of classes.

\begin{table}
\centering
    \begin{tabular}{ | l || l | l | l | l |}
    \hline
   Algorithm            & \multicolumn{2}{l}{Forward}                                             & \multicolumn{2}{|l|}{Forward and substitution}           \\\hline
   Complexity per bag   & \multicolumn{2}{l}{$O(|\textbf{Y}_b|2^{|\textbf{Y}_b|}n_b^2+n_bCd)$}    & \multicolumn{2}{|l|}{$O(|\textbf{Y}_b|2^{|\textbf{Y}_b|}n_b+n_bCd)$} \\ \hline\hline
   Dataset              & $\sum_{b=1}^{B}|\textbf{Y}_b|2^{|\textbf{Y}_b|}n_b^2$ &$\sum_{b=1}^{B}n_bCd$ &$\sum_{b=1}^{B}|\textbf{Y}_b|2^{|\textbf{Y}_b|}n_b$ &$\sum_{b=1}^{B}n_bCd$ \\\hline
   HJA bird             & 1,300,502     &2,469,012          &99,314             &2,469,012          \\\hline
   HJA union            & 1,105,808     &2,469,012          &83,584             &2,469,012          \\\hline
   MSCV2                & 680,112       &1,940,832          &105,584            &1,940,832          \\\hline
   Letter Carroll       & 4,825,522     &275,328            &499,282            &275,328            \\\hline
   Letter Frost         & 1,921,792     &216,960            &211,248            &216,960            \\\hline
   Voc12                & 566,464       &3,976,320          &79,888             &3,976,320          \\\hline
   50Salad              & 4,981,778     &690,840            &120,738            &690,840            \\\hline
    \end{tabular}
\caption{Computational complexity in the E-step for the two implementations of MLR for every dataset}
\label{table:9}
\end{table}

\subsubsection{Speed up using pruning}
\begin{figure}
\begin{minipage}[b]{0.48\linewidth}
  \centering
  \centerline{\includegraphics[width=7.4cm]{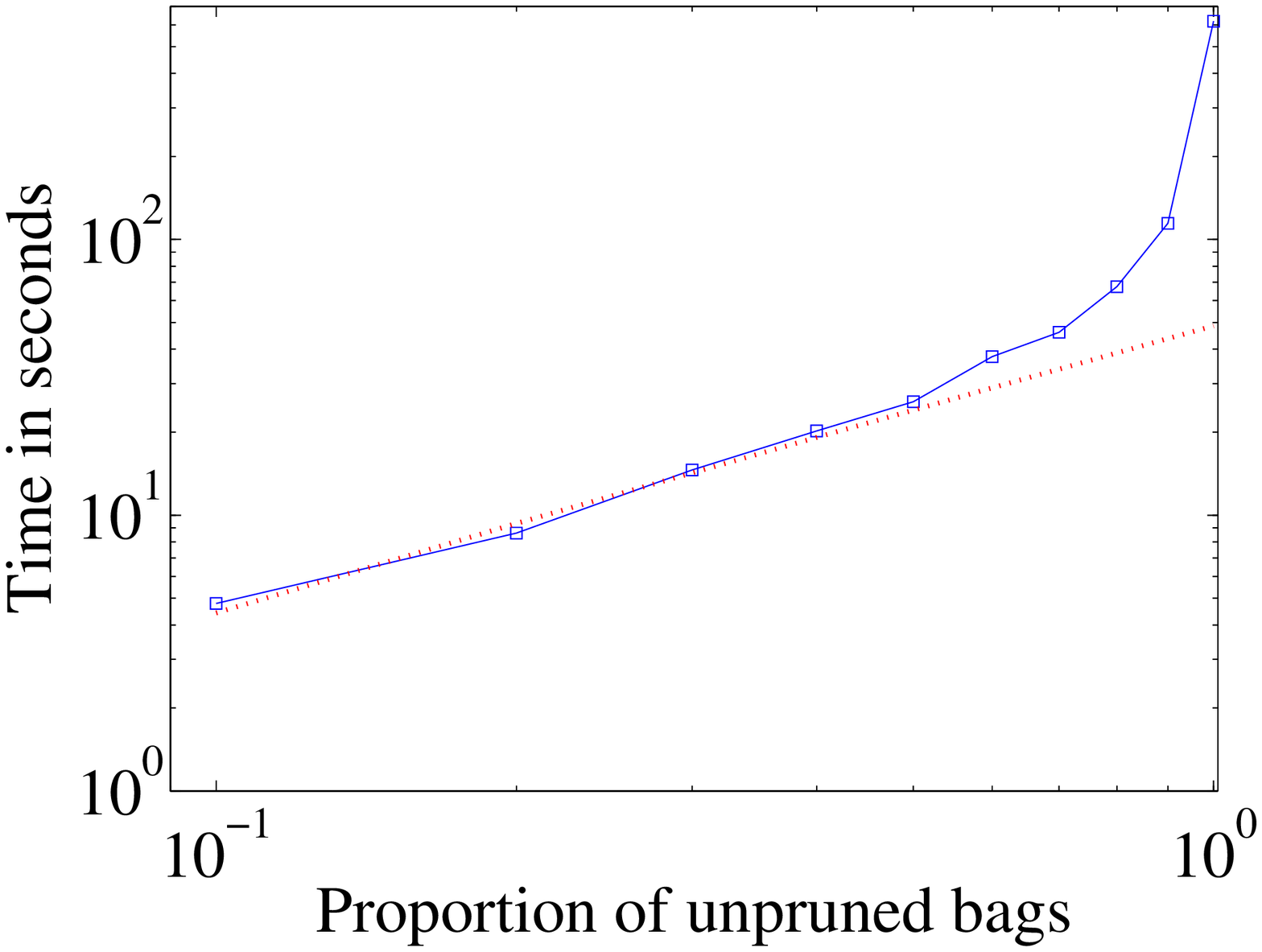}}
  \centerline{Letter Carroll}\medskip
\end{minipage}
\hfill
\begin{minipage}[b]{0.48\linewidth}
  \centering
  \centerline{\includegraphics[width=7.4cm]{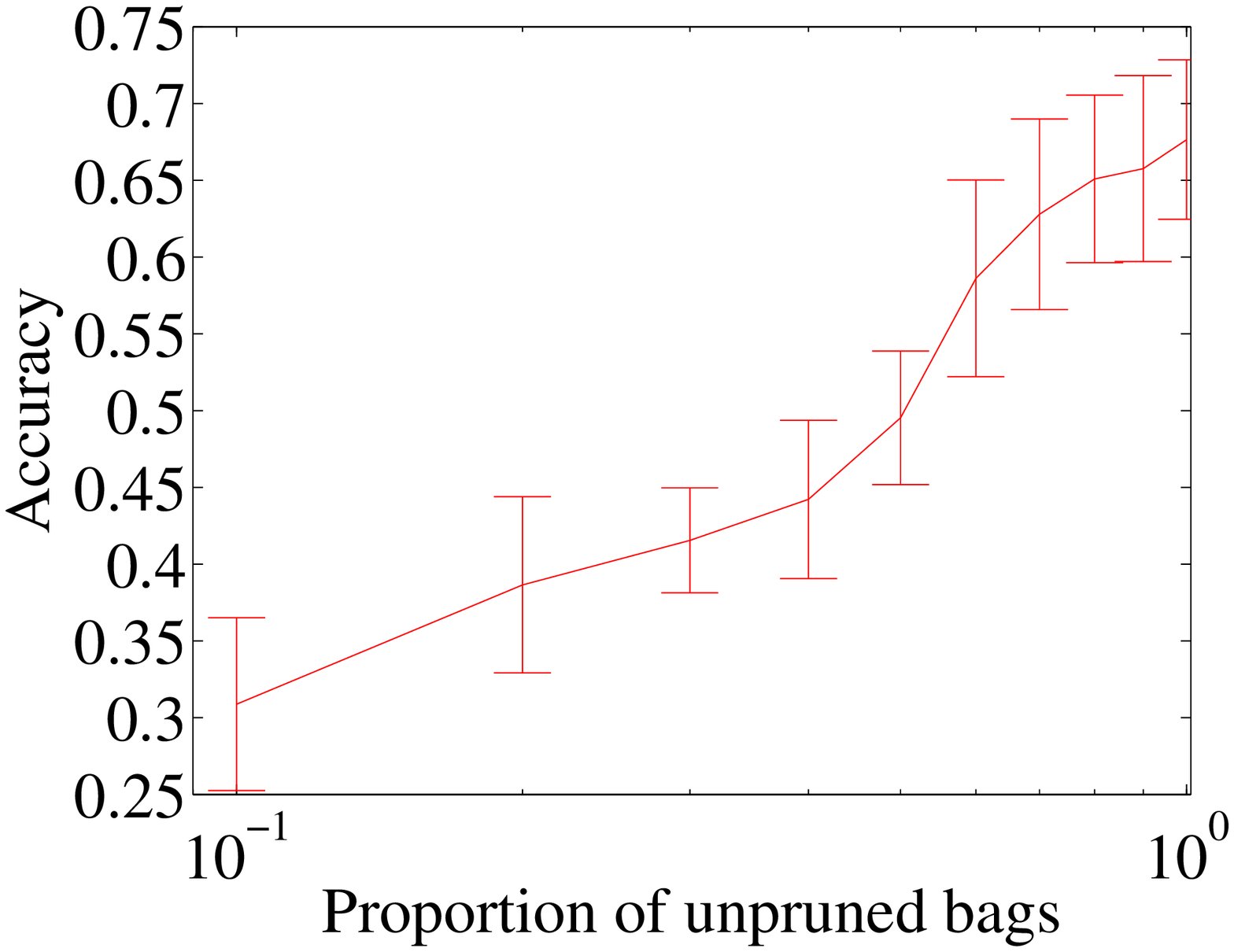}}
  \centerline{Letter Carroll}\medskip
\end{minipage}
\begin{minipage}[b]{0.48\linewidth}
  \centering
  \centerline{\includegraphics[width=7.4cm]{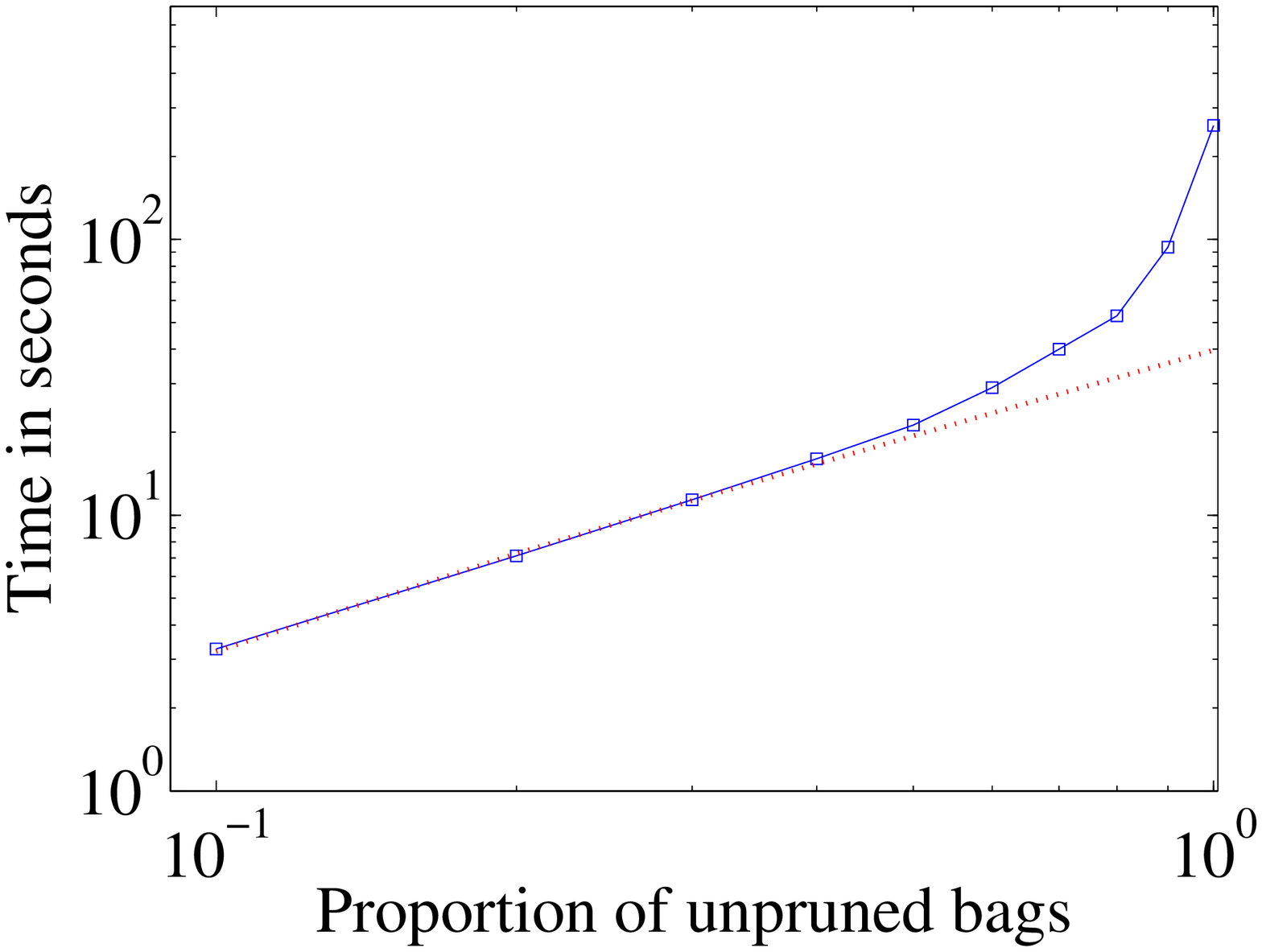}}
  \centerline{Letter Frost}\medskip
\end{minipage}
\hfill
\begin{minipage}[b]{0.48\linewidth}
  \centering
  \centerline{\includegraphics[width=7.4cm]{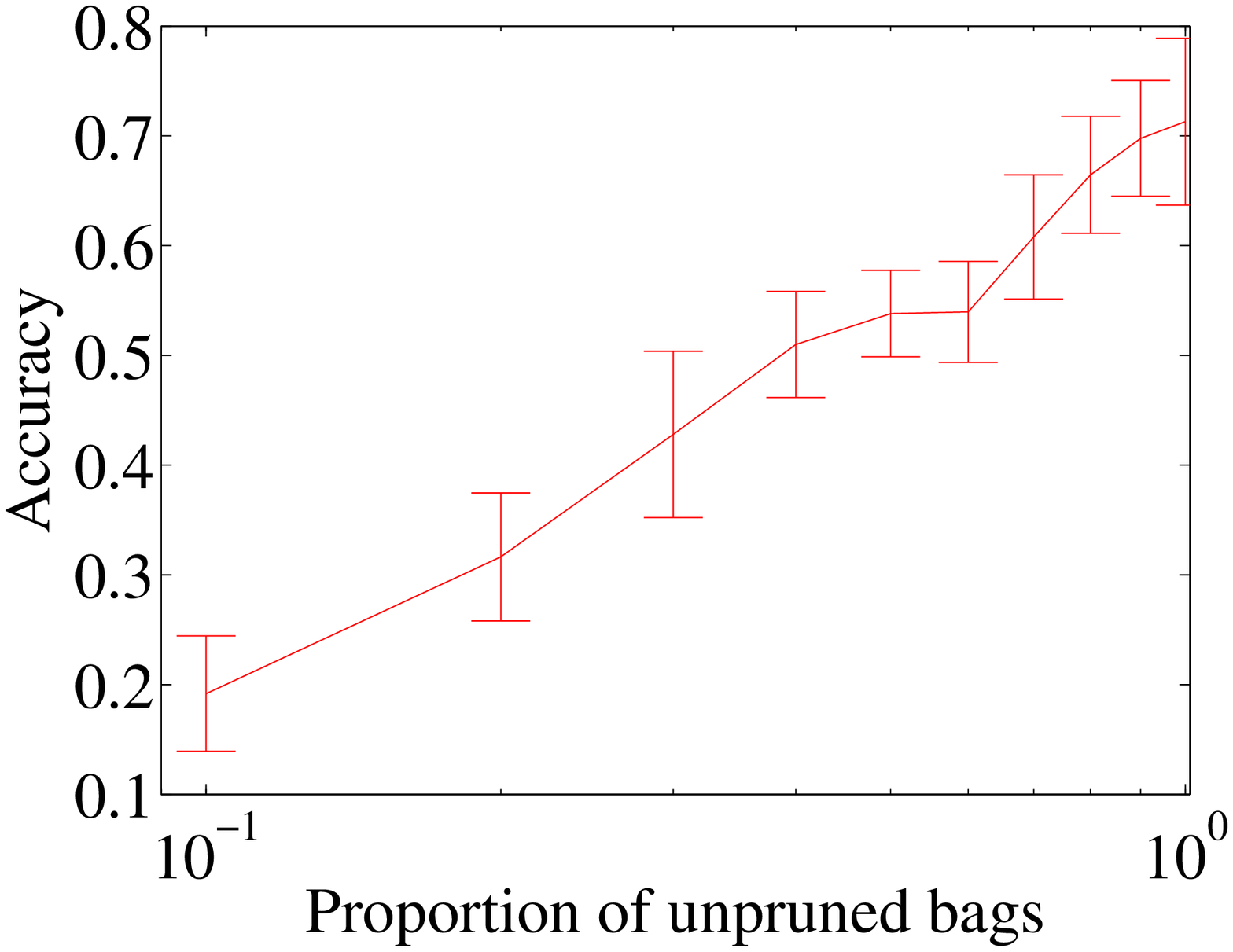}}
  \centerline{Letter Frost}\medskip
\end{minipage}

\begin{minipage}[b]{0.48\linewidth}
  \centering
  \centerline{\includegraphics[width=7.4cm]{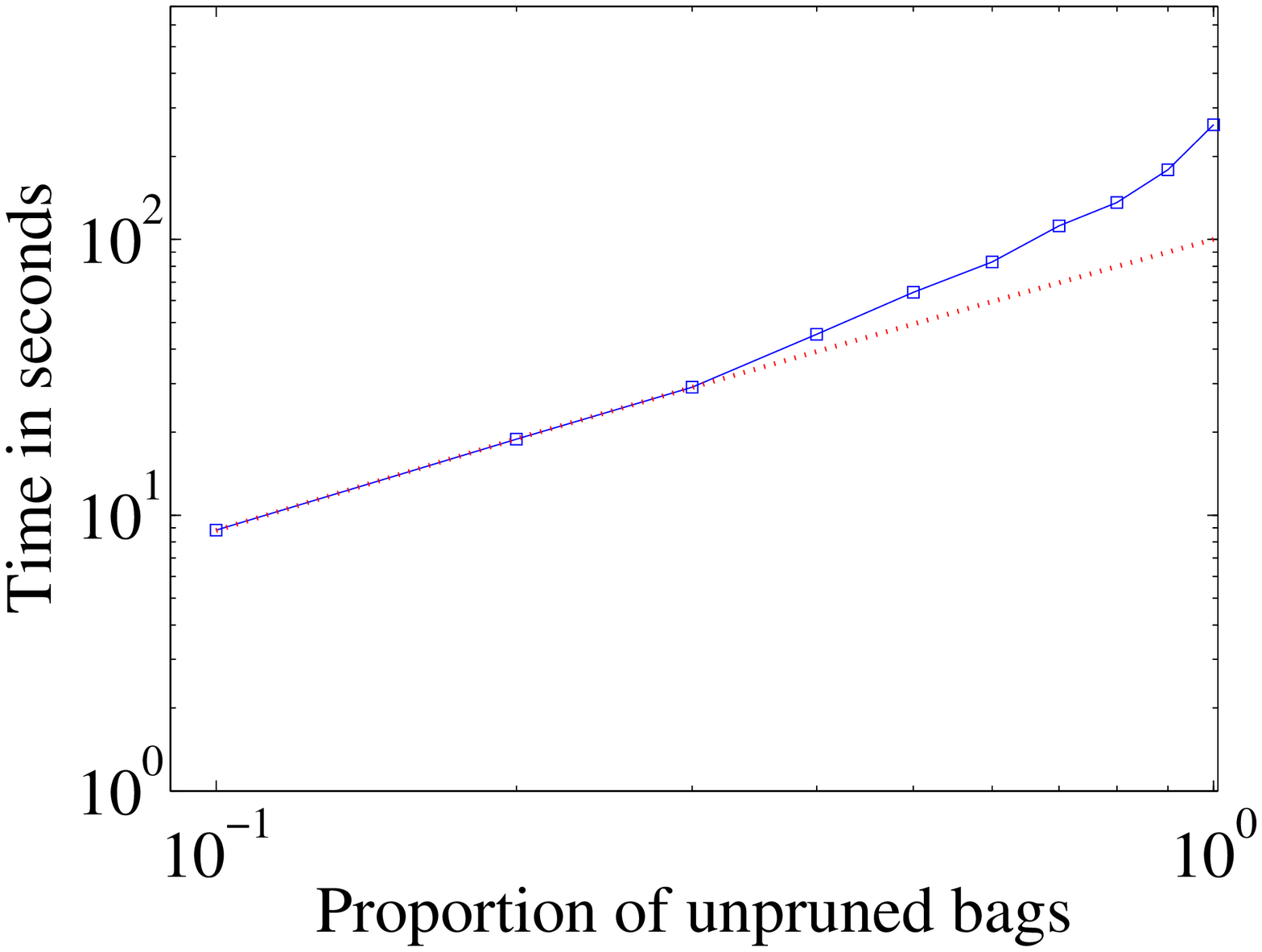}}
  \centerline{HJA bird}\medskip
\end{minipage}
\hfill
\begin{minipage}[b]{0.48\linewidth}
  \centering
  \centerline{\includegraphics[width=7.4cm]{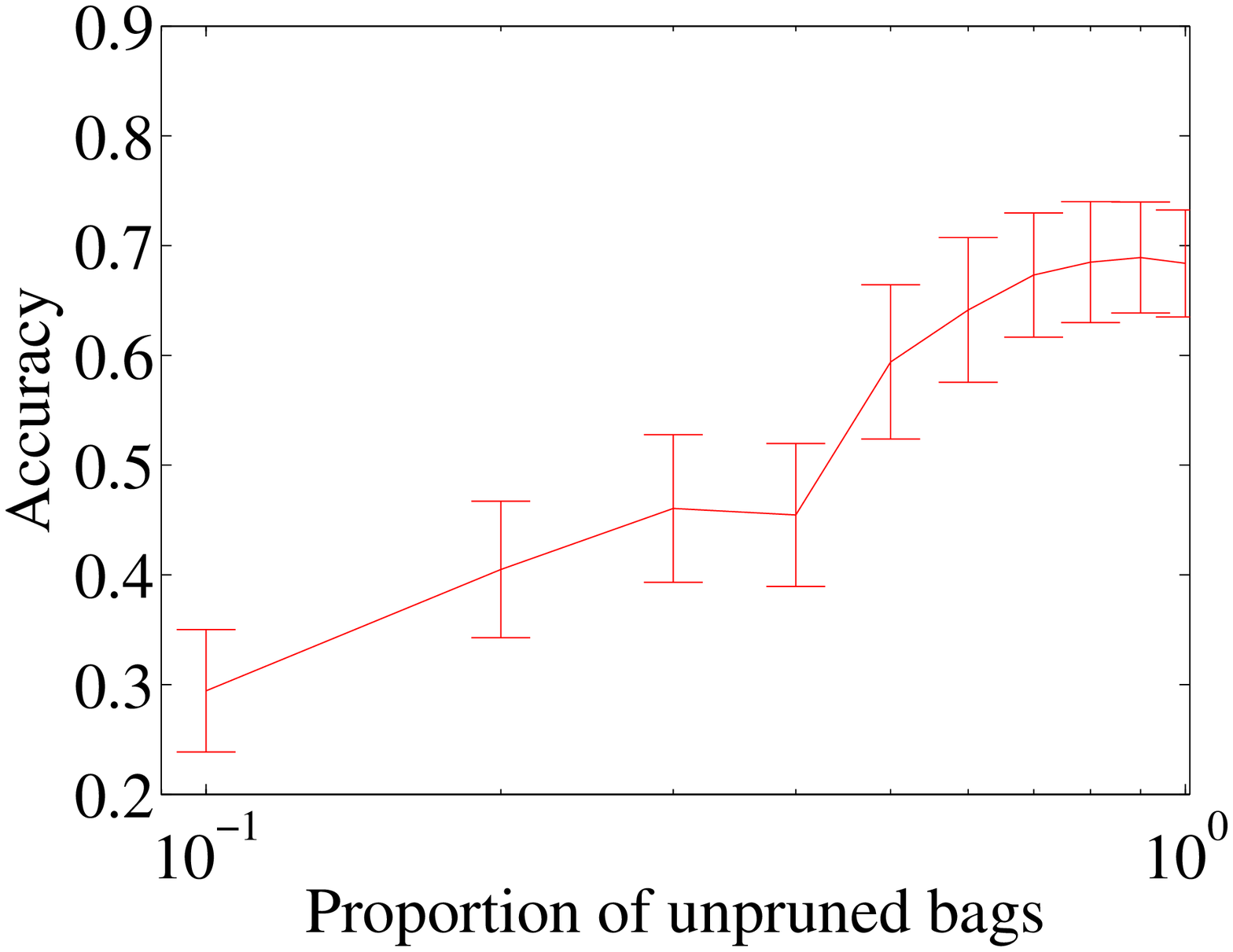}}
  \centerline{HJA bird}\medskip
\end{minipage}
\caption{Accuracy and runtime when pruning bags with largest bag label cardinality. Dotted lines are estimated using the linear least square from the first three points starting from the left in each time plot. The difference between the solid and dotted lines at the right of each time plot shows how significant bags having a large number of classes contribute to the overall runtime.}
\label{fig:prune}
\end{figure}

Pruning is a technique in our paper that removes bags containing a large number of classes in the dataset. In order to prune, we sort bags based on the value of $n_b|\textbf{Y}_b|2^{|\textbf{Y}_b|}$ in an ascending order and remove a fixed percentage of bags with these highest values from the training set. We perform 10-fold cross validation. Assume that $n_b$ is equal for every bag, thus a bag contains a large number of classes $|\textbf{Y}_b|$ would have a significantly high value of $n_b|\textbf{Y}_b|2^{|\textbf{Y}_b|}$ proportional to the time MLR spends on the $b$th bag in the expectation step. In fact, those bags are less informative, due to the label ambiguity, which increases as a function of the size of the label set. As a result, we expect the accuracy is hardly changed after pruning while the runtime decreases.

To understand the intuition behind the pruning technique used for MLR in this section, we present the following experiments on HJA, MSCV2, Letter Carroll, and Letter Frost datasets. First, we sort the bags in term of $n_b|\textbf{Y}_b|2^{|\textbf{Y}_b|}$ in an ascending order. Then, we select $B_0$ bags which contain 80\% of the bags with lowest values, and report the ratio $(\sum_{b=1}^{B}n_b|\textbf{Y}_b|2^{|\textbf{Y}_b|})/(\sum_{b=1}^{B_0}n_b|\textbf{Y}_b|2^{|\textbf{Y}_b|})$ for each dataset as Table \ref{table:10}. From Table \ref{table:10}, we could observe that the $B_0$ selected bags only constitute to a small proportion of the sum, which are around 1/3, 1/7, 1/20, and 1/12 for HJA, MSCV2, Letter Carroll, and Letter Frost, respectively.

\begin{table}
\centering
    \begin{tabular}{ | m{3cm} || l | l | l | l |}
    \hline
   Dataset           & HJA bird                        & MSCV2                   & Letter Carroll                 & Letter Frost          \\\hline
   $\frac{\sum_{b=1}^{B}n_b|\textbf{Y}_b|2^{|\textbf{Y}_b|}}{\sum_{b=1}^{B_0}n_b|\textbf{Y}_b|2^{|\textbf{Y}_b|}}$           & 3.3                         & 6.6                     & 20.3                           & 11.6  \\\hline
    \end{tabular}
\caption{The ratio between the theoretical time complexity of all bags and 80\% bags}
\label{table:10}
\end{table}

We perform the experiment on HJA, Letter Carroll, and Letter Frost. We prune \{0\%, 10\%, 20\%, 30\%, 40\%, 50\%, 60\%, 70\%, 80\%, 90\%\} the number of bags having highest value. The accuracy and runtime as functions of the proportion of keeping bags on Letter Carroll, Letter Frost, and HJA are reported in Figure \ref{fig:prune}.

From Figure \ref{fig:prune}, we observe a significant speed up on Letter Carroll and Letter Frost datasets. Specifically, we can speed up the computation time in Letter Carroll and Letter Frost by 9.2 and 5.0 times by just removing 20\% of the data and the accuracy is hardly changed. In the Letter Carroll and Letter Frost, there are few bags having 10 classes (words have more than 10 different letters). Removing these words does not affect the overall accuracy result. However, since the computational complexity is exponential in the number of classes per bag, the runtime for such bags constitutes a large percentage of the overall runtime. In the HJA dataset, the effect is less pronounced. We can maintain the accuracy by removing 20\% bags and the runtime decreases by a factor of 2. From Figure \ref{fig:prune}, for a small proportion of keeping bags, the runtime seems to decrease linearly with the number of keeping bags. It can be explained as for those datasets, after removing a high proportion of bags with high number of classes, the remaining bags almost have the same number of classes. As a result, the computational complexity per bag using the forward and substitution algorithm from Table \ref{table:9} depends on the number of instances per bag only which is assumed to be equal. Consequently, the runtime depends linearly on the number of keeping bags.

\subsubsection{Speed up using stochastic gradient ascent}
\begin{figure}
\begin{minipage}[b]{0.48\linewidth}
  \centering
  \centerline{\includegraphics[width=7.4cm]{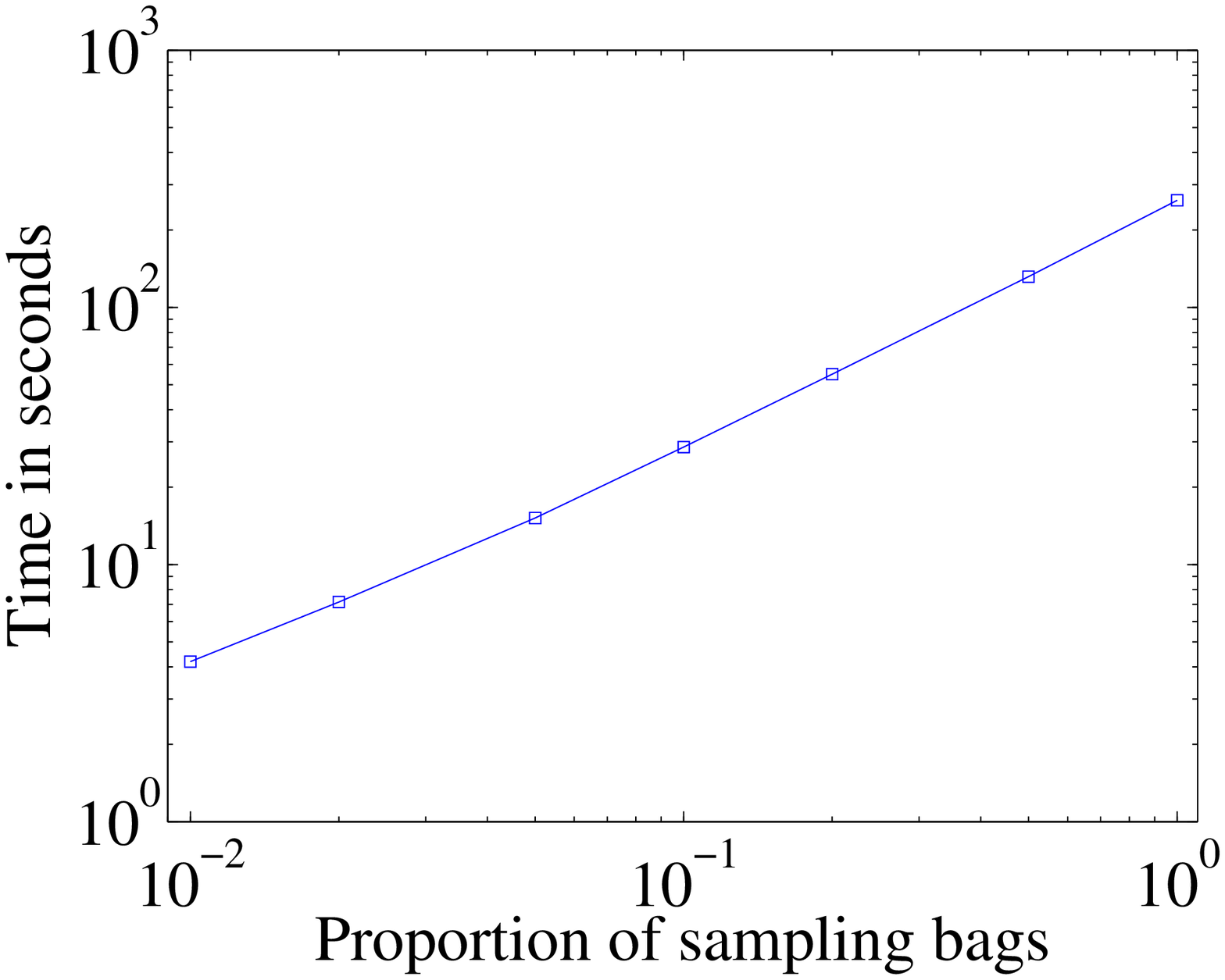}}
  \centerline{HJA bird}\medskip
\end{minipage}
\hfill
\begin{minipage}[b]{0.48\linewidth}
  \centering
  \centerline{\includegraphics[width=7.4cm]{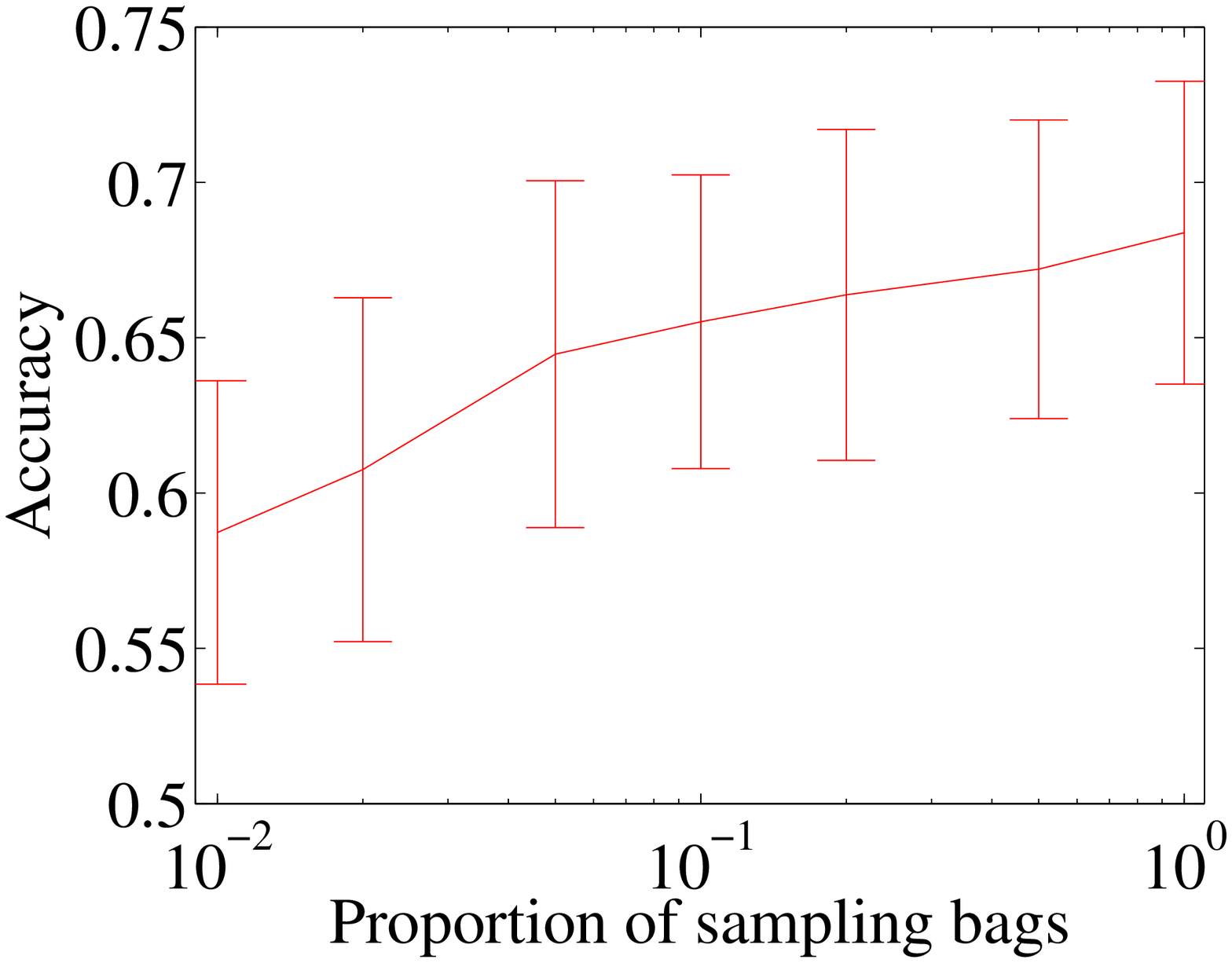}}
  \centerline{HJA bird}\medskip
\end{minipage}

\begin{minipage}[b]{0.48\linewidth}
  \centering
  \centerline{\includegraphics[width=7.4cm]{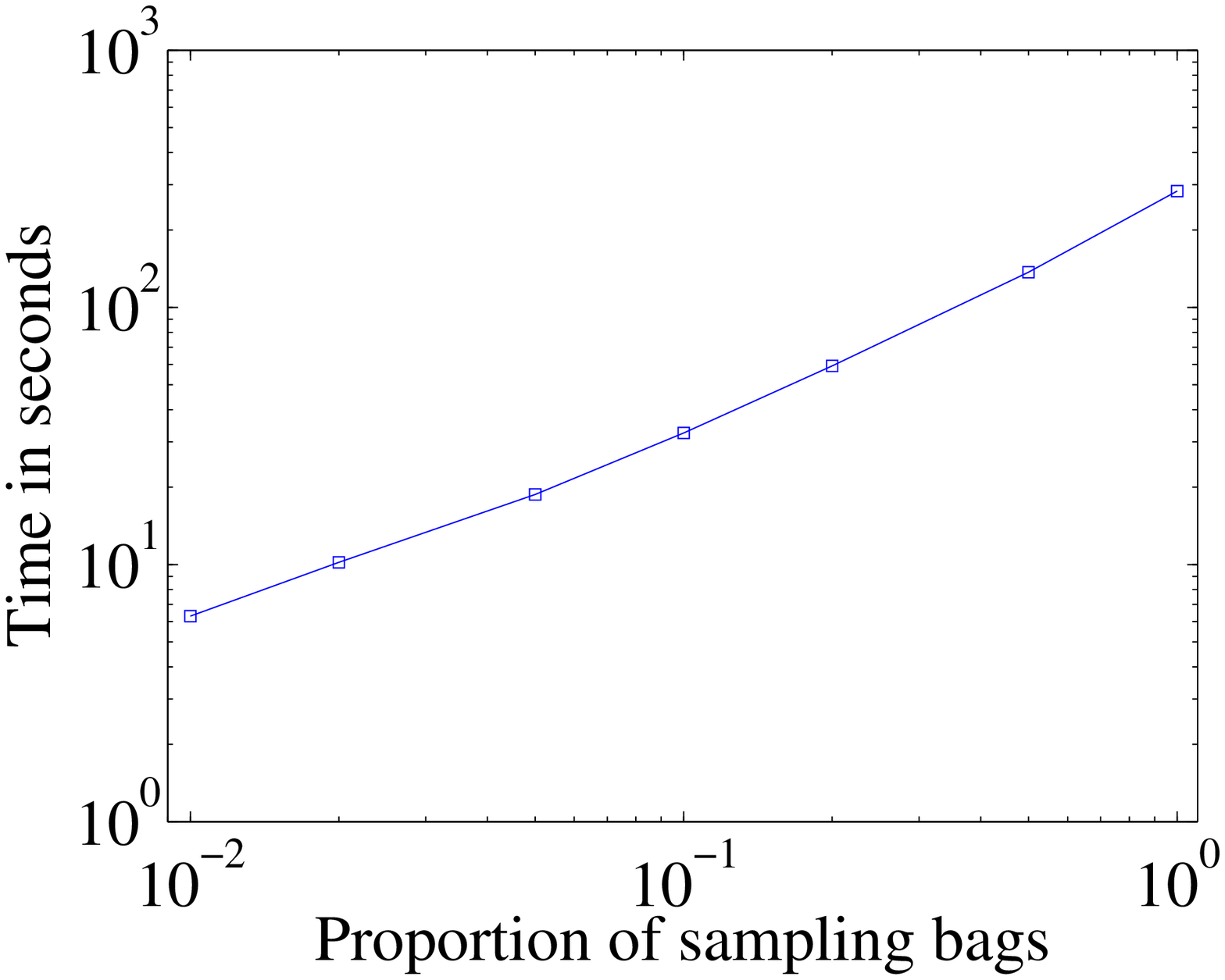}}
  \centerline{MSCV2}\medskip
\end{minipage}
\hfill
\begin{minipage}[b]{0.48\linewidth}
  \centering
  \centerline{\includegraphics[width=7.4cm]{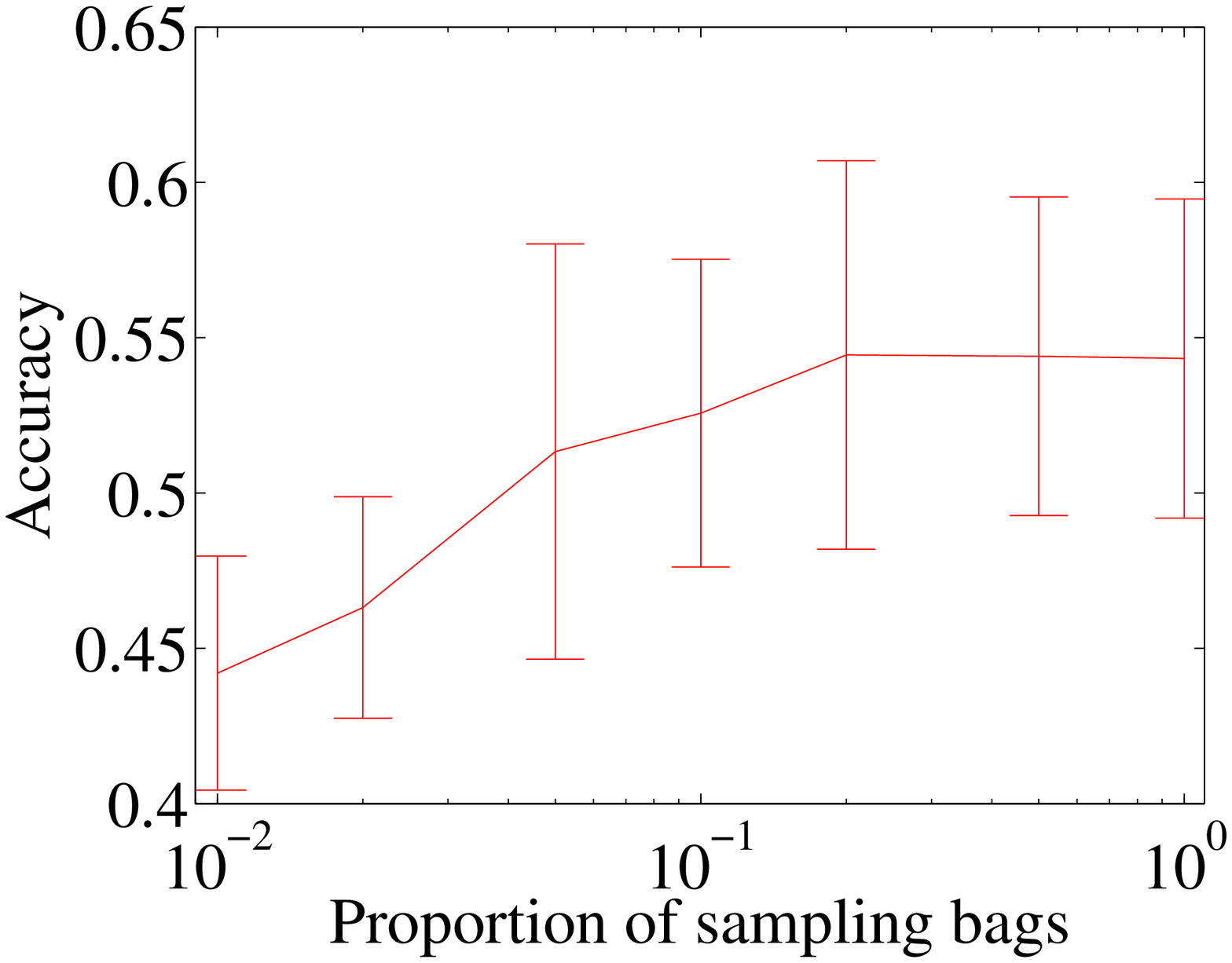}}
  \centerline{MSCV2}\medskip
\end{minipage}

\begin{minipage}[b]{0.48\linewidth}
  \centering
  \centerline{\includegraphics[width=7.4cm]{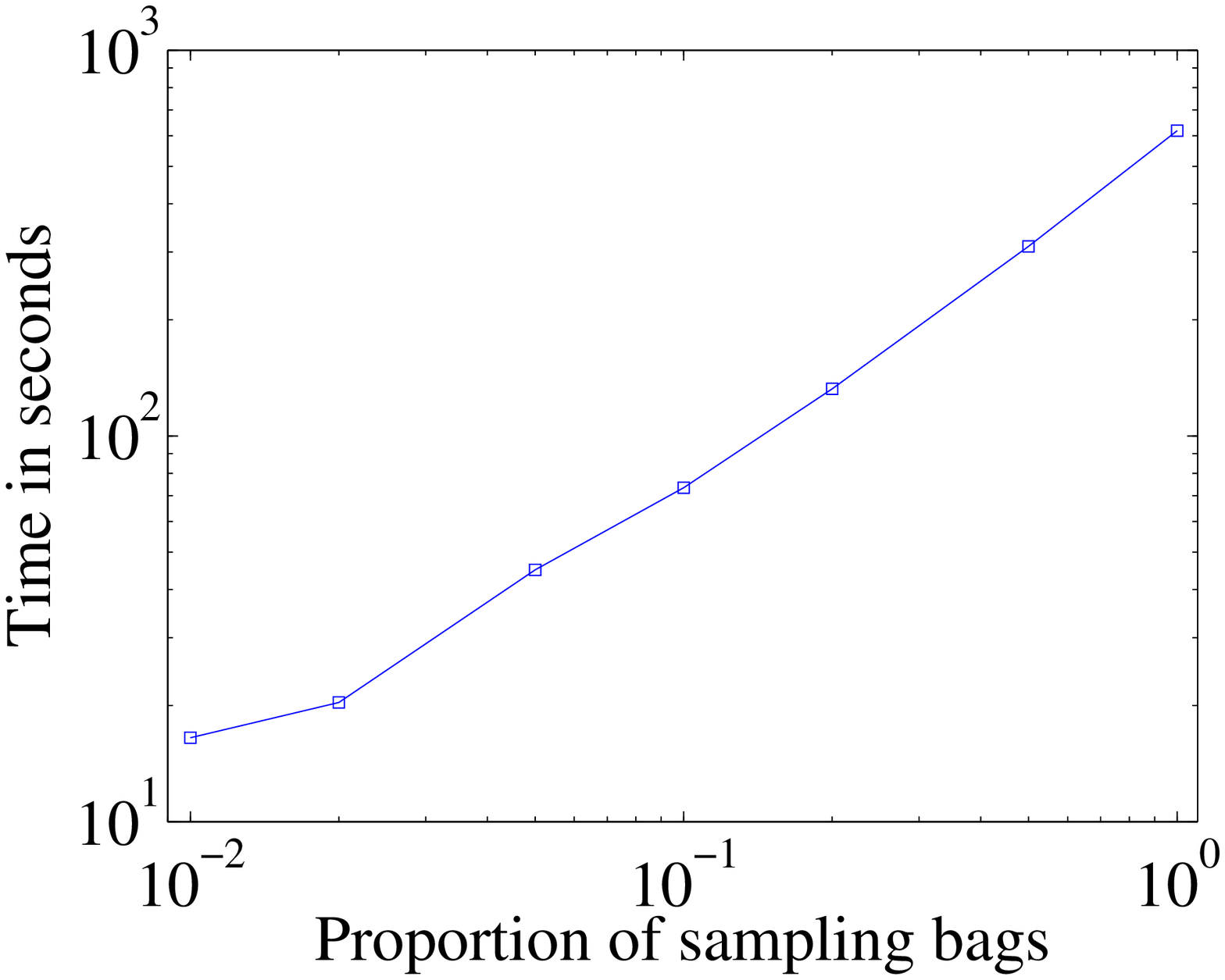}}
  \centerline{Letter Carroll}\medskip
\end{minipage}
\hfill
\begin{minipage}[b]{0.48\linewidth}
  \centering
  \centerline{\includegraphics[width=7.4cm]{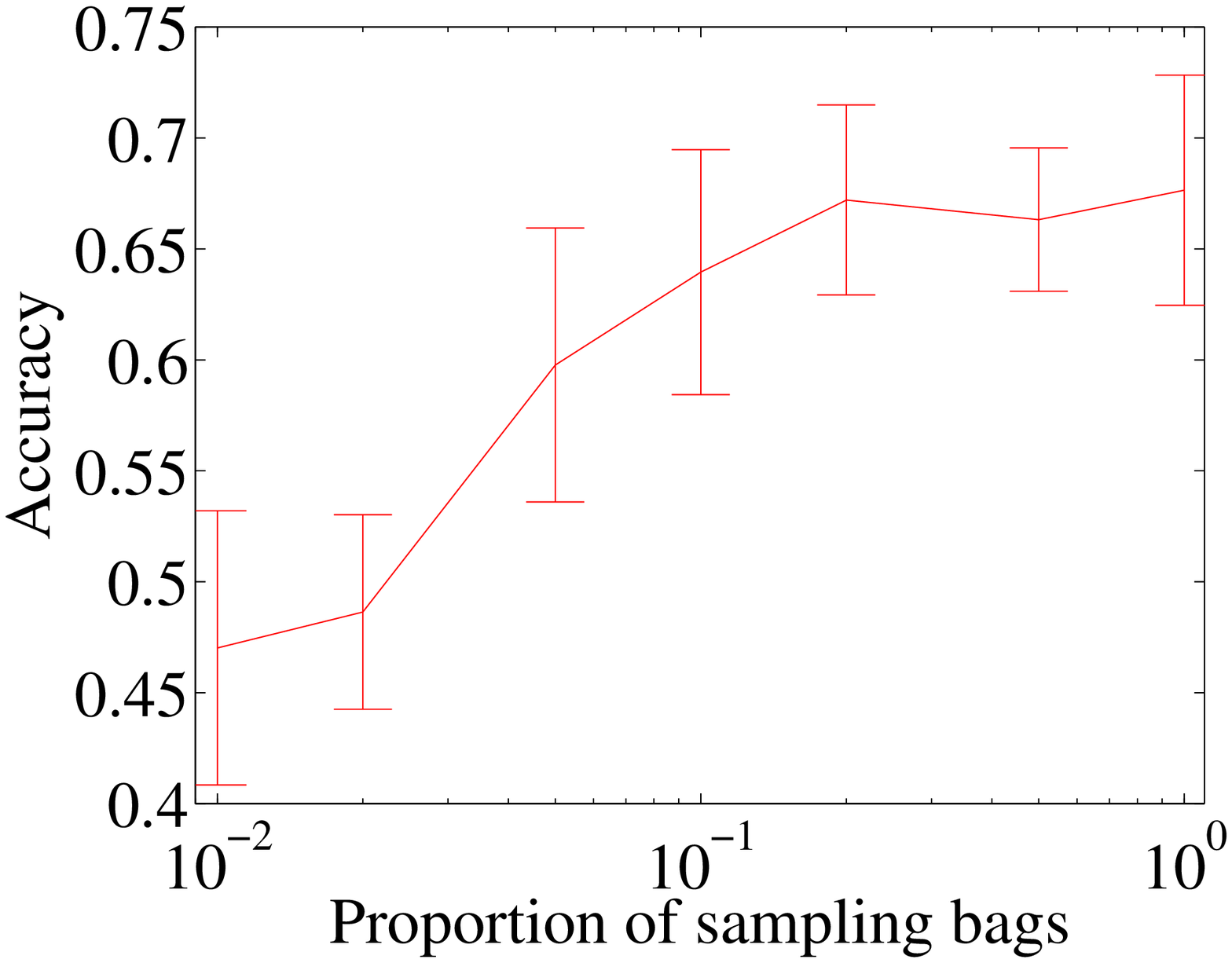}}
  \centerline{Letter Carroll}\medskip
\end{minipage}
\caption{Accuracy and runtime when using stochastic gradient ascent}
\label{fig:sampling}
\end{figure}

In Section \ref{ss:em}, we observe that in each expectation maximization iteration, for all bags, we need to compute the instance labels probability given their bag label which is costly in runtime. Stochastic sampling means that, for each iteration, we sample only a few bags then compute the instance labels probability for those bags only. Specifically, the new expectation maximization at the $k$th iteration is as follows:\\

\noindent E-step:
\begin{itemize}[leftmargin=10pt]
 \item Sample a subset $S$ from $B$ bags
 \item Compute $p(y_{bi}=c|\textbf{Y}_b,\textbf{X}_b,\textbf{w}^{(k)}), \forall b\in S$
\end{itemize}
\noindent M-step:
\begin{equation*}
\textbf{w}_c^{(k+1)}=\textbf{w}_c^{(k)}+ \frac{\partial \widetilde{g}(\textbf{w},\textbf{w}^{(k)})}{\partial\textbf{w}_c}\bigg|_{\textbf{w}=\textbf{w}^{(k)}} \times \eta,
\end{equation*}
where
\begin{equation}
\frac{\partial  \widetilde{g}(\textbf{w},\textbf{w}^{(k)})}{\partial\textbf{w}_c}=\sum_{b\in S}\sum_{i=1}^{n_b}[p(y_{bi}=c|\textbf{Y}_b,\textbf{X}_b,\textbf{w}^{(k)})-p(y_{bi}=c|\textbf{x}_{bi},\textbf{w})]\textbf{x}_{bi}.
\end{equation}

\noindent Since computing instance labels probability on a subset of bags is faster with a factor linear w.r.t.~the size of the subset, we expect that the runtime will decrease proportionally. In each iteration, the subset of bags $S$ would change, so we expect we could cover all the bags after few iterations, such that the accuracy does not change significantly.

The percentage of sampled bags is selected from the set \{1\%, 2\%, 5\%, 10\%, 20\%, 50\%, 100\%\}. The accuracy and runtime as functions of the proportion of sampled bags of our approach on HJA, MSCV2, and Letter Carroll are reported in Figure \ref{fig:sampling}.

From Figure \ref{fig:sampling}, we observe that with high percentage of sampled bags, when we decrease the percentage of sampled bags, the runtime decreases at almost the same rate. For example, the runtime for HJA dataset with sampling percentage of \{20\%, 50\%, 100\%\} are \{55, 131, 260\} seconds per cross validation, respectively. The reason is that in both expectation step and maximization step, MLR with stochastic gradient ascent only considers sampled bags. When we decrease the percentage of sampled bags, the overall runtime of MLR decreases at almost the same rate. In general, by sampling 20\% of the number of the bags, we can reduce the runtime by a factor of 5 while keeping the accuracy at the same level on HJA, MSCV2, and Letter Carroll.

\subsubsection{Speed up by shrinking the dictionary}
\begin{figure}

\begin{minipage}[b]{0.48\linewidth}
  \centering
  \centerline{\includegraphics[width=7.4cm]{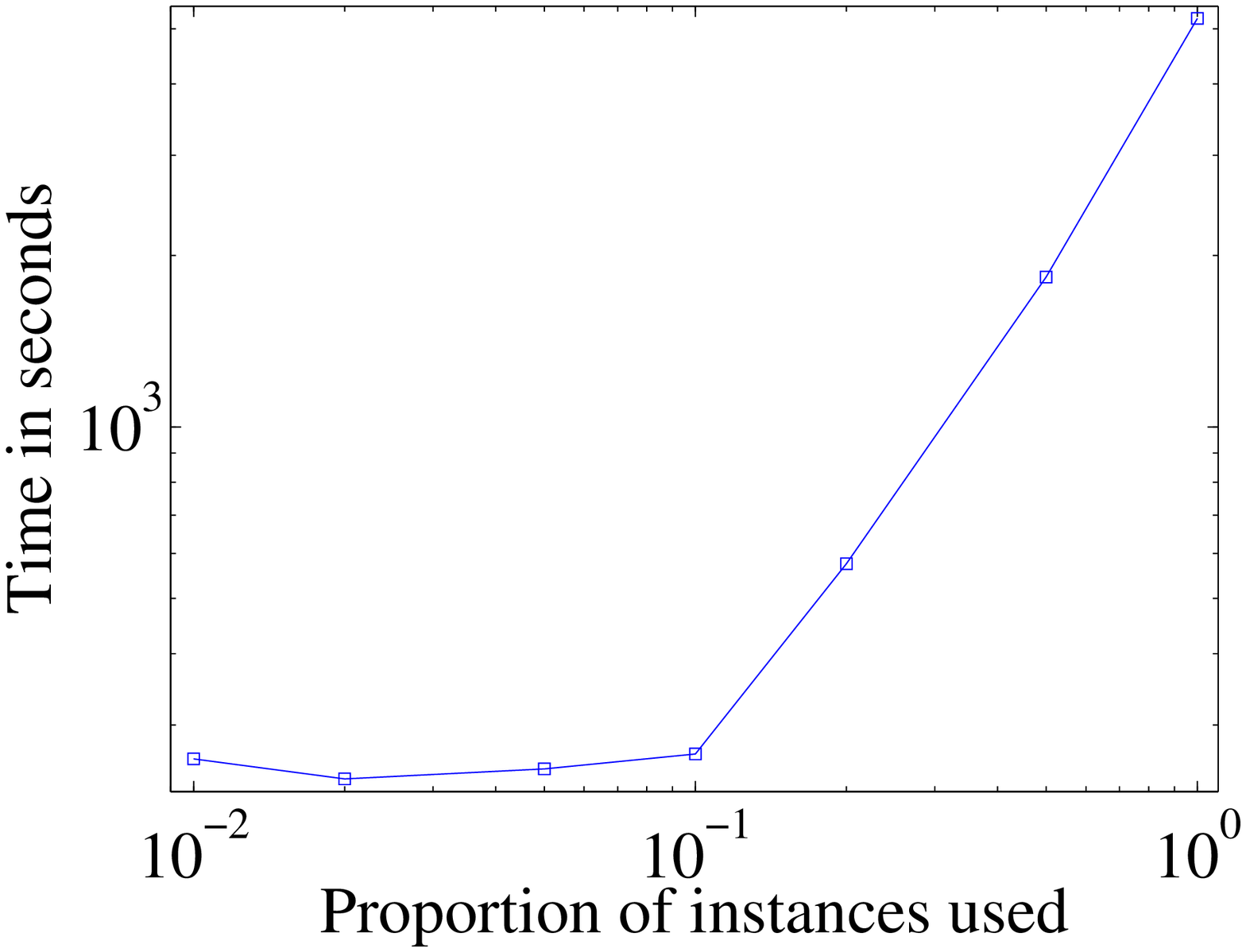}}
  \centerline{HJA bird}\medskip
\end{minipage}
\hfill
\begin{minipage}[b]{0.48\linewidth}
  \centering
  \centerline{\includegraphics[width=7.4cm]{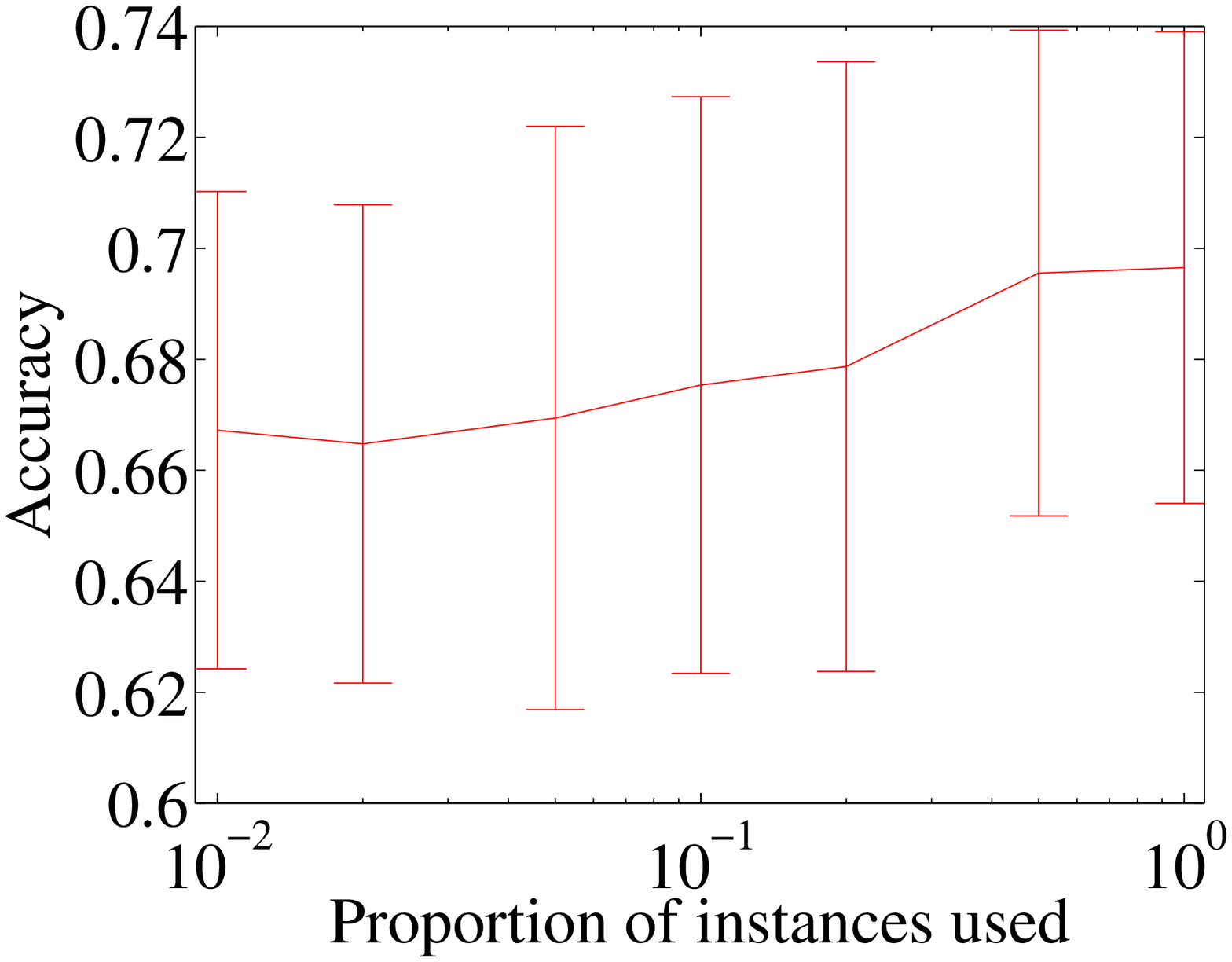}}
  \centerline{HJA bird}\medskip
\end{minipage}
\begin{minipage}[b]{0.48\linewidth}
  \centering
  \centerline{\includegraphics[width=7.4cm]{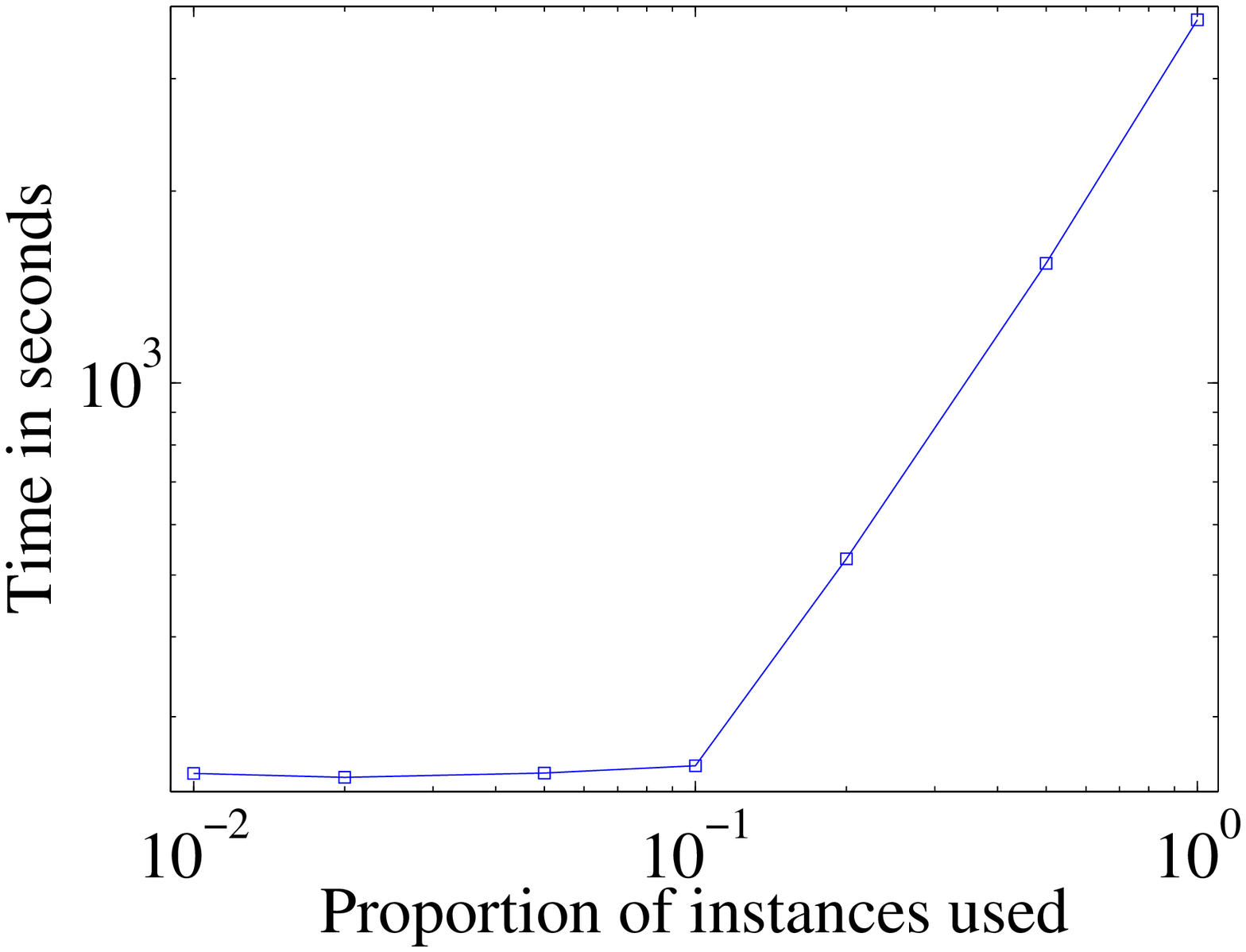}}
  \centerline{MSCV2}\medskip
\end{minipage}
\hfill
\begin{minipage}[b]{0.48\linewidth}
  \centering
  \centerline{\includegraphics[width=7.4cm]{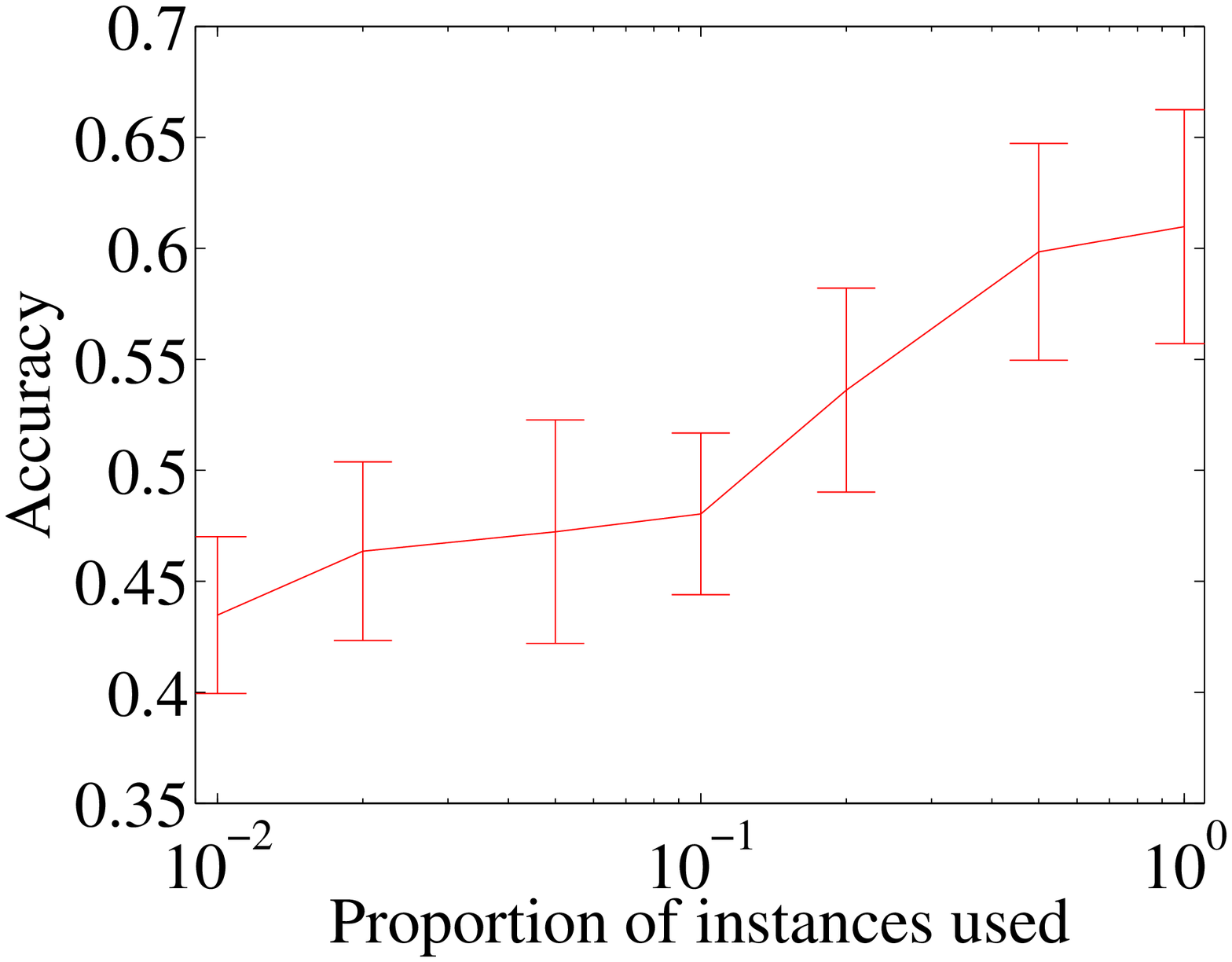}}
  \centerline{MSCV2}\medskip
\end{minipage}

\begin{minipage}[b]{0.48\linewidth}
  \centering
  \centerline{\includegraphics[width=7.4cm]{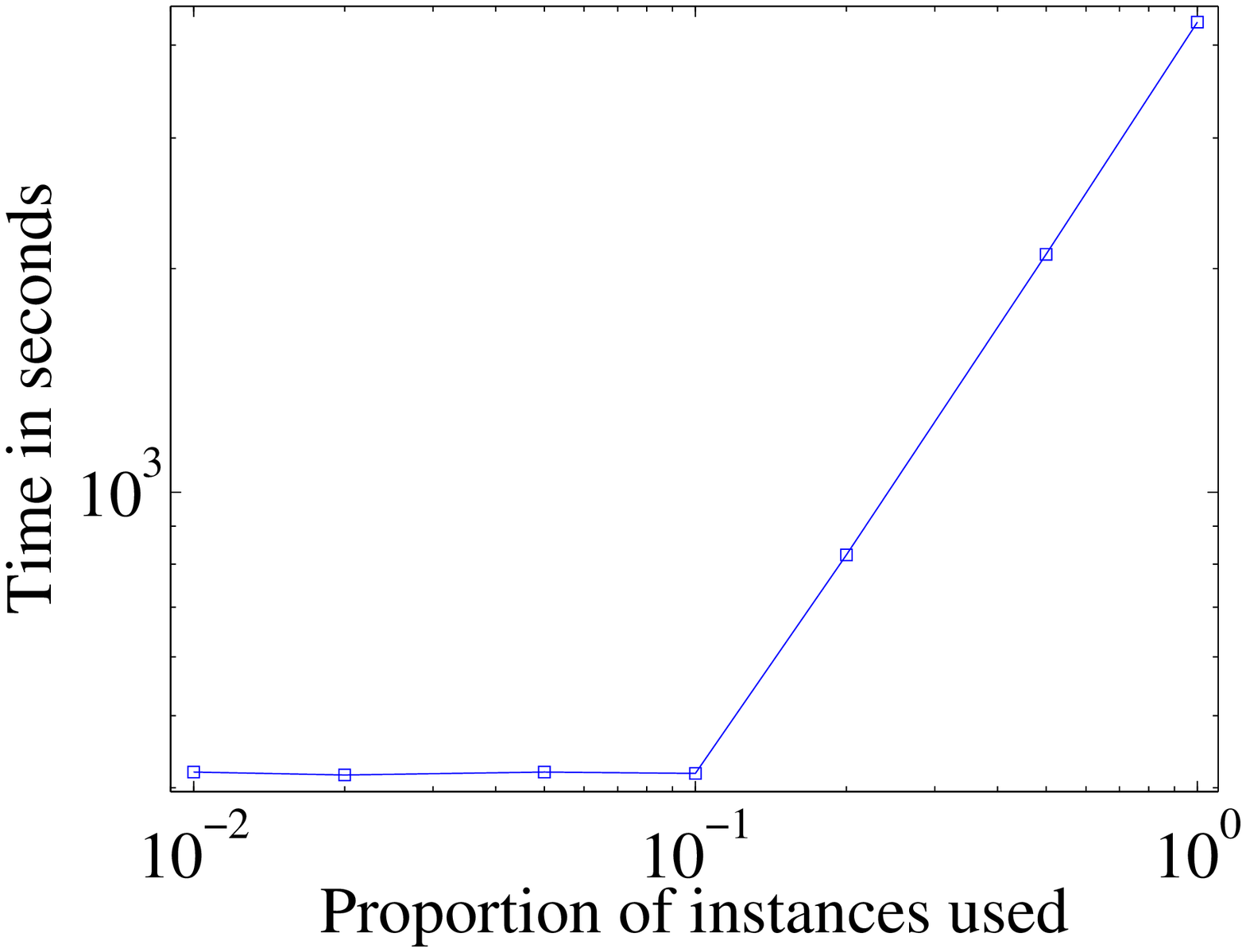}}
  \centerline{Letter Carroll}\medskip
\end{minipage}
\hfill
\begin{minipage}[b]{0.48\linewidth}
  \centering
  \centerline{\includegraphics[width=7.4cm]{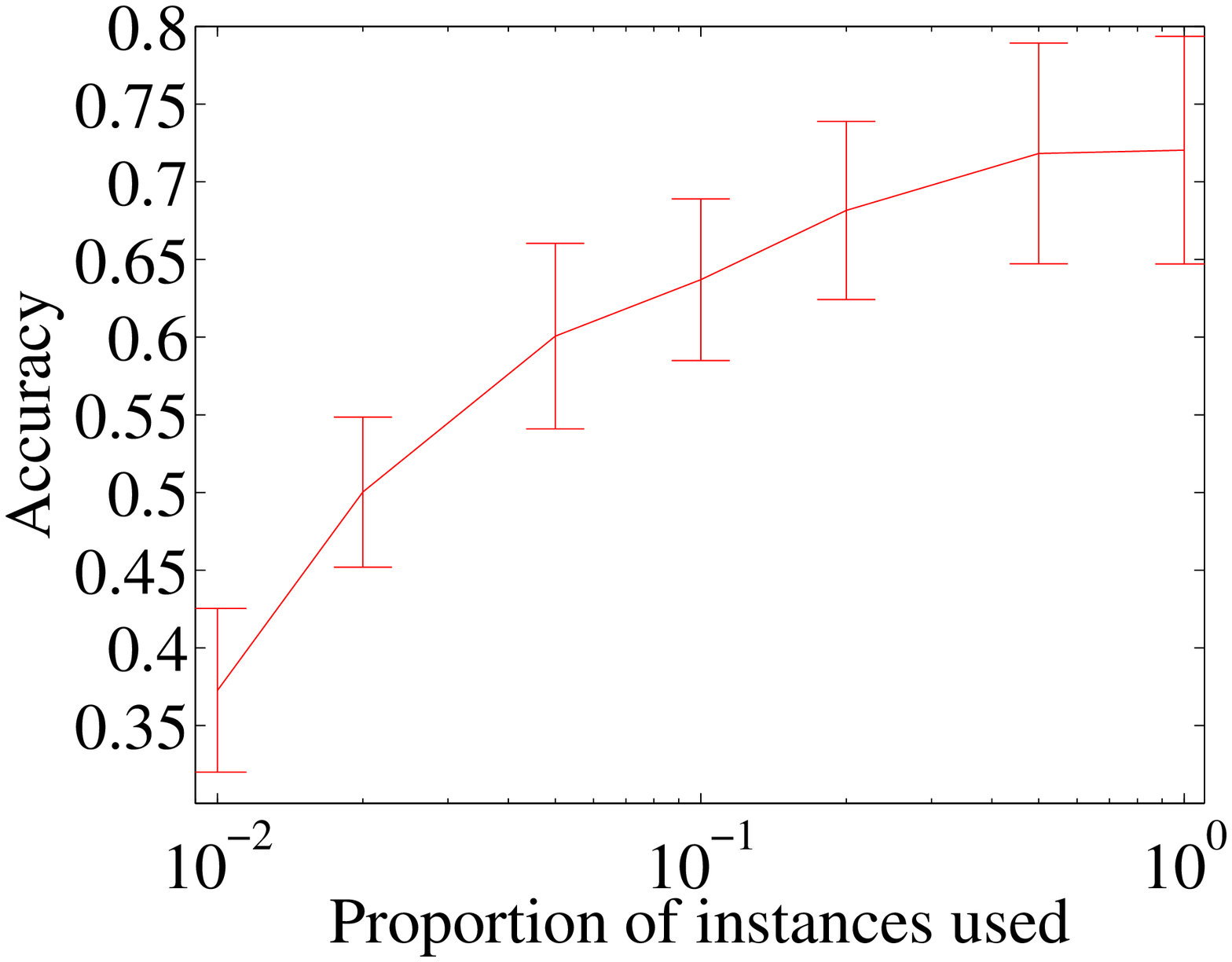}}
  \centerline{Letter Carroll}\medskip
\end{minipage}

\caption{Accuracy and runtime when changing the proportion of instances used for the kernel vector}
\label{fig:dict}
\end{figure}

In kernel learning, in order to create the kernel vector $\textbf{k}(\textbf{x})$, we compute the similarities of $\textbf{x}$ with all other instances (called dictionary), to obtain $\textbf{k}(\textbf{x})=[K(\textbf{x},\textbf{x}_{11}),K(\textbf{x},\textbf{x}_{12}),\dots,K(\textbf{x},\allowbreak\textbf{x}_{Bn_B})]$.
Shrinking dictionary means reducing the dimension of $\textbf{k}(\textbf{x})$. Specifically, from the data, we sample $S$ instances $\hat{\textbf{x}}_1,\hat{\textbf{x}}_2,\dots,\hat{\textbf{x}}_S$, and create $\textbf{k}(\textbf{x})=[K(\textbf{x},\hat{\textbf{x}}_1),K(\textbf{x},\hat{\textbf{x}}_2),\dots,K(\textbf{x},\allowbreak\hat{\textbf{x}}_S)]$.
In fact, the larger the dimension of $\textbf{k}(\textbf{x})$, the longer MLR takes to converge since one dimension of the parameter $\boldsymbol\alpha$ in Section \ref{ss:kernel} is equal to the dimension of $\textbf{k}(\textbf{x})$. Even though there are a lot of instances, they may lie on a low dimensional space. As a result, even though some instances are removed in the kernel vector, we hope that the accuracy will only mildly change.

We select $q$ the percentage of instances used to create the kernel vector in the set \{1\%, 2\%, 5\%, 10\%, 20\%, 50\%, 100\%\}. If $q\le10\%$, we set the number of EM iterations to 50. If $q>10\%$, then the dimension of the kernel vector and parameter $\boldsymbol\alpha$ become large. Consequently, in order to guarantee the convergence of the proposed EM solution, we set the number of iterations to $(q/10\%)*50=q*500$ which is proportional to $q$. For example, if we use $20\%$ of instances, we set the number of iterations to $20\%\times500=100$. The accuracy and runtime as functions of $q$ are reported in Figure \ref{fig:dict}.

From Figure \ref{fig:dict}, on the HJA, MSCV2, and Letter Carroll datasets, we observe that by creating the kernelized feature vectors containing the similarities with only 50\% of the number of instances in the dataset, we can still keep a similar level of accuracy when using all instances in the datasets. Note that for $q\le10\%$, the runtime is hardly changed since the number of expectation maximization iterations is fixed to be 50.

In summary, even though there is an exponential factor w.r.t.~the number of classes per bag in the computational complexity of the proposed MLR approach, the average number of classes per bag is not high in reality making MLR a practical instance annotation solution. Additionally, there are several speed up methods including pruning and sampling that help to reduce the runtime while keeping the accuracy almost unchanged. Especially, pruning is a technique designed for MLR that help to reduce the runtime on datasets with high average number of classes per bag such as Letter Carroll and Letter Frost by a factor from 4 to 7 times.

\section{Conclusions}
This paper focuses on the instance annotation problem in multi-label multi-instance setting. We proposed an OR-ed logistic regression model for the problem and used an expectation maximization framework to facilitate maximum likelihood estimation of the model parameters. We focused on the challenge of how to efficiently compute the exact instance label posterior probabilities given their bag level labels to keep the accuracy as high as possible. We address this challenge by using a dynamic programming method, with the computational complexity linear in the number of instances per bag. Experiments on different datasets indicate that the proposed approach outperforms other state-of-the-art methods including SIM, LSB, and Mfast, especially on MSCV2 dataset where the accuracy of the proposed approach is around 7\% higher than those of SIM, LSB, and Mfast. Our MLR approach is not only free of parameter tuning but also achieves higher accuracy, as well as uses less bags to get to the same accuracy level compared to other methods. In MSCV2, MLR just requires around 17\% training bags to achieve the same level of accuracy as SIM and LSB. Experiments on bag level classification show that even though without using bag level features, MLR is comparable to other bag level classifiers such as M-SVM and M-NN. Other experiments show that the proposed approach can be extended well to linearly inseparable datasets using kernel learning and efficiently speeded up using pruning and sampling techniques.

\acks{This work is partially supported by the National Science Foundation grants CCF-1254218 and IIS-1055113. The authors would like to thank Dr.~Jesus Perez for his help with proofreading of this paper, Dr.~Forrest Briggs for help with the HJA data and advice regarding the application of SIM \citep{briggs2012} to the data, and Mr.~Liping Liu for his advice on the application of LSB-CMM \citep{liu2012conditional} to the data. Some of the material in this paper was appeared in \cite{anh2014a} and \cite{anh2014b}.}

\vspace{15pt}
{\centering \textbf{\normalsize APPENDIX A: PROOF OF PROPOSITION \ref{p:1} }\\}
\vspace{15pt}
\noindent For $i=0$, from our model, $p(\textbf{Y}_b^{1}|\textbf{X}_b,\textbf{w})=p(y_{b1}|\textbf{X}_b,\textbf{w})$. Therefore, if $\textbf{\L}$ contains only one class $l$, $p(\textbf{Y}_b^{1}=\textbf{\L}|\textbf{X}_b,\textbf{w})=p(y_{b1}=l|\textbf{x}_{b1},\textbf{w})$. Otherwise, $p(\textbf{Y}_b^{1}=\textbf{\L},|\textbf{X}_b,\textbf{w})=0$. \\

\noindent For $i\ge 1$, in the graphical model in Figure \ref{fig:Dynamic4}(e), by marginalizing $p(\textbf{Y}_b^{i+1},\textbf{Y}_b^{i},y_{b(i+1)}|\textbf{X}_b,\textbf{w})$ over $\textbf{Y}_b^{i}$ and $y_{b(i+1)}$, we obtain
\begin{equation}
\label{e:a1}
p(\textbf{Y}_b^{i+1}=\textbf{\L}|\textbf{X}_b,\textbf{w})=\sum_{\textbf{\L}^{'}}\sum_{c}p(\textbf{Y}_b^{i+1}=\textbf{\L},\textbf{Y}_b^{i}=\textbf{\L}^{'},y_{b(i+1)}=c|\textbf{X}_b,\textbf{w}).
\end{equation}
Furthermore, since in the model $\textbf{Y}_b^{i}$ and $y_{b(i+1)}$ are independent, and from the conditional probability rule, the RHS of \eqref{e:a1} becomes
\begin{align}
p(\textbf{Y}_b^{i+1}=\textbf{\L}|\textbf{X}_b,\textbf{w})=\sum_{\textbf{\L}^{'}}\sum_{c}&p(\textbf{Y}_b^{i+1}=\textbf{\L}|\textbf{Y}_b^{i}=\textbf{\L}^{'},y_{b(i+1)}=c,\textbf{X}_b,\textbf{w})\times\nonumber\\
&p(\textbf{Y}_b^{i}=\textbf{\L}^{'}|\textbf{X}_b,\textbf{w})p(y_{b(i+1)}=c|\textbf{X}_b,\textbf{w}).\label{e:a2}
\end{align}
Using the assumption that $\textbf{Y}_b^{i+1} = \textbf{Y}_b^{i} \bigcup y_{b(i+1)}$, we can rewrite the RHS of \eqref{e:a2} as follows
\begin{align}
p(\textbf{Y}_b^{i+1}=\textbf{\L}|\textbf{X}_b,\textbf{w})=&\sum_{\textbf{\L}^{'}}\sum_{c}I(\textbf{\L}=\textbf{\L}^{'}\bigcup c)p(\textbf{Y}_b^{i}=\textbf{\L}^{'}|\textbf{X}_b,\textbf{w})p(y_{b(i+1)}=c|\textbf{X}_b,\textbf{w}) \nonumber\\
=&\sum_{c \in \textbf{\L}}p(y_{b(i+1)}=c|\textbf{X}_b,\textbf{w})\times \nonumber\\
&\hspace{6mm}[p(\textbf{Y}_b^{i}=\textbf{\L}_{\backslash c}|\textbf{X}_b,\textbf{w})+p(\textbf{Y}_b^{i}=\textbf{\L}|\textbf{X}_b,\textbf{w})].\label{e:appa}
\end{align}
From our model, $y_{b(i+1)}$ is only depend on $\textbf{x}_{b(i+1)}$ and $\textbf{w}$, therefore we obtain
\begin{align*}
p(\textbf{Y}_b^{i+1}=\textbf{\L}|\textbf{X}_b,\textbf{w})=\sum_{c \in \textbf{\L}}&p(y_{b(i+1)}=c|\textbf{x}_{b(i+1)},\textbf{w})\times\\
&[p(\textbf{Y}_b^{i}=\textbf{\L}_{\backslash c}|\textbf{X}_b,\textbf{w})+p(\textbf{Y}_b^{i}=\textbf{\L}|\textbf{X}_b,\textbf{w})].
\end{align*}

\vspace{15pt}
{\centering \textbf{\normalsize APPENDIX B: PROOF OF PROPOSITION \ref{p:2} }\\}
\vspace{15pt}
\noindent Similar to the proof for Proposition \ref{p:1}, by marginalizing $p(\textbf{Y}_b^{n_b},\textbf{Y}_b^{n_b-1},y_{bn_b}|\textbf{X}_b,\textbf{w})$ over $\textbf{Y}_b^{n_b-1}$, we have
\begin{equation}
\label{e:b1}
p(y_{bn_b}=c, \textbf{Y}_b^{n_b}=\textbf{\L}|\textbf{X}_b,\textbf{w})=\sum_{\textbf{\L}^{'}}p(\textbf{Y}_b^{n_b}=\textbf{\L},\textbf{Y}_b^{n_b-1}=\textbf{\L}^{'},y_{bn_b}=c|\textbf{X}_b,\textbf{w}).
\end{equation}
Since $\textbf{Y}_b^{n_b-1}$ and $y_{bn_b}$ are independent, and from the conditional probability rule, the RHS of \eqref{e:b1} becomes
\begin{align}
p(y_{bn_b}=c, \textbf{Y}_b^{n_b}=\textbf{\L}|\textbf{X}_b,\textbf{w})=\sum_{\textbf{\L}^{'}}&p(\textbf{Y}_b^{n_b}=\textbf{\L}|\textbf{Y}_b^{n_b-1}=\textbf{\L}^{'},y_{bn_b}=c,\textbf{X}_b,\textbf{w})\times\nonumber\\
&p(\textbf{Y}_b^{n_b-1}=\textbf{\L}^{'}|\textbf{X}_b,\textbf{w})p(y_{bn_b}=c|\textbf{X}_b,\textbf{w}).\label{e:b2}
\end{align}
Using the assumption that $\textbf{Y}_b^{n_b} = \textbf{Y}_b^{n_b-1} \bigcup y_{bn_b}$, and since $y_{bn_b}$ depends on only $\textbf{x}_{bn_b}$ and $\textbf{w}$, we can rewrite the RHS of \eqref{e:b2} as follows
\begin{align*}
p(y_{bn_b}=c, \textbf{Y}_b^{n_b}=\textbf{\L}|\textbf{X}_b,\textbf{w})=&\sum_{\textbf{\L}^{'}}I(\textbf{\L}=\textbf{\L}^{'}\bigcup c)p(\textbf{Y}_b^{n_b-1}=\textbf{\L}^{'}|\textbf{X}_b,\textbf{w})p(y_{bn_b}=c|\textbf{X}_b,\textbf{w})\\
=&p(y_{bn_b}=c|\textbf{x}_{bn_b},\textbf{w})[p(\textbf{Y}_b^{n_b-1}=\textbf{\L}_{\backslash c}|\textbf{X}_b,\textbf{w})+p(\textbf{Y}_b^{n_b-1}=\textbf{\L}|\textbf{X}_b,\textbf{w})].
\end{align*}

\vspace{15pt}
{\centering \textbf{\normalsize APPENDIX C: PROOF OF PROPOSITION \ref{p:4} }\\}
\vspace{15pt}
\noindent From \eqref{e:appa} in the proof of Proposition \ref{p:1} we obtain
\begin{equation}
p(\textbf{Y}_b=\textbf{\L}|\textbf{X}_b,\textbf{w})=\sum_{\textbf{\L}^{'}}\sum_{c}I(\textbf{\L}=\textbf{\L}^{'} \bigcup c)p(\textbf{Y}_b^{\backslash i}=\textbf{\L}^{'}|\textbf{X}_b,\textbf{w})p(y_{bi}=c|\textbf{x}_{bi},\textbf{w}).
\label{e:forwardremoving}
\end{equation}
Then,  from \eqref{e:forwardremoving} and by definition of $\textbf{A}$, $\textbf{u}$, and $\textbf{v}$, we obtain the relation $\textbf{u}=\textbf{A}\textbf{v}$.

\bibliography{strings,refs}

\end{document}